\documentclass[default,iicol]{sn-jnl}

\usepackage{graphicx}
\usepackage{multirow}
\usepackage{amsfonts}%
\usepackage{amsthm}%
\usepackage{mathrsfs}%
\usepackage[title]{appendix}%
\usepackage{textcomp}%
\usepackage{manyfoot}%
\usepackage{algorithm}%
\usepackage{algorithmicx}%
\usepackage{algpseudocode}%
\usepackage{listings}%
\usepackage{booktabs}
\usepackage{amsmath}
\usepackage{amssymb}

\usepackage{bm}
\usepackage{xspace}
\usepackage{subcaption}
\usepackage{enumitem}
\usepackage{threeparttable}
\usepackage[table,xcdraw]{xcolor}
\usepackage{hyperref}
\newcommand{\best}[1]{\textcolor{red}{{#1}}}
\newcommand{\second}[1]{\textcolor{blue}{{#1}}}

\usepackage[capitalize]{cleveref}
\crefname{section}{Sec.}{Secs.}
\Crefname{section}{Section}{Sections}
\Crefname{table}{Table}{Tables}
\crefname{table}{Tab.}{Tabs.}
\def\y{\mathbf{y}}
\def\z{\mathbf{z}}
\def\M{\mathbf{M}}
\def\v{\mathbf{v}}
\newcommand{\round}[1]{\left\lfloor #1 \right\rceil}

\usepackage{color,soul}
\usepackage{xcolor}

\newcommand{\wh}[1]{\textcolor{black}{#1}}

\theoremstyle{thmstyleone}%
\newtheorem{theorem}{Theorem}
\theoremstyle{thmstyletwo}%

\theoremstyle{thmstylethree}%

\begin{document}

\title[Article Title]{Beyond Learned Metadata-based Raw Image Reconstruction}


\author[1]{\fnm{Yufei} \sur{Wang}}\email{yufei001@ntu.edu.sg}

\author[1]{\fnm{Yi} \sur{Yu}}\email{yuyi0010@ntu.edu.sg}

\author[2]{\fnm{Wenhan} \sur{Yang}}\email{yangwh@pcl.ac.cn}

\author[1]{\fnm{Lanqing} \sur{Guo}}\email{lanqing001@ntu.edu.sg}

\author[3]{\fnm{Lap-Pui} \sur{Chau}}\email{lap-pui.chau@polyu.edu.hk}

\author[1]{\fnm{Alex C.} \sur{Kot}}\email{eackot@ntu.edu.sg}

\author*[1]{\fnm{Bihan} \sur{Wen}}\email{bihan.wen@ntu.edu.sg}

\affil[1]{\orgdiv{ROSE Lab}, \orgname{Nanyang Technological University}, \orgaddress{\street{50 Nanyang Ave}, \postcode{639798}, \country{Singapore}}}

\affil[2]{\orgdiv{Peng Cheng Laboratory}, \orgaddress{\street{No. 2, Xingke 1st Street}, \city{Shenzhen}, \postcode{518066}, \country{China}}}

\affil[3]{\orgdiv{Department of Electronic and Information Engineering}, \orgname{The Hong Kong Polytechnic University}, \orgaddress{\street{11 Yuk Choi Rd}, \city{Hong Kong}, \country{China}}}


\abstract{While raw images possess distinct advantages over sRGB images, \textit{e.g.}, linearity and fine-grained quantization levels, they are not widely adopted by general users due to their substantial storage requirements.
Very recent studies propose to compress raw images by designing sampling masks within the pixel space of the raw image. However, these approaches often leave space for pursuing more effective image representations and compact metadata.
In this work, we propose a novel framework that learns a compact representation in the latent space, serving as metadata, in an end-to-end manner. 
Compared with lossy image compression, we analyze the intrinsic difference of the raw image reconstruction task caused by rich information from the sRGB image. 
Based on the analysis, a novel design of the backbone with asymmetric and hybrid spatial feature resolutions is proposed, which significantly improves the rate-distortion performance.
Besides, we propose a novel design of the sRGB-guided context model, which can better predict the order masks of encoding/decoding based on both the sRGB image and the the masks of already processed features.
Benefited from the better modeling of the correlation between order masks, the already processed information can be better utilized.
Moreover, a novel sRGB-guided adaptive quantization precision strategy, which dynamically assigns varying levels of quantization precision to different regions, further enhances the representation ability of the model. 
Finally, based on the iterative properties of the proposed context model, we propose a novel strategy to achieve variable bit rates using a single model. This strategy allows for the continuous convergence of a wide range of bit rates.
We demonstrate how our raw image compression scheme effectively allocates more bits to image regions that hold greater global importance.
Extensive experimental results validate the superior performance of the proposed method, achieving high-quality raw image reconstruction with a smaller metadata size, compared with existing SOTA methods.}




\maketitle
\section{Introduction}

\begin{figure}[t]
    \centering
     \begin{subfigure}{1\linewidth}
    \includegraphics[width=1\linewidth]{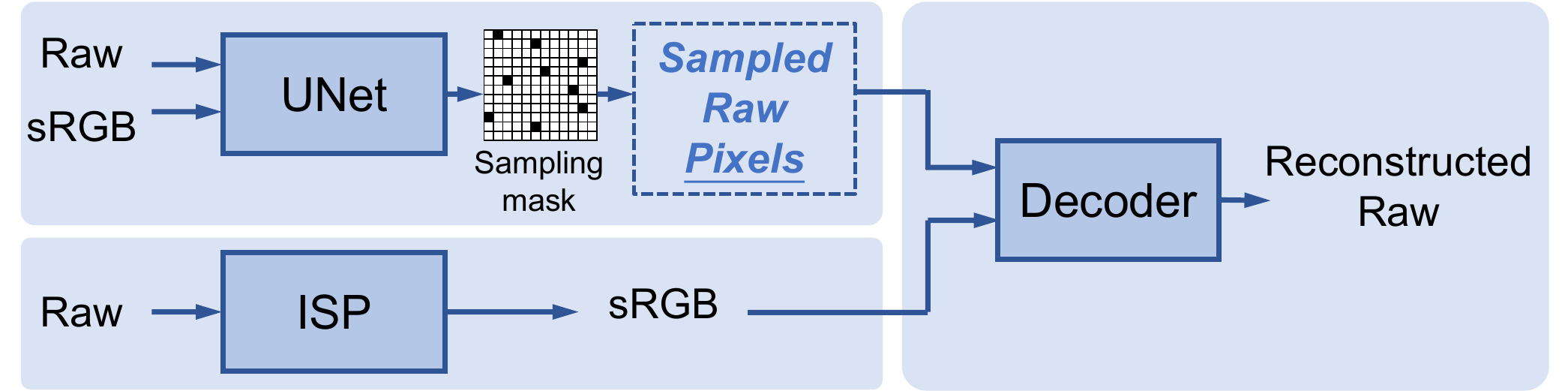}
    \caption{Previous SOTA methods sample in the raw space~\citep{punnappurath2021spatially, nam2022learning}.}
     \end{subfigure}
     \begin{subfigure}{1\linewidth}
    \includegraphics[width=1\linewidth]{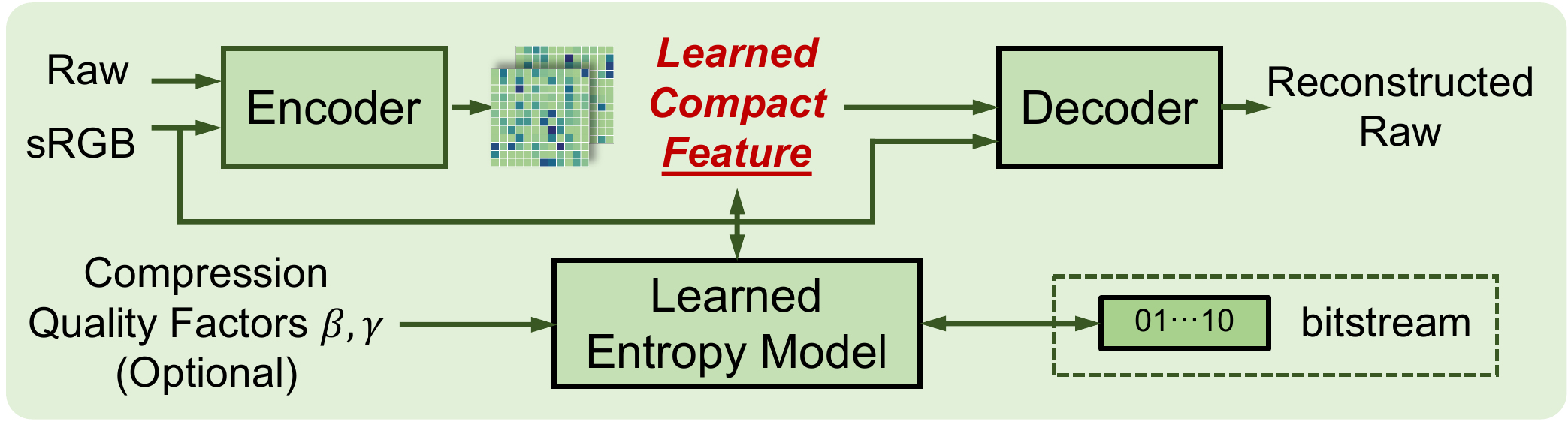}
    \vspace{-0.3cm}
    \caption{The proposed method samples in the latent space.}
     \end{subfigure}
    \vspace{-0.4cm}
    \caption{
    The comparison between the previous SOTA methods (in blue) and our proposed method (in green).
    Different from the previous work where the sampling strategy is hand-crafted or learned by a pre-defined sampling loss, we learn the sampling and reconstruction process in a unified end-to-end manner. 
    In addition, the sampling of previous works is in the raw pixel space, which in fact still includes a large amount of spatial redundancy and precision redundancy.
    Instead, we conduct sampling in the feature space, and more compact metadata is obtained for pixels in the feature space via the adaptive allocation. The saved metadata is annotated in the dashed box. 
    Compared with our conference version~\cite{wang2023raw}, we propose a variable bit rate strategy that can continuously converge a large range of bit rates. Besides, better RD performance is achieved by the improved entropy model and backbone design.
    }
    \vspace{-0.15cm}
    \label{fig:intro}

\end{figure}

Raw images, acquired directly from camera sensors and in an unprocessed and uncompressed data format, provide distinct benefits for computer vision applications, \textit{e.g.}, image denoising~\citep{zhang2021rethinking, abdelhamed2019noise}, low-light enhancement~\citep{wei2020physics, huang2022towards, wang2022low}, brightness correction, and artistic manipulation.

%
%

Despite the benefits of raw images, they are not widely used by ordinary users due to their large file sizes. 
%
%
To address the storage efficiency issue, there is a growing interest in the raw-image reconstruction task, which aims to minimize the amount of required metadata to convert sRGB images back to the raw space.
Classic metadata-based raw image reconstruction methods model the image signal processing (ISP) pipeline and save the required ISP parameters as metadata~\citep{nguyen2016raw}.
\begin{figure}[tbp]
    \centering
    \begin{subfigure}{0.98\linewidth}
    \includegraphics[width=1\linewidth, trim=0 15 0 0, clip]{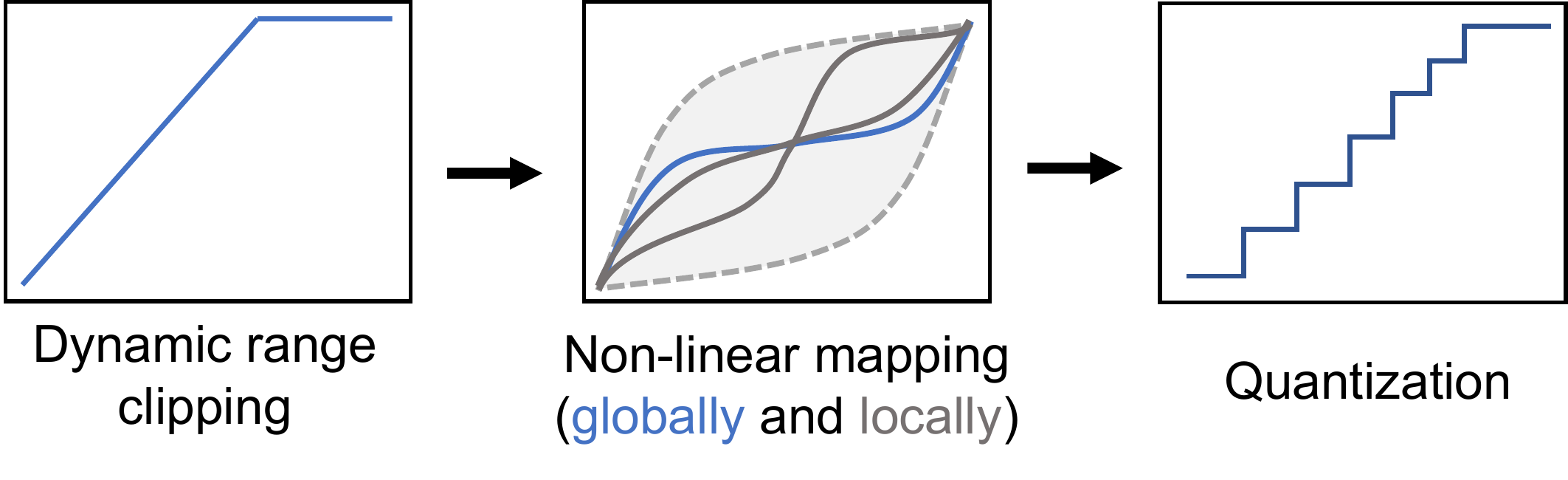}
    \caption{Simplified ISP adopted from \citep{liu2020single}}
    \end{subfigure}\\
    \begin{subfigure}{0.3\linewidth}
    \includegraphics[width=1\linewidth]{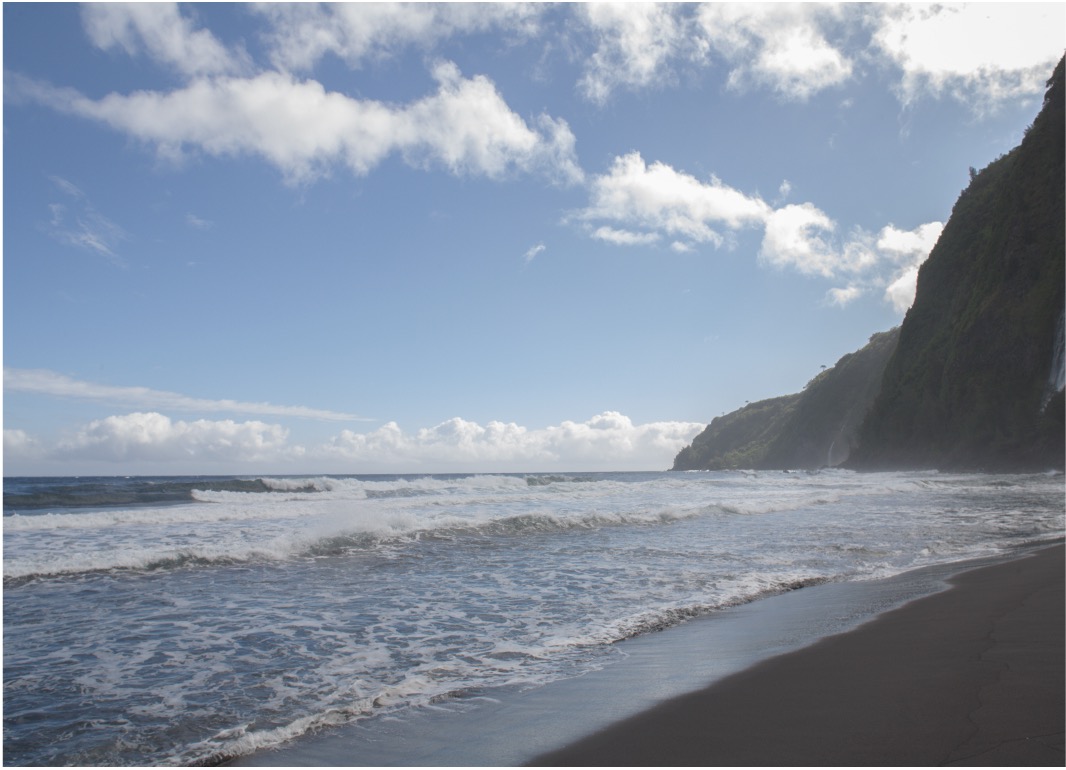}
    \caption{Processed Raw}
    \end{subfigure}
    \begin{subfigure}{0.3\linewidth}
    \includegraphics[width=1\linewidth]{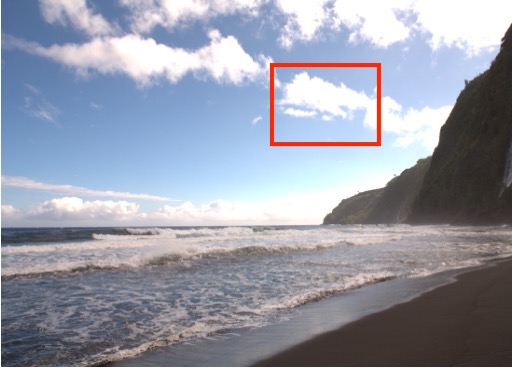}
    \caption{Quantized sRGB}
    \end{subfigure}
    \begin{subfigure}{0.37\linewidth}
    \includegraphics[width=1\linewidth]{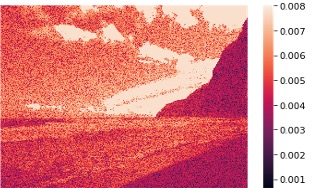}
    \caption{Quantization error map}
    \end{subfigure}
    
    \vspace{-0.1cm}
    \caption{
    An illustration of the non-uniform information loss caused by the ISP. 
    (a) A simplified ISP, which involves non-linear mapping both globally and locally. 
    (b) Raw image after post-process to better display the details. 
    (c) Quantized sRGB image after ISP which suffers information loss, \textit{e.g.}, the red bounding box area.
    (d) The quantization error map. 
    As we can see from the above figures, the information loss caused by the quantization is non-uniformly distributed in both over-exposed areas and normally-exposed areas.
    }
    \label{fig:isp}
    \vspace{-0.2cm}
\end{figure}
Recent methods focus on sparse sampling of raw image pixels to achieve a lightweight and flexible reverse ISP reconstruction while reducing storage and computational complexity. This approach is discussed in two very recent studies \citep{punnappurath2021spatially, nam2022learning}. In \citep{punnappurath2021spatially}, a uniform sampling strategy is proposed, which is combined with an interpolation algorithm by solving systems of linear equations. The other study \citep{nam2022learning} proposes a sampling network and uses a deep convolutional network to approximate the reconstruction process, thereby further improving the sampling strategy and reconstruction quality.

While existing sparse sampling-based raw image reconstruction methods have made great progress, they still have limitations in terms of coding efficiency and image reconstruction quality. 
Given the non-linear transformation and quantization steps in ISP as shown in Fig. \ref{fig:isp}, to achieve better rate-distortion (RD) performance, the bit allocation for an image should be adaptive and globally optimized. For instance, smooth regions of a raw image can be accurately reconstructed with sparser samples, while texture-rich regions require denser sampling. However, in existing practices, including the state-of-the-art method \citep{nam2022learning} which enforces locally non-uniform sampling, the sampling is almost uniform from a global perspective. This results in metadata redundancy and limits the reconstruction performance. Additionally, recent works~\citep{punnappurath2021spatially, nam2022learning} use a fixed sampling space, namely the raw image space, with a fixed bit depth for sampled pixels, which limits the representation ability and results in precision redundancy.

To overcome the issues mentioned above, we propose a novel end-to-end learned raw image joint \textit{sample-reconstruction} framework, which does not rely on pre-defined sampling strategies or sampling losses, \textit{e.g.}, super-pixel loss. As shown in Fig. \ref{fig:intro}, our framework is based on the encoded latent feature with the entropy constraint. 
Additionally, we propose an sRGB-guided context model to further improve the rate-distortion performance. 
Compared with the context model in our conference work~\cite{wang2023raw}, our improved context model additionally considers the correlations between different steps in the sampling order mask. By incorporating both sRGB images and the mask of already processed features to guide the distribution modeling process, our enhanced context model significantly boosts the RD performance of the model compared with the model w/o a context model and the model w/ the proposed context model in our conference work~\cite{wang2023raw}.

Besides, although the proposed framework shares a similar core with the existing works on end-to-end lossy image compression~\cite{theis2017lossy, balle2016end, agustsson2017soft}, directly applying their backbone design and deep entropy coding strategy leads to a sub-optimal solution due to the intrinsic difference of the metadata-based image reconstruction task. 
Specifically, since sRGB images themselves already contain most of the information necessary to describe the scene, reconstructing a raw image  requires significantly less information compared with situations where sRGB is not available. This results in a smaller coding range for the latent code values and a more serious spatial redundancy of the latent codes.
Therefore, while added uniform noise performs well in lossy image compression as a substitute for the non-differentiable quantization operation, the smaller coding range greatly amplifies the approximation error between training and evaluation.
Additionally, the fixed quantization width restricts the representation ability of the model within a narrow coding range.

To further enhance the rate-distortion (RD) performance of the proposed joint sample-reconstruction framework, we propose an improved backbone design. This design utilizes down-sampled features for entropy coding while preserving high spatial resolution for decoding. Such a design leads to improved bit rate, faster coding speed, and higher reconstruct quality.
Besides, we propose a novel adaptive quantization width strategy, which adaptively assigns different quantization precision to the latent code to achieve better representation ability. Additionally, we adopt an asymmetric hybrid approximation strategy for the quantization step, \textit{i.e.}, the spatial resolution of features of encoder and decoder is different and asymmetric, which effectively bridges the gap between training and evaluation. 

\wh{Our contributions are summarized} as follows\wh{,}
\begin{enumerate}
\setlength{\itemsep}{0pt}
\setlength{\parsep}{0pt}
\setlength{\parskip}{0pt}
    \item We propose the first end-to-end deep encoding framework for raw image reconstruction, by fully optimizing the use of stored metadata.
    %
    \item A novel sRGB-guided context model is proposed, which leads to better reconstruction quality, smaller size of metadata, and faster speed.
    %
    \item We evaluate our method over popular raw image datasets. The experimental results demonstrate that we can achieve better reconstruction quality with less metadata required compared with SOTA methods.
\end{enumerate}

As an extension of our conference paper~\cite{wang2023raw}, the improvements are mainly in the following aspects:
\begin{enumerate}
    \item 
    We propose a more precise model to predict the ordering mask used in the context model. Specifically, we incorporate the mask of already processed latent features as an additional input to predict the mask of latent features yet to be processed.
    Besides, in conjunction with our context model, we further propose a novel variable bit rate strategy that enables a single model to cover a wide range of bit rates.
    
    \item We propose a novel approach \wh{for} adaptive quantization bin width, which allows the improved model to dynamically allocate varying quantization precision to different latent codes.
    Additionally, we introduce a hybrid strategy for approximating the quantization step during training. These advancements in the quantization step contribute to further improvements in RD performance.
    
    \item We examine the advantages and disadvantages of the backbone design by analyzing the causes of the approximation error between training and evaluation. 
    To address these issues, we propose a novel asymmetric hybrid backbone that combines the strengths of both the mainstream lossy image compression backbone and our previous conference version.
    This hybrid backbone achieves superior performance compared to existing methods. Additionally, we introduce a unified design for the hyper-prior module, simplifying the design of the overall framework.

    \item We further assess the performance of our models under a more challenging and realistic scenario, where we consider sRGB images after a complex ISP instead of a simplified software ISP. 
    Additionally, we investigate the ability of different methods to preserve information in shadow and highlight areas. Our proposed method consistently demonstrates superior rate-distortion (RD) performance and robustness across various settings.
\end{enumerate}
Overall, compared with our conference work~\cite{wang2023raw}, we achieve significant improvement in terms of both rate-distortion performance (around 70\% saving of the size of metadata) and the coding speed (2 times faster) on standard benchmark~\cite{nam2022learning} with a better PSNR.

\section{Related Work}
\subsection{Raw image reconstruction}
There are two main categories of raw image reconstruction techniques: blind raw reconstruction and raw reconstruction with metadata.

\begin{figure*}[tbp]
    \centering
    \includegraphics[width=\linewidth, trim=0 0 0 0 0 ,clip]{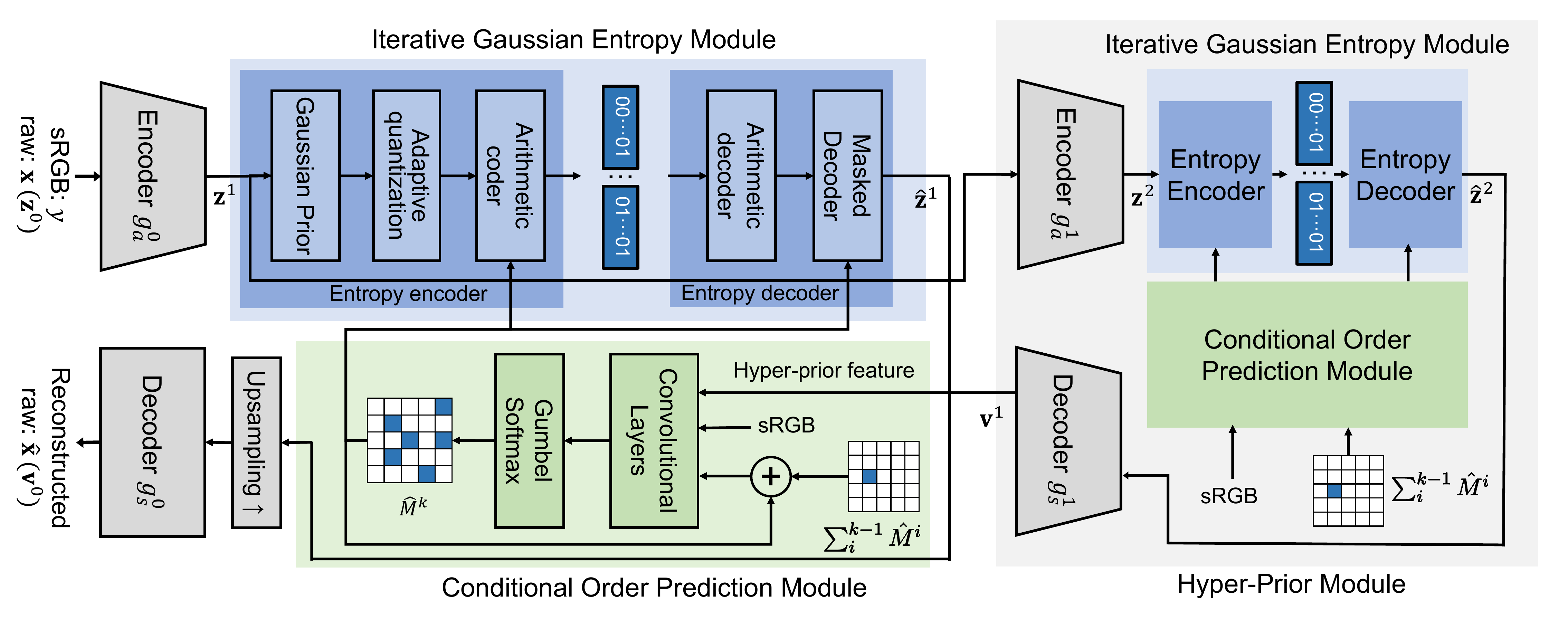}
    \vspace{-0.4cm}
    \caption{
    The overall framework of our method includes two levels that share a similar design. 
    The second level (the hyper prior module) aims to further reduce the redundancy in the latent feature of the first level.
    For each level, the encoding/decoding process \wh{maps} the raw image/latent feature into the latent space and vice versa.
    The conditional order prediction module predicts the order of encoding/decoding, and more details can be seen in Fig.\ref{fig:compare_mask_pre}. 
    Given the predicted order mask, the sRGB-guided context model, which is based on the iterative Gaussian entropy model as elaborated in Fig.~\ref{fig:context}, \wh{encodes/decodes} the latent variable $\hat{\mathbf{z}}$ into/from the \wh{bitstream}. For the sake of brevity, we omit part of the sRGB image that serves as the input of the modules in the figure.
    }
    \label{fig:framework}
    \vspace{-0.1cm}
\end{figure*}

\noindent\textbf{Blind raw reconstruction.} Blind raw reconstruction focuses on reconstructing the raw image solely based on the rendered sRGB image \cite{wang2021deep, zheng2021ultra}. Initially, early works aimed to restore the image's linearity through radiometric calibration \cite{debevec2008recovering}. Subsequently, more sophisticated models such as \cite{chakrabarti2014modeling, kim2012new, gong2018rank} were proposed to better describe the ISP (Image Signal Processing) pipeline workflows. The advancement of deep learning has led to the rapid development of deep learning-based models. For instance, \cite{marnerides2018expandnet} directly learns the mapping from low dynamic range (LDR) to high dynamic range (HDR), while \cite{liu2020single} utilizes three specialized CNNs to preserve the subdivided pipeline from HDR to LDR. More recently, \cite{xing2021invertible} proposes the use of an invertible network to learn mappings between sRGB space and raw space, and vice versa. Despite significant progress, the fidelity of the reconstructed images is inevitably constrained due to information loss during the ISP pipeline, such as quantization and dynamic range clipping.

\noindent\textbf{Raw reconstruction with metadata.} To further enhance the fidelity of raw image reconstruction, an alternative approach is to save additional metadata that assists in the reconstruction process \cite{punnappurath2019learning, yuan2011high, nam2022learning}. For example, Yuan \cite{yuan2011high} proposes saving a low-resolution raw file to model the tone mapping curve, while Nguyen \cite{nguyen2016raw, nguyen2018raw} suggests saving the estimated parameters of the simplified ISP pipeline. Punnappurath et al. \cite{punnappurath2021spatially} introduce a spatially-aware algorithm that estimates interpolation parameters based on uniformly sampled raw image pixels during test time. In a recent work by Nam et al. \cite{nam2022learning}, the sampling strategy is improved by selecting representative raw pixels based on the superpixel algorithm. Additionally, a UNet is adopted \cite{nam2022learning} to further expedite the inference process. However, the training of its sampling network relies on a predefined loss function, resulting in a suboptimal sampling strategy that affects the restoration performance. In contrast to previous approaches that typically save discrete pixels of raw images, we propose an end-to-end network capable of learning to extract necessary metadata in the latent space.

\subsection{Learned image compression}
Recently, a significant number of deep learning-based image compression methods have been proposed, achieving promising results \cite{balle2016end, li2018learning, hu2021learning}. The development of differential quantization and rate estimation has enabled end-to-end training \cite{theis2017lossy, balle2016end, agustsson2017soft}. Moreover, the introduction of contextual models \cite{cheng2020learned, minnen2018joint, lee2018context} has substantially improved the compression rate of learned compression models, garnering increasing attention. Specifically, works in \cite{minnen2018joint, lee2018context} propose the utilization of autoregressive models to leverage information that has already been decompressed from the bitstream. However, due to the nature of context models, both the compression and decompression processes become excessively slow for high-resolution images. To mitigate this issue, \cite{minnen2020channel} introduces a channel-conditioning approach, while \cite{he2021checkerboard} proposes a checkerboard context model. In our conference paper~\cite{wang2023raw}, we propose to predict the ordering mask of the context model given the sRGB image. Additionally, although learning-based image compression exhibits promising results at low bit rate scenarios, the network architecture needs to be meticulously designed for high-fidelity requirements \cite{helminger2020lossy, mentzer2020high}. Besides, compression approaches to support variable rates using a single mode are explored recently~\cite{choi2019variable, song2021variable}.
\section{Methodology}
\subsection{Motivation}
Our objective is to reconstruct the original raw image, denoted as $\mathbf{x}$, from the sRGB image $\mathbf{y}$ obtained after the ISP pipeline, which exhibits a linear relationship with the scene radiance. However, the process from the raw image to the sRGB image involves non-linear operations such as quantization and tone mapping, leading to spatially non-uniform information loss, as depicted in Fig. \ref{fig:isp}.
Unlike previous approaches that uniformly or approximately uniformly store the sparse raw pixel values using a fixed number of bits, we propose an end-to-end learning-based approach to encode the information in the latent space. Our method adaptively allocates a variable number of bits for each pixel, aiming for a more compact representation while ensuring high-fidelity reconstruction.
As illustrated in Fig. \ref{fig:framework}, our approach aims to obtain a compact representation, denoted as $\mathbf{\hat{z}}^1$, of the image conditioned on its corresponding sRGB image. The latent feature $\mathbf{\hat{z}}^1$ contains the necessary information for reconstructing the raw image accurately, with the goal of minimizing the code length.


\begin{figure*}[t]
    \centering
    \includegraphics[width=\linewidth]{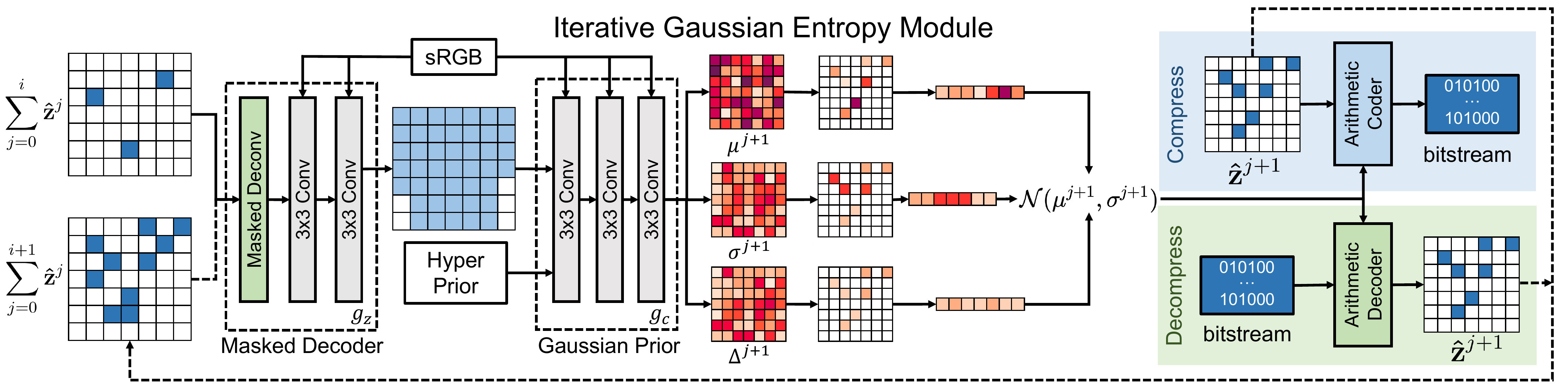}
    \caption{
    An illustration of a step of the proposed iterative Gaussian entropy model with adaptive quantization precision. 
    We model the distribution of $\mathbf{\hat{z}}^{i+1}$ based on the existing information $\sum_{j=0}^{i}\mathbf{\hat{z}}^j$ and the \wh{hyper-prior} latent feature.
    The quantitation bin width $\Delta^{i+1}$ is used when predicting the likelihood of latent features. 
    Then arithmetic coding is used to compress/decompress the latent code $\mathbf{\hat{z}}^{i+1}$ losslessly. The dashed arrows represent the following step. The superscript in this figure represents the index of the iterative step.
    }
    \label{fig:context}
    \vspace{-0.0cm}
\end{figure*}

\subsection{The overall of the framework} 
The motivation for the design of the overall framework is that we utilize a cascade pipeline to improve the coding efficiency and make the modeling of distributions unified as shown in Fig. \ref{fig:framework}.
Specifically, our framework can be formulated by
\begin{equation}
\begin{split}
    \z^{i+1} &= g_a^i(\z^i , \y ; \phi^i) \\
    \hat{\z}^{i+1} &= Q_\Delta(\z^{i+1}) \\
    \v^{i} &= g_s^i(\hat{\z}^{i+1},\y;\theta^i) \\
    i &\in \{0, 1\} ,
\end{split}
\end{equation}
where $Q_\Delta$ is the quantization operation with the bin width $\Delta$ (abbreviated as $Q$), $g_a^i$ and $g_s^i$ are the analysis and synthesis transforms of the $i_{th}$ level, and $\phi^i$ and $\theta^i$ represent the parameters of these two transforms respectively. $\z^0$ and $\v^0$ are the ground-truth raw image and the reconstructed raw image respectively, \textit{i.e.}, $\hat{\mathbf{x}}=\mathbf{v}^0$ and $\z^0 = \mathbf{x}$. Similar to the works with hyper-prior~\cite{balle2018variational, cheng2020learned}, we use a cascade architecture of two levels, and the second level is also called hyper-prior for consistency.

Given a hyper-parameter $\lambda$, the optimization objective that simultaneously minimizes the raw image reconstruction loss and the codelength of latent codes is defined as follows
\begin{equation}
\begin{split}
    \mathcal{L} &(\lambda) = \underbrace{\sum_{i} \mathcal{R}(\hat{\z}^i)}_{\text{rate}}  + \lambda \cdot \underbrace{\mathcal{D} (\hat{\mathbf{x}}, \mathbf{x})}_{\text{distortion}} \\
    &= \sum_i \mathbb{E}[-\log_2(q({\hat{\z}^i|{\v}^{i},\mathbf{y}}))] + \lambda  \mathcal{D} (\hat{\mathbf{x}}, \mathbf{x}), 
\end{split}
\label{eq:loss}
\end{equation}
where the rate $\mathcal{R}$ is the bit-per-pixel (bpp) loss, and $\mathcal{D}$ is the mean absolute error to measure the reconstruction loss. 
As shown in Eq.~\ref{eq:loss}, the hyper-prior feature $\mathbf{v}^{1}$ helps model the distribution of the latent code $\mathbf{z}^1$, therefore improving the coding efficiency. In our setting where two levels are applied, the prior knowledge of the last level $\mathbf{v}^2$ is set to an empty feature.
The analysis of the backbone design and the modeling of distributions $q_{{\hat{\z}^i|\v^{i},\mathbf{y}}}$ will be introduced below respectively.

\subsection{Analysis of the backbone design}
\begin{figure}
    \centering
    \hspace{-0.1cm}
    \begin{subfigure}{0.192\linewidth}
        \includegraphics[width=\linewidth]{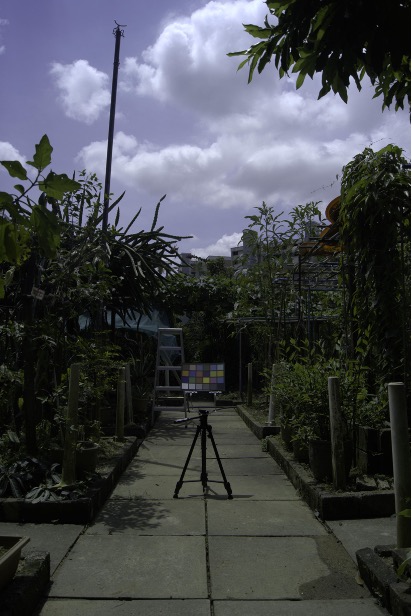}
        \caption{}
    \end{subfigure}
    \hspace{-0.1cm}
    \begin{subfigure}{0.192\linewidth}
        \includegraphics[width=\linewidth]{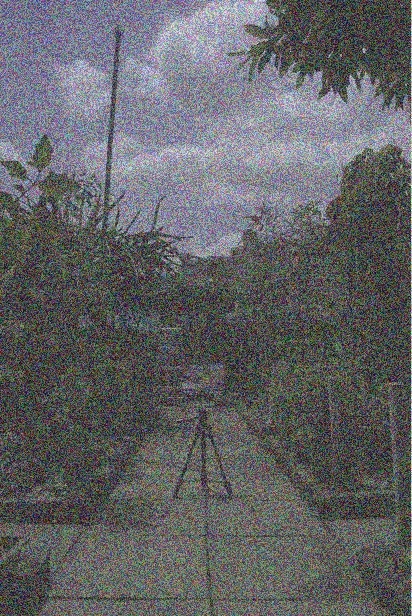}
        \caption{}
    \end{subfigure}
    \hspace{-0.1cm}
    \begin{subfigure}{0.192\linewidth}
        \includegraphics[width=\linewidth]{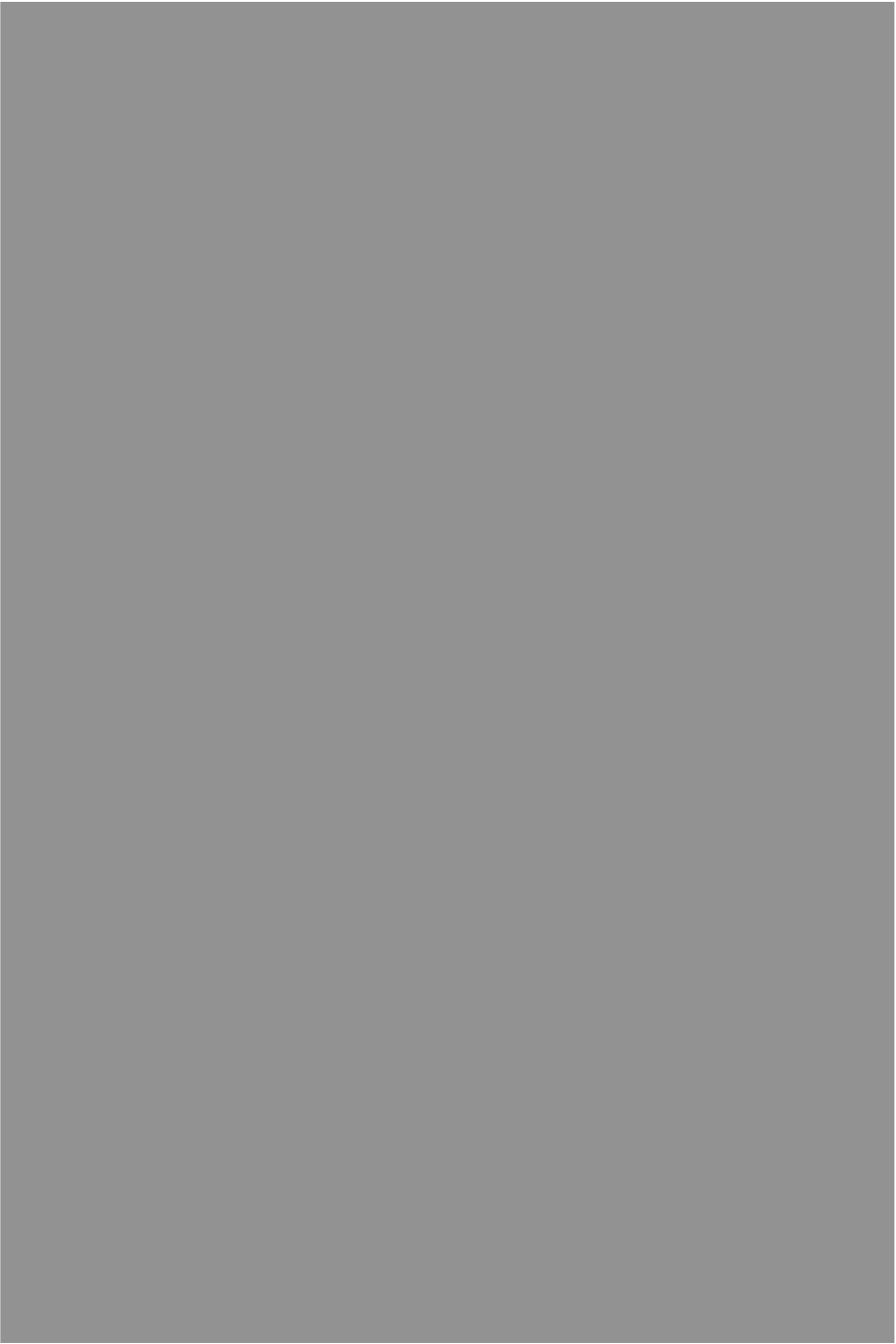}
        \caption{}
    \end{subfigure}
    \hspace{-0.1cm}
    \begin{subfigure}{0.192\linewidth}
        \includegraphics[width=\linewidth]{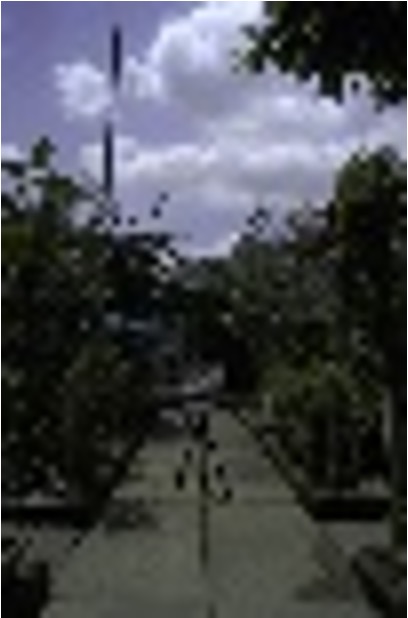}
        \caption{}
    \end{subfigure}
    \hspace{-0.1cm}
    \begin{subfigure}{0.192\linewidth}
        \includegraphics[width=\linewidth]{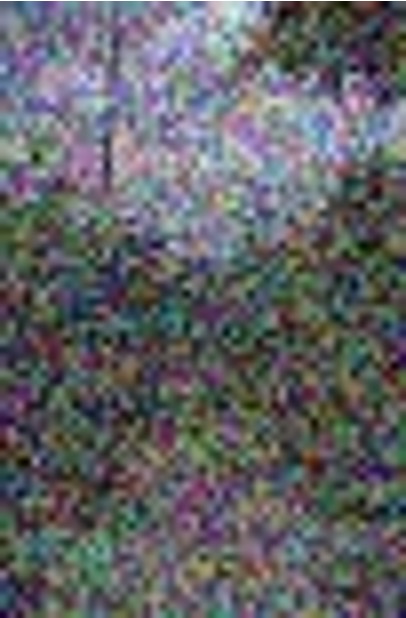}
        \caption{}
    \end{subfigure}
    \hspace{-0.1cm}
    \caption{An illustration of the impact of the latent feature resolution. Without loss of generality, we utilize an image as a substitution for the latent feature. (a) An image range from -0.5 to 0.5. (b) For training, uniform noise from -0.5 to 0.5 is added to simulate the quantization step. (c) The image after the round operation for testing. (d) The image after down-sampling. (e) The down-sampled image w/ added uniform noise. The image w/ added noise still exists lots of information when the resolution of the latent feature is high even if there is little/no information for evaluation, leading to a large gap between training and testing. This gap can be eliminated by using a smaller spatial resolution of the feature. (Figures are normalized for display.)}
    \label{fig:quantization}
\end{figure}
\begin{figure}[tbp] 
    \centering
    \begin{subfigure}{0.48\linewidth}
        \includegraphics[width=\linewidth]{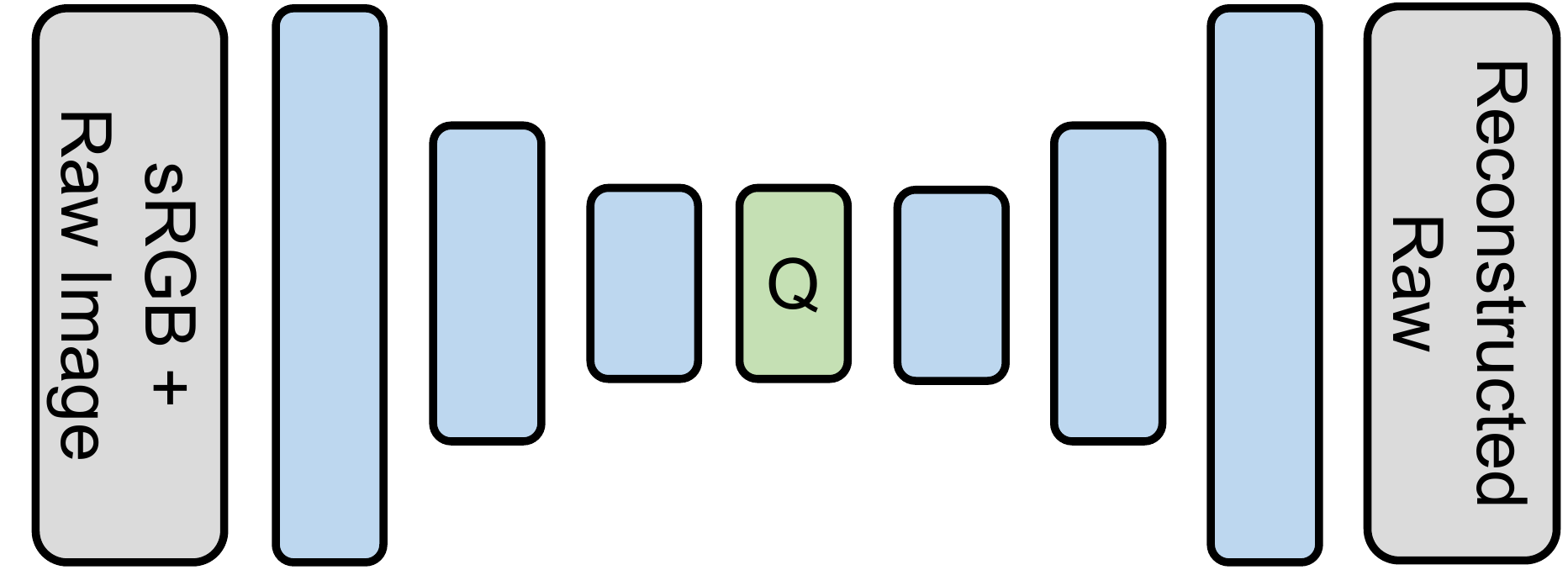}
        \caption{Downsampling\\oriented}
    \end{subfigure}\hspace{0.2cm}
    \begin{subfigure}{0.48\linewidth}
        \includegraphics[width=\linewidth]{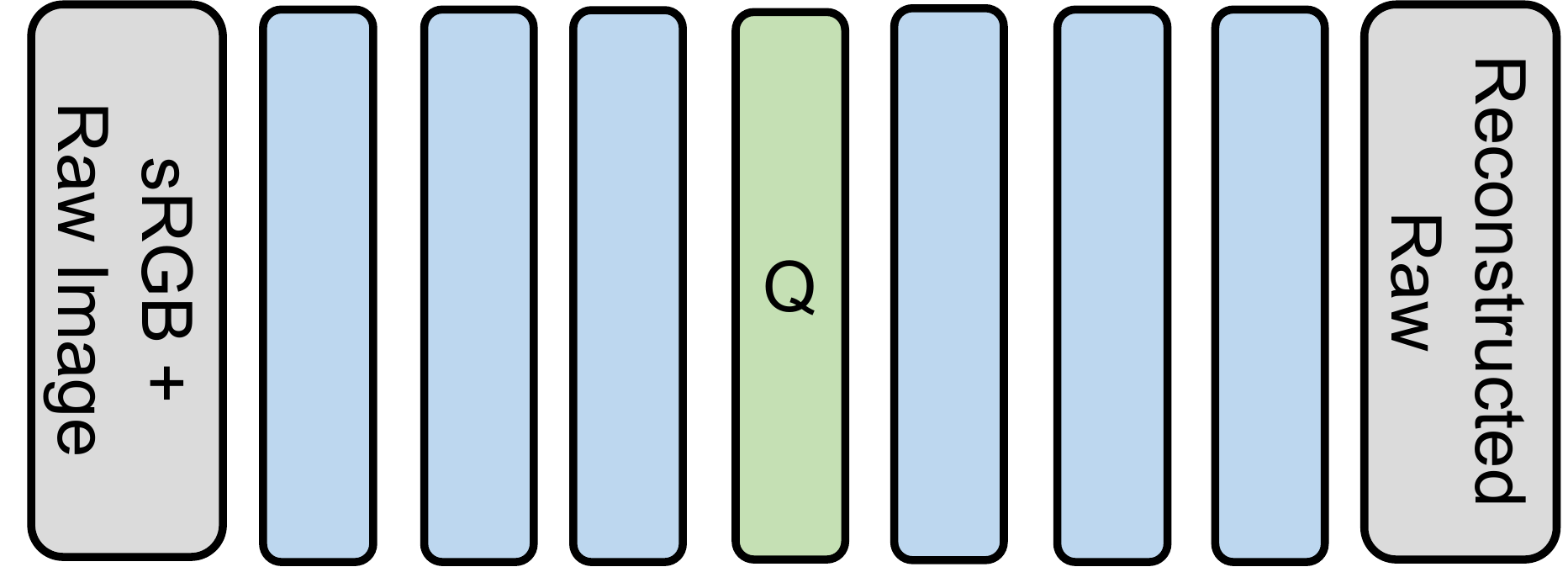}
        \caption{Resolution maintained}
    \end{subfigure}\hspace{0.2cm}
    \begin{subfigure}{0.48\linewidth}
        \includegraphics[width=\linewidth]{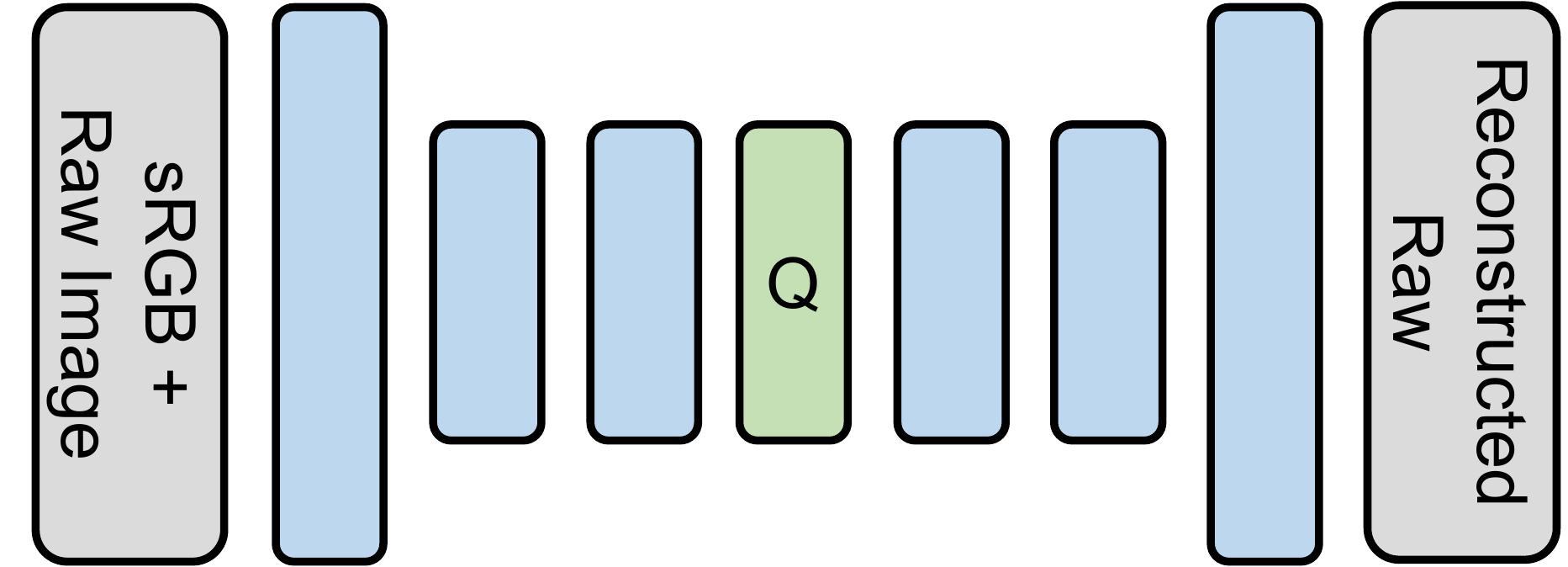}
        \caption{Finetuned\\downsampling}
    \end{subfigure}\hspace{0.2cm}
    \begin{subfigure}{0.48\linewidth}
        \includegraphics[width=\linewidth]{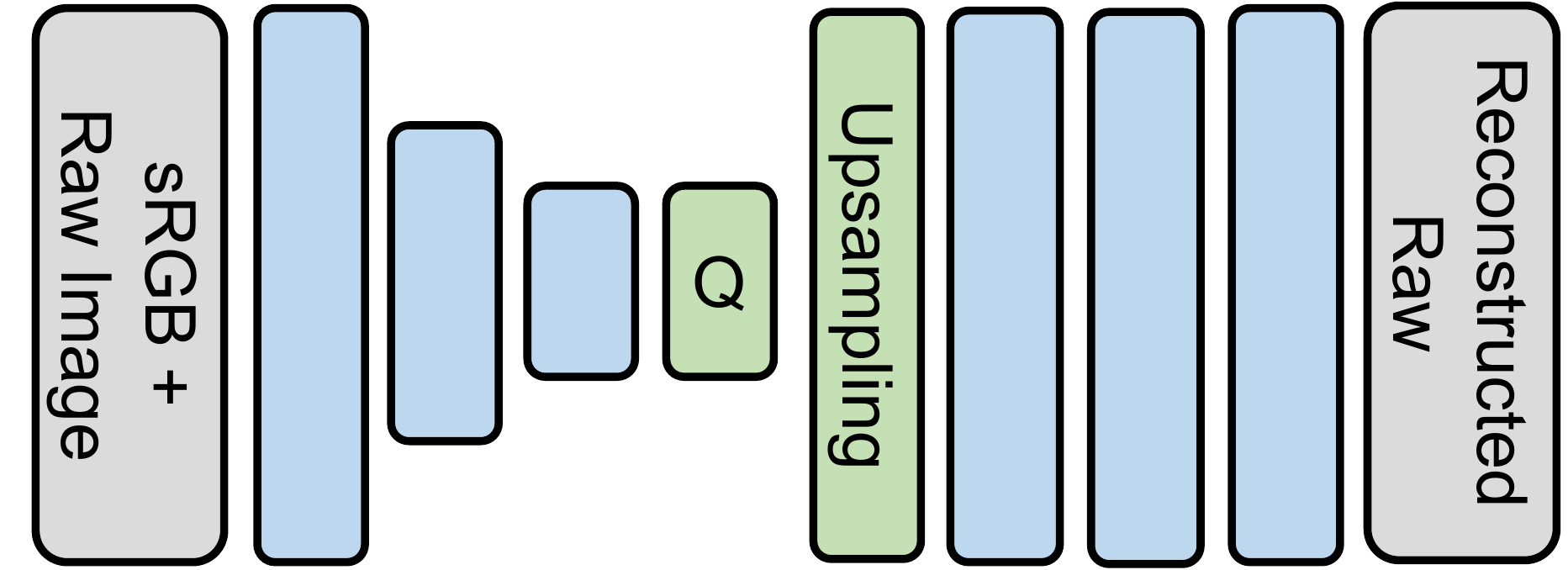}
        \caption{Asymmetric\\hybrid}
    \end{subfigure}
    \caption{Different streams of the backbone design where $Q$ is the quantization step. (a) Downsampling-oriented methods~\cite{balle2016end, li2018learning, hu2021learning} have a lower ceiling of the reconstruction quality. (b) Resolution maintained method~\cite{wang2023raw} leads to a large gap between training and evaluation. (c) The proposed fine-tuned downsampling backbone is a compromise between the gap and ceiling of fidelity. (d) The proposed asymmetric design of the backbone leads to a high ceiling of fidelity and a small gap between training and testing.}
    \label{fig:backbones}
\end{figure}

The proposed method is based on an encoder-decoder architecture, and both the analysis and synthesis transforms are implemented by deep neural networks. 
Different from the backbones of image compression methods w/o prior knowledge, we find that the distinctive properties of the metadata-based raw image reconstruction task lead to a different preference for the design of the backbone networks. Specifically, since we aim to achieve a high ceiling of the reconstruction quality, the latent feature during the synthesis transform (decoding process) can not be as small as the mainstream lossy image compression works~\cite{balle2016end, li2018learning, hu2021learning}~(Fig. \ref{fig:backbones} (a)). Therefore, our conference version proposes to utilize a backbone without downsampling to achieve a higher ceiling of fidelity~(Fig. \ref{fig:backbones} (b)). However, such a paradigm leads to a large gap between training and evaluation caused by the approximation error as shown in Fig. \ref{fig:quantization}. As a result, meticulous control of the training process is required and the performance tends to be sub-optimal. 

Therefore, a proper design of the backbone network plays a significant role in our task. Inspired by the observation that the latent features are relatively smooth, \textit{i.e.}, spatial redundancy commonly exists, only reducing the spatial resolution of the encoded latent features will not significantly diminish the representation ability of the model. The proposed asymmetric design in Fig. \ref{fig:backbones} (d) helps alleviate the gap between training and evaluation while retaining a high ceiling of the reconstruction quality.
A summary of the existing streams of the backbone design and the proposed ones are summarized in Fig.~\ref{fig:backbones}. The advantages of the proposed asymmetric hybrid backbone can be summarized as follows
\begin{enumerate}
\setlength{\itemsep}{0pt}
\setlength{\parsep}{0pt}
\setlength{\parskip}{0pt}
\item It has a relatively higher ceiling of the reconstruction quality since the input and the latent features of the decoder have the same spatial resolution as the raw image. For our specific task, a lower spatial resolution of the features from the encoder is sufficient, given the local uniformity of the mapping from sRGB to raw space.
\item The gap between training and testing caused by the approximated quantization step can be alleviated as illustrated in Fig. \ref{fig:quantization}.
\item By utilizing a more lightweight encoder and employing latent features with reduced spatial resolution, the encoding process could be significantly accelerated. Besides, lower latency of the encoding can be achieved.
\end{enumerate}

\subsection{The estimation of the likelihood}
%
The mainstream of previous works utilizes a hyper-prior model to further reduce the spatial redundancy in the latent feature $\z^1$ by introducing the new auxiliary latent code $\z^2$. To model the distribution of $\z^2$, a non-parametric fully factorized density model is adopted~\cite{balle2018variational, cheng2020learned}. 

The non-parametric-based modeling of the distribution has shown promising results in previous deep image compression works, including our previous conference version~\cite{wang2023raw}. However, further improvements can be made to our task. Specifically, the hyper-prior model's domain must be univariate and the network must be monotonic \cite{balle2018variational} due to limitations of the network design, which impose constraints on its expressive capacity. Moreover, previous works do not effectively utilize the side information, \textit{i.e.}, the sRGB image. To enhance coding efficiency, we propose a unified framework that incorporates the sRGB-guided context module to model both the latent code $\z^1$ and the hyper-prior feature $\z^2$.

Besides, compared with our conference version~\cite{wang2023raw}, other improvements in terms of the likelihood estimation are as follows:
\begin{enumerate}
\setlength{\itemsep}{0pt}
\setlength{\parsep}{0pt}
\setlength{\parskip}{0pt}
\item We propose an adaptive quantization bin width strategy that can adaptively assign different quantization precision for different elements in latent feature features based on both sRGB images and the already processed features.
\item Different approximations of the non-differentiable quantization step are used for features from different levels during training so that the gap between training and inference can be better mitigated.
\item A more accurate modeling of the order masks in the proposed sRGB-guided context model is proposed. 
\end{enumerate}
In the following subsections, we first introduce the limitations of the existing approximation strategy for the quantization step, and then the details of the improved estimation.

\subsubsection{The approximation of the quantization step} For inference purposes, all latent codes need to be quantized to discrete values since entropy coding can only handle discrete symbols. However, the quantization step is non-differentiable, making it unsuitable for direct application in training. To allow optimization, previous approaches tend to introduce a uniform noise to relax the quantization process as follows
\begin{equation}
    \tilde{z} = z + u, \quad u \sim \left[ -\frac{1}{2}\Delta, \frac{1}{2}\Delta \right], 
\end{equation}
where $\Delta$ is the quantization bin width which is usually set to $1$. The reason for using the above approximation is that the probability density function (PDF) of $\tilde{z}$ is a continuous function that interpolates the discrete probability values at integer positions~\cite{balle2016end}, \textit{i.e.},
\begin{equation}
    (q*\mathcal{U}(-\frac{1}{2}\Delta, \frac{1}{2}\Delta))(\tilde{z})= P_q(\tilde{z}),
\end{equation}
where $q$ is the Gaussian distribution parameterized by the network, $*$ is the convolution operation, and $P_q$ is the probability mass function measured on the quantized latent space.

In vanilla deep image compression tasks, a fixed quantization bin width, denoted as $\Delta=1$, performs well because the latent codes are distributed over a large range. However, in our scenario, where only a small amount of information needs to be encoded, the latent codes are typically close to or equal to zero. The aforementioned relaxation technique results in a relatively large non-zero lower bound for the bit-per-pixel loss, as demonstrated in Theorem \ref{theorem1}, which makes the training process less tractable. 
%
To address this limitation, we propose an adaptive quantization precision strategy, which is explained in detail in the following section.

\begin{figure}[t]
\begin{minipage}{\linewidth}
\begin{theorem}
Given a fixed quantization bin width $\Delta$ and a lower bound of the predicted scale $\sigma_\theta$, the lower bound of the bit-per-pixel loss during training is as follows if using the relaxation of uniform noise for any latent code $z$
\begin{equation*}
    E_{u\sim U(-\frac{1}{2}\Delta, \frac{1}{2}\Delta)} [\mathcal{R}(z+u)] \geq -\log_2 \left[\frac{2\int_0^{\Delta}c(x)\text{d}_x}{\Delta} - 1 \right]
\end{equation*}
where $c$ is the cumulative distribution function (CDF) of the distribution $\mathcal{N}(0, \sigma_\theta)$.
\label{theorem1}
\end{theorem}
\end{minipage}
\end{figure}

\subsubsection{The adaptive quantization bin width}
%
In our task, the value range of the metadata is relatively small, and therefore the gap is more obvious as shown in Fig. \ref{fig:quantization}.
Besides, a fixed quantization precision also limits the expression ability of the latent code. 
To address these challenges, different from a pre-defined set of quantization bin widths~\cite{choi2019variable}, we propose a novel sRGB-guided adaptive quantization bin width strategy that can assign different quantization precision to different elements in latent features given the information from both sRGB and the hyper-prior feature. 
Specifically, the likelihood of a univariate latent code with the predicted adaptive quantization width $\Delta_\theta$ is as follows
\begin{equation}
\begin{split}
    q(z) &= (\mathcal{N}(\mu_\theta, \sigma_\theta)*\mathcal{U}(-\frac{1}{2}\Delta_\theta ,\frac{1}{2}\Delta_\theta))(z) \\    
    &= \mathbb{E}_{u\sim U} \left[ c(z+u+\frac{1}{2}\Delta_\theta)-c(z+u-\frac{1}{2}\Delta_\theta) \right],
\end{split}    
\end{equation} 
where $U$ is the abbreviation of the uniform distribution $U\left[-\frac{1}{2}\Delta_\theta,\frac{1}{2}\Delta_\theta\right]$.
Compared with previous methods that adopt a fixed quantization bin width, the proposed method not only predicts the mean and std of the Gaussian distribution, and also predicts the quantization bin width $\Delta_\theta$ for each univariate latent code.

\subsection{The hybrid quantitation approximation strategy}
We observed that the features in the hyper-prior module cannot be effectively optimized due to the approximation error introduced by the uniform noise relaxation during training, even though the proposed adaptive quantization width mitigates the gap. To address this issue, we propose the utilization of straight-through gradients~\cite{rumelhart1986learning, theis2017lossy}, which enables coherent estimation of the bit rate between training and inference. Specifically, we apply the same rounding strategy during the forward pass, while employing a smooth approximation during the backward pass, as follows
\begin{equation}
\left\{\begin{matrix}
Q(z) = \Delta_\theta \cdot \round{\frac{z}{\Delta_\theta}}\\ 
\frac{\mathrm{d} }{\mathrm{d} z} Q(z)  := \frac{\mathrm{d} }{\mathrm{d} z} z
\end{matrix}\right.
,
\label{eq:round}
\end{equation}
where $\round{\cdot}$ is the round operation, and $\Delta_\theta$ is the quantization bin width predicted by the network. The likelihood of its feature is estimated as follows

\begin{equation}
\begin{split}
    & q(z) = P_q(Q(z)) = \int_{Q(z)-\frac{1}{2}\Delta_\theta}^{Q(z)+\frac{1}{2}\Delta_\theta} p_{u_\theta, \sigma_\theta}(z)\text{d}z \\
        &= c_\theta(Q(z)+\frac{1}{2}\Delta_\theta) - c_\theta(Q(z)-\frac{1}{2}\Delta_\theta),
\end{split}
\end{equation}
where $c_\theta$ is the CDF of the distribution $\mathcal{N}(u_\theta, \sigma_\theta)$. 

While compared with the approximation that uses added uniform noise, the straight-through approximation can obtain exactly the same result during the forward of training, its gradient is an approximation. We empirically find that only adopting it to the hyper-prior module and keeping the use of uniform noise approximation to the first level achieves the best results, \textit{i.e.}, a hybrid strategy is better.

\subsubsection{The sRGB-guided context model}
\begin{figure}
    \centering
    \begin{subfigure}{1\linewidth}
        \includegraphics[width=1\linewidth]{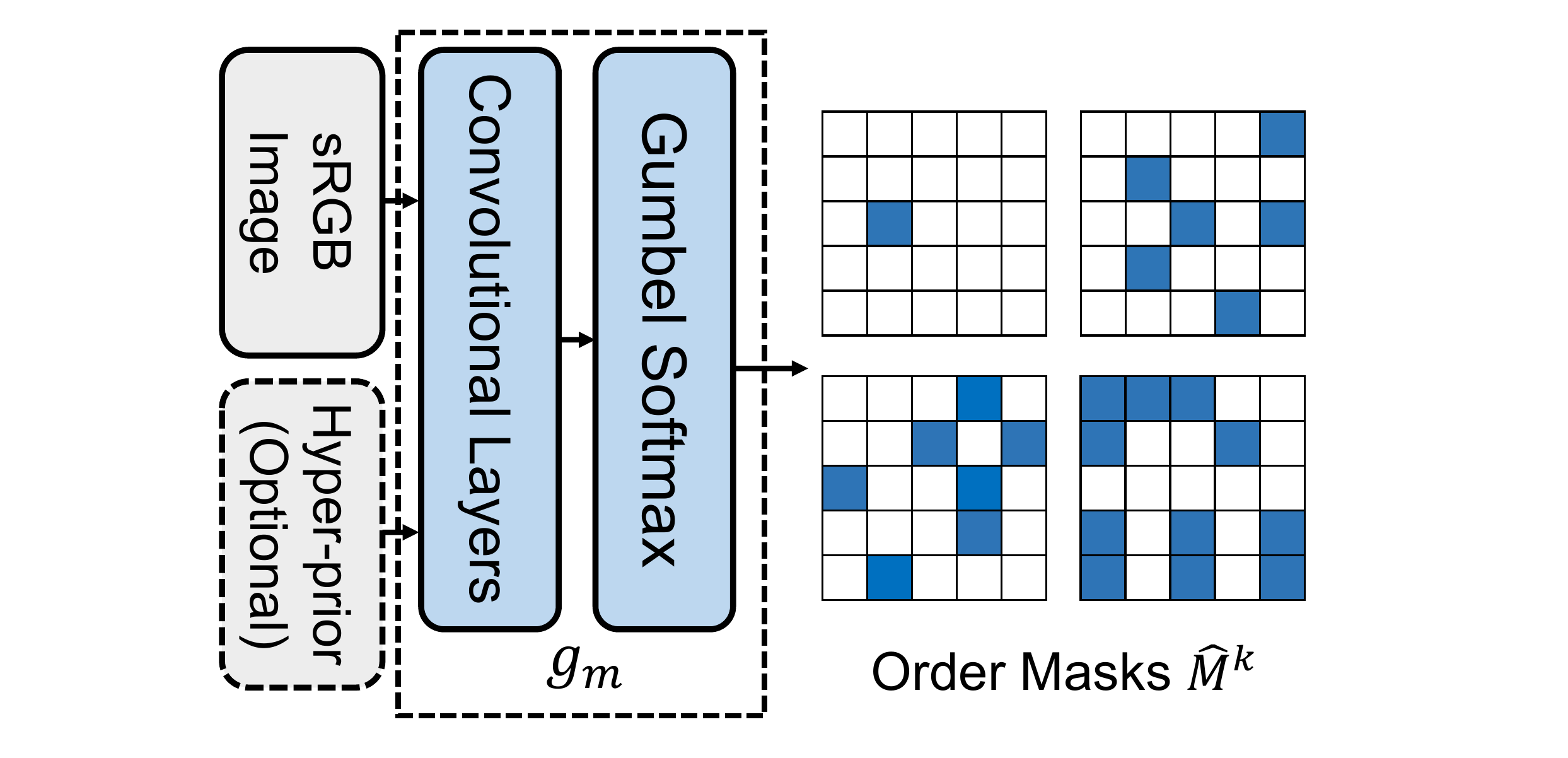}
    \caption{The order prediction module proposed in our conference work \cite{wang2023raw} where the sRGB image and hyper-prior feature are used as input.}
    \end{subfigure}
    \begin{subfigure}{1\linewidth}
        \includegraphics[width=1\linewidth]{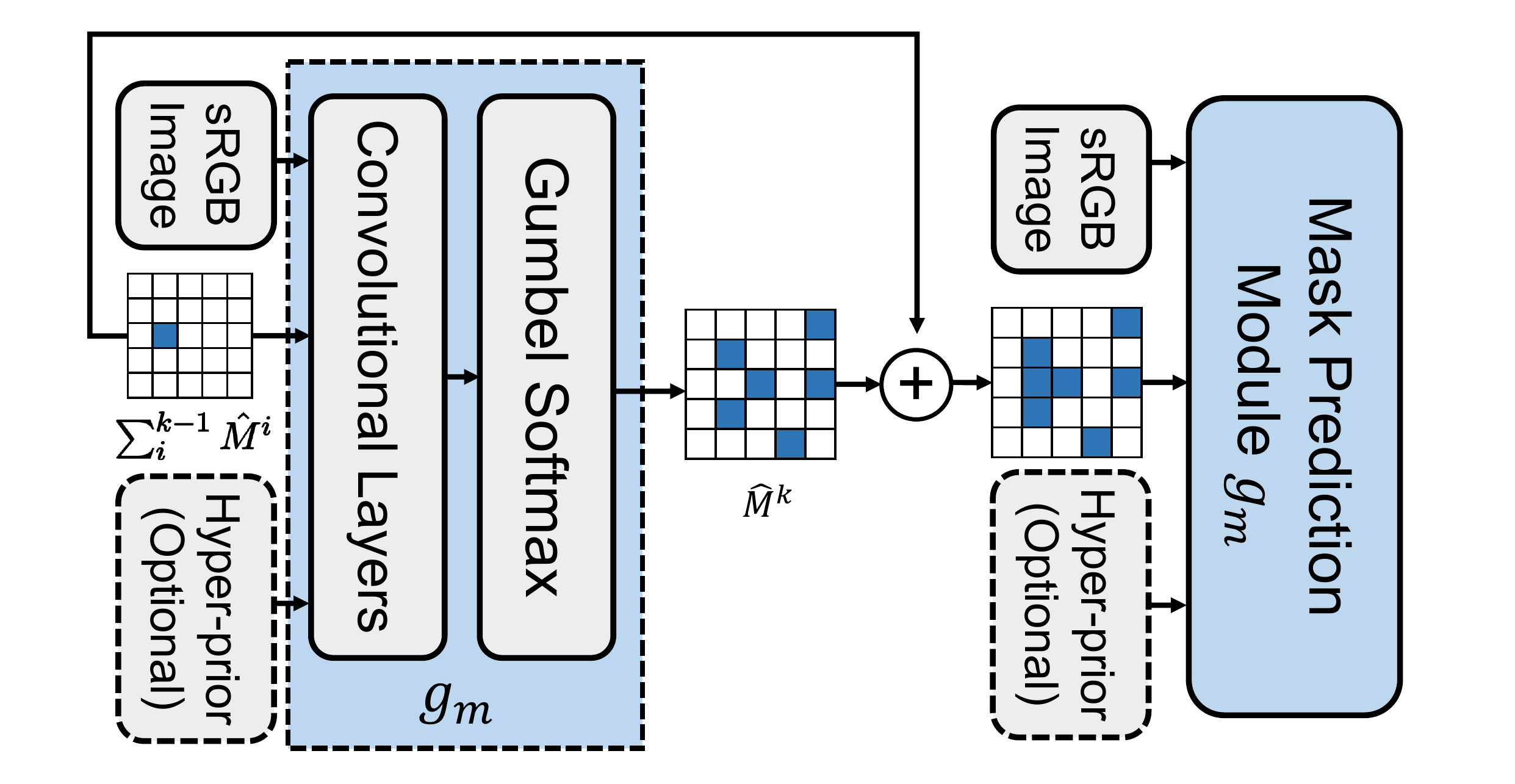}
    \caption{The improved mask prediction module can predict the mask of the to-be-processed features in the next step further based on already processed ones.}
    \end{subfigure}
    \caption{A comparison between (a) the mask prediction module in our conference version \cite{wang2023raw} and (b) the one proposed in this work.}
    \label{fig:compare_mask_pre}
\end{figure}

Previous works have demonstrated significant improvement by introducing the context model~\cite{minnen2018joint, lee2018context}, which uses the already processed pixels to predict the pixels that have not been processed yet, resulting in improved coding efficiency. However, the autoregressive model used in the context model has a serialization property that incurs a substantial computational cost, making it unsuitable for high-resolution raw image reconstruction tasks.
In order to enhance computational efficiency while retaining the benefits of the autoregressive model, we propose a novel sRGB-guided context model. 
More specifically, our proposed context model includes two parts: a learnable order prediction network $g_{m}$ as shown in Fig. \ref{fig:compare_mask_pre}, and an iterative Gaussian entropy model as shown in Fig.~\ref{fig:context}.

\noindent\textbf{The learnable order prediction module.}
Our proposed sRGB-guided context model relies on the order masks of compression/decompression as a prerequisite. 
To make sampling order masks learnable, we propose a training strategy that enables end-to-end training of the entire framework.
Different from our conference version that predicts the order only based on sRGB and the side information from the hyper-prior, the newly proposed one is additionally conditioned on the mask of already processed latent codes. By further conditioning on the accumulated mask, the proposed context model can better predict the to-be-processed latent features that can benefit most from the existing ones as shown in Fig. \ref{fig:compare_mask_pre}.
Specifically, we utilize \wh{Gumbel}-softmax \cite{jang2016categorical} to make the binary mask derivable for training as follows
\begin{equation}
    M_{i,j}^k = \frac{\exp((\log(m_{0,i,j}^k)+g^{k}_{ 0,i,j})/\tau)}{\sum_{t=0}^{1}\exp((\log(m_{t,i,j}^k)+g^{k}_{t,i,j})/\tau)},
\end{equation}
where $k \in \{0,1,...,N\}$ is the index of the sampling mask, $\mathbf{g}^k\in\mathbb{R}^{2\times h\times w}$ is a random matrix i.i.d sampled from Gumbel(0,1) distribution, $\tau$ is a temperature hyper-parameter, and $\mathbf{m}^k\in\mathbb{R}^{2\times h\times w}$ denotes unnormalized conditional log probabilities predicted by a subnetwork, \textit{i.e.}, $\mathbf{m}^{k+1}=g_m(\sum_{i=0}^{k} \hat{\mathbf{M}}^k, y, v)$ where $v$ is the hyper-prior feature.  
%
The binary mask is obtained as follows to make sure $\sum_{i=0}^{N} \hat{M}_{i,j}^k=1$, 
\begin{equation}
\hat{M}_{i,j}^k =  \left\{ \begin{matrix}
   0 & \text{ if } M_{i,j}^k < 0.5 \text{ or } \sum_{i=0}^{k-1}\hat{M}_{i,j}^k>0\\
    1 & \text{otherwise}
\end{matrix}
\right. .
\end{equation}
To make the binary operation derivable, we utilize the same straight-through gradients~\cite{rumelhart1986learning, theis2017lossy} strategy.
For inference, to make sure that we have exactly the same random vector $\mathbf{g}$ during compressing/decompressing processes, we add a registered buffer to the model to save a pre-sampled $\mathbf{g}$. 
The pre-sampled $\mathbf{g}$ is then cropped to the same size \wh{as} the $\mathbf{m}$ to generate a set of sparsely sampled $\hat{\mathbf{M}}^k$.
\begin{figure}[tbp]
    \centering
    \includegraphics[width=0.8\linewidth]{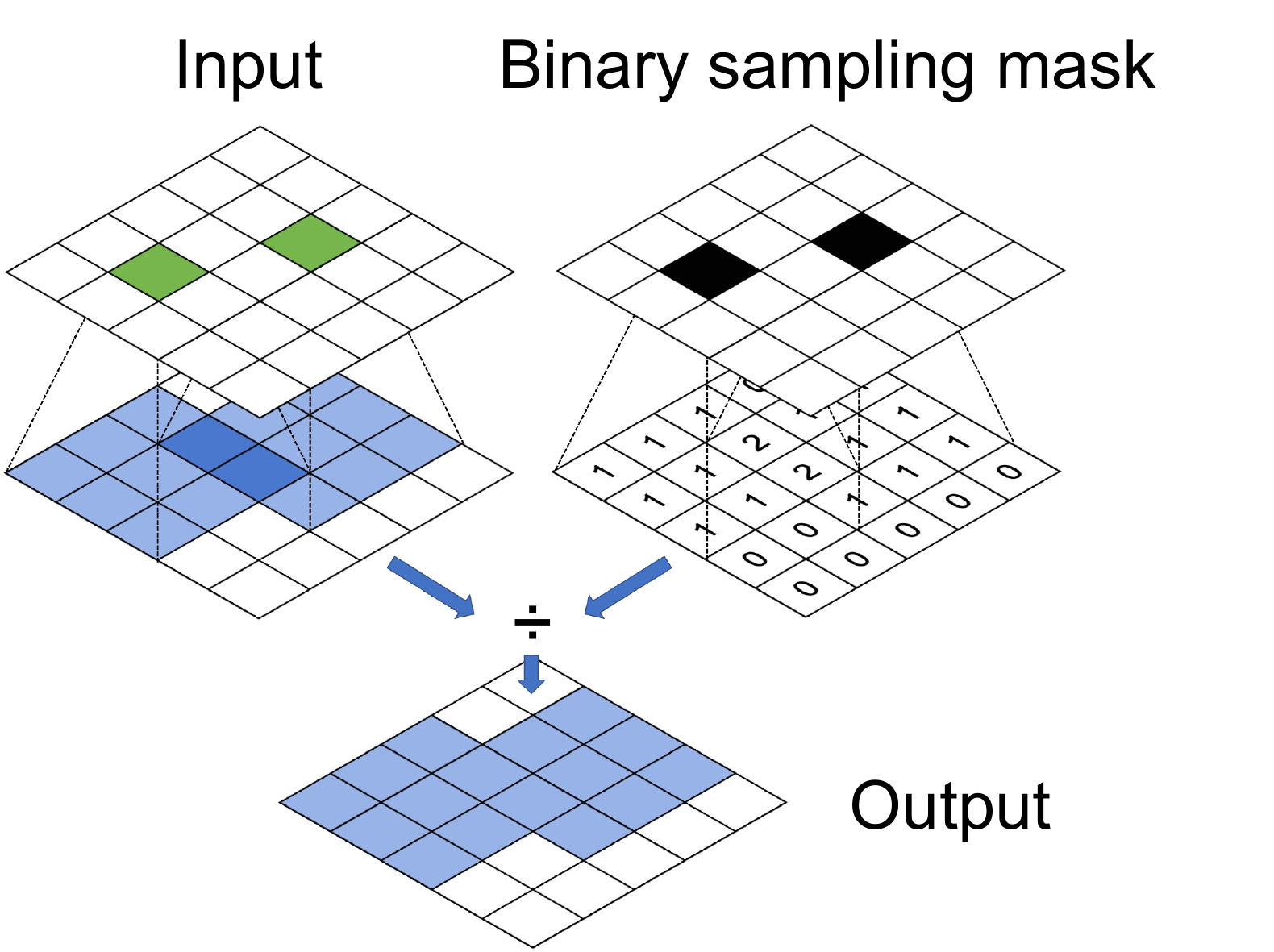}
    \vspace{0.2cm}
    \caption{The proposed masked deconvolution layer.}
    \label{fig:mask_deconv}
\end{figure}

In addition, we find that \wh{the vanilla convolutional layer cannot} well utilize the information from the randomly sparsely sampled features (can refer to the sampling mask in Fig. \ref{fig:context_vis}).
Therefore, we further propose a new masked deconvolution layer that can alleviate negative impacts from the randomness and sparsity as shown in Fig. \ref{fig:mask_deconv}. 
For an input feature $\mathbf{\hat{z}}$ and its corresponding mask $\hat{\mathbf{M}}^c=\sum_{i=0}^{k} \hat{\mathbf{M}}^k$ which records the positions of all already decoded ones, the output $\mathbf{\hat{z}}'$ is as follows
\begin{equation}
    \mathbf{\hat{z}}' = \frac{\text{Deconv}(\hat{\z})}{\max(1,\text{Deconv}_{ \mathbf{1}}(\hat{\mathbf{M}}^c))},
\end{equation}
where $\text{Deconv}$ and $\text{Conv}_{ \mathbf{1}}$ are deconvolution layers with the same kernel size and stride size of 1. Besides,  $\text{Deconv}_{\mathbf{1}}$ is a fixed layer that the weights are all one and the bias is zero. 

\noindent\textbf{The iterative Gaussian entropy model.}
After obtaining the predicted order masks, we can iteratively compress/decompress the information as shown in Fig \ref{fig:context}. Specifically, we use the information from the auxiliary latent variable $\mathbf{\hat{v}}$ and already encoded/decoded partial of $\mathbf{\hat{z}}$ to predict the distribution of the \wh{to-be-processed} part of $\mathbf{\hat{z}}$ as follows
\begin{equation}
    \bm{\mu}^{i+1}, \bm{\sigma}^{i+1}, \mathbf{\Delta}^{i+1}_\theta = g_c(g_z((\sum_{k=0}^i \hat{\mathbf{M}}^k) \odot \hat{\mathbf{z}}, \y), g_{s}(\v)),
\label{eq:predict_distri}
\end{equation}
where $\hat{\z}$ and $\v$ are from the same level so that the superscript originally for the cascade index is omitted, and is used for indicating the step-index of the context model. Besides, $\odot$ is a pixel-wise multiplication, $g_z$ is the masked decoder, and $g_c$ is the Gaussian prior module to predict the distribution of $\hat{\mathbf{z}}^{i+1}$ that are not encoded. Then, the likelihood of $\mathbf{\hat{z}}$ is formulated as
\begin{equation}
\begin{split}
    q&_{\hat{\mathbf{z}} |\hat{\v},\mathbf{y}}(\hat{z}_i|\hat{\v},\mathbf{y}) = P_q(\hat{z}_i) \\
    &= c_{\mu^{k_i}_i, \sigma^{k_i}_i}(\hat{z}_i + \frac{1}{2}\Delta_i^{k_i})  - c_{\mu^{k_i}_i, \sigma^{k_i}_i}(\hat{z}_i- \frac{1}{2}\Delta_i^{k_i}),
\end{split}
\end{equation}
where the subscript $i$ is the index of the pixel position, $k_i$ is the index of \wh{parameter} groups defined in Eq. \ref{eq:predict_distri}, $\Delta_i^{k_i}$ is the predicted element-wised quantization bin width, and $c_{\mu^{k_i}_i, \sigma^{k_i}_i}(\cdot)$ is its corresponding cumulative function. 

\begin{figure}[tbp]
    \centering
    \includegraphics[trim=0 50 0 50, clip, width=0.75\linewidth
    ]{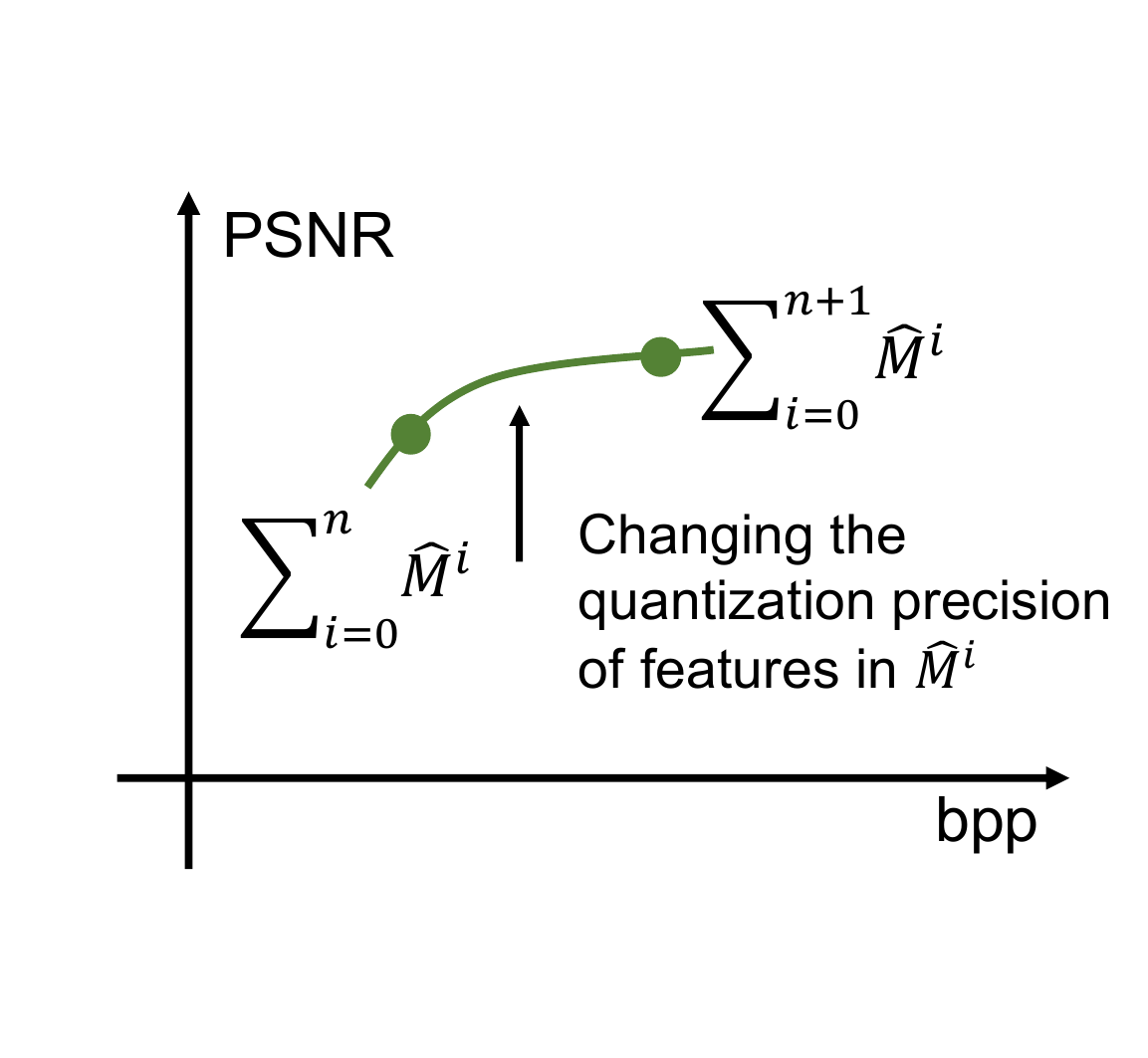}
    \caption{The proposed variable bit rates strategy for a single model with a continuous and monotonic change of the bit rate. The discrete points represent encoding/decoding latent features in a different number of sampling masks to/from the bitstreams.}
    \label{fig:var_bpp}
\end{figure}

\begin{figure}[t]
\begin{algorithm}[H]
    \caption{The variable bit rate strategy for encoding}
    \begin{minipage}{1\linewidth}
    \begin{algorithmic}[1]
    \State \textbf{Input:} to-be-encoded latent code $\z$, ordering masks $\{\hat{\M}^i | i\in \{0, 1, ..., N\}\}$, quality factor $\beta \in \mathbb{N}$ and $\gamma \in \mathbb{R}$ 
    \State \textbf{Output:} the bitstream
    \For{$k \gets 0$ to $N-1$}
    \If{$k<\beta$}
        \If{$k<\beta-1$}
            \State $\Delta=\Delta_\theta$
        \Else
            \State $\Delta=\Delta_\theta \times 2 ^ \gamma$
        \EndIf
        \State Encode $\hat{\M}^k \odot \z$ to the bitstream using the quantization bin width $\Delta$
    \EndIf   
    \EndFor
    \State \Return the bitstream
    \end{algorithmic}
    \end{minipage}
    \label{alg:encode}
\end{algorithm}
\vspace{-0.55cm}
\end{figure}

\begin{figure}[t]
\begin{algorithm}[H]
    \caption{Decoding strategy of the latent code with a variable bit rate}
    \begin{minipage}{1\linewidth}
    \begin{algorithmic}[1]
    \State \textbf{Input:} the bitstream, ordering masks $\{\hat{\M}^i | i\in \{0, 1, ..., N\}\}$, quality factor $\beta \in \mathbb{N}$, and $\gamma \in \mathbb{R}$ 
    \State \textbf{Output:} the latent code $\hat{\z}$
    \State $\hat{\z} = \textbf{0}$ for initialization
    \For{$k \gets 0$ to $N-1$}
    \If{$k<\beta$}
        \If{$k<\beta-1$}
            \State $\Delta=\Delta_\theta$
        \Else
            \State $\Delta=\Delta_\theta \times 2 ^ \gamma$
        \EndIf
        \State Decode a part of the latent code $\hat{\z}'$ from the bitstream under the quantization bin width $\Delta$
    \Else
        \State Predict $\hat{\z}'$ given the already decompressed latent feature $\hat{\z}$
    \EndIf   
    \State $\hat{\z} = \hat{\z} + \hat{\M}^k \odot \hat{\z}'$ 
    \EndFor
    \State \Return $\hat{\z}$
    \end{algorithmic}
    \end{minipage}
    \label{alg:decoding}
\end{algorithm}
\vspace{-0.55cm}
\end{figure}

\noindent\textbf{The variable bit rate strategy for a single model.} 
In this work, we propose a strategy that enables achieving variable bit rates in a wide range, where the bit rate changes monotonically based on the hyper-parameters we set. 
Intuitively, we can reduce the bit rate by lowering the quantization precision as follows
\begin{equation}
    \Delta = 2^{\gamma} \cdot \Delta_{\theta},
\end{equation}
where $\gamma$ is a hyper-parameter to control the quantization precision and $\Delta_{\theta}$ is the adaptively predicted bin width in Eq.~\ref{eq:predict_distri}. 
An example of the performance that adopts different $\gamma$ is in Fig.~\ref{fig:complicate_isp_result}.

Besides, a wider range of bit rate can be achieved by selectively foregoing the encoding of a portion of latent features as shown in Algorithm \ref{alg:encode} and \ref{alg:decoding}, \textit{i.e.}, the estimated values based on already processed ones are used instead of its exact value after quantization. 
Gradually increasing the number of used sampling masks allows us to incorporate more latent features, resulting in a higher bit rate and improved reconstruction quality. An example of using different numbers of sampling masks is illustrated in Fig.~\ref{fig:nus_curve}.
As demonstrated in Fig.~\ref{fig:var_bpp}, by combining the strategies of variable quantitation precision and dropping pixels according to the predicted order masks, we can continuously and monotonically change the bit rate in a wide range as shown in Fig.~\ref{fig:adaptive_quant}. 
\begin{figure*}[htbp]
    \centering
    \hspace{-0.4cm}
    \scalebox{0.84}{
    \subcaptionbox{Input\\(8 bit sRGB image)}{
    \includegraphics[height=0.625\linewidth, clip]{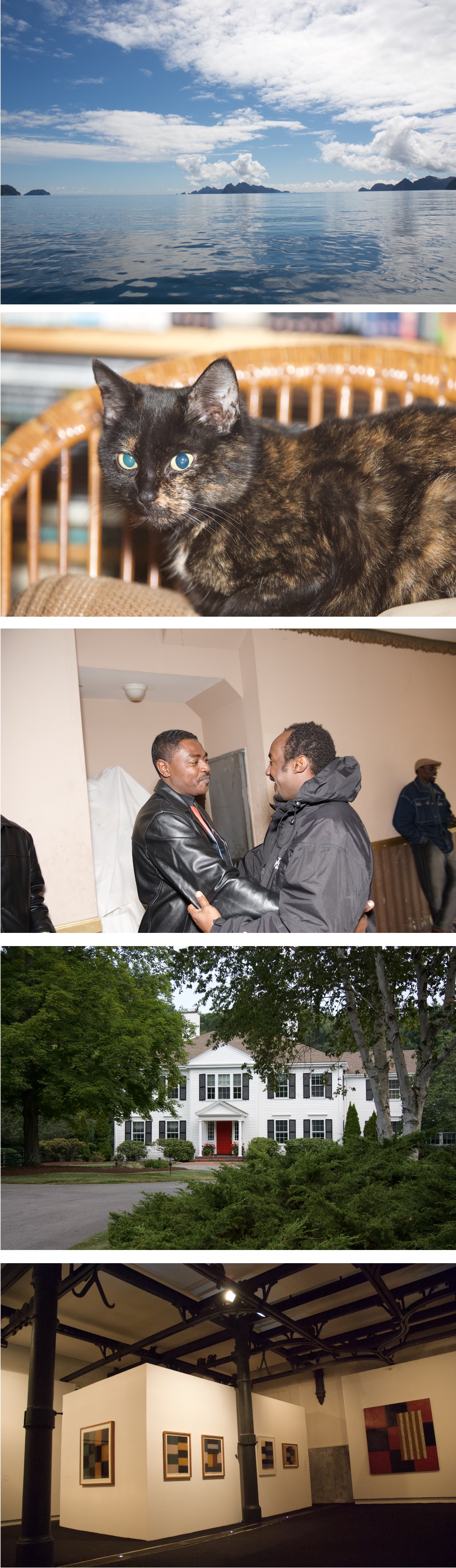}
    }
    \hspace{-0.32cm}
    \subcaptionbox{InvISP~\cite{xing2021invertible}\\(bpp: N/A)}{
    \includegraphics[height=0.625\linewidth, trim=30 0 0 0,clip ]{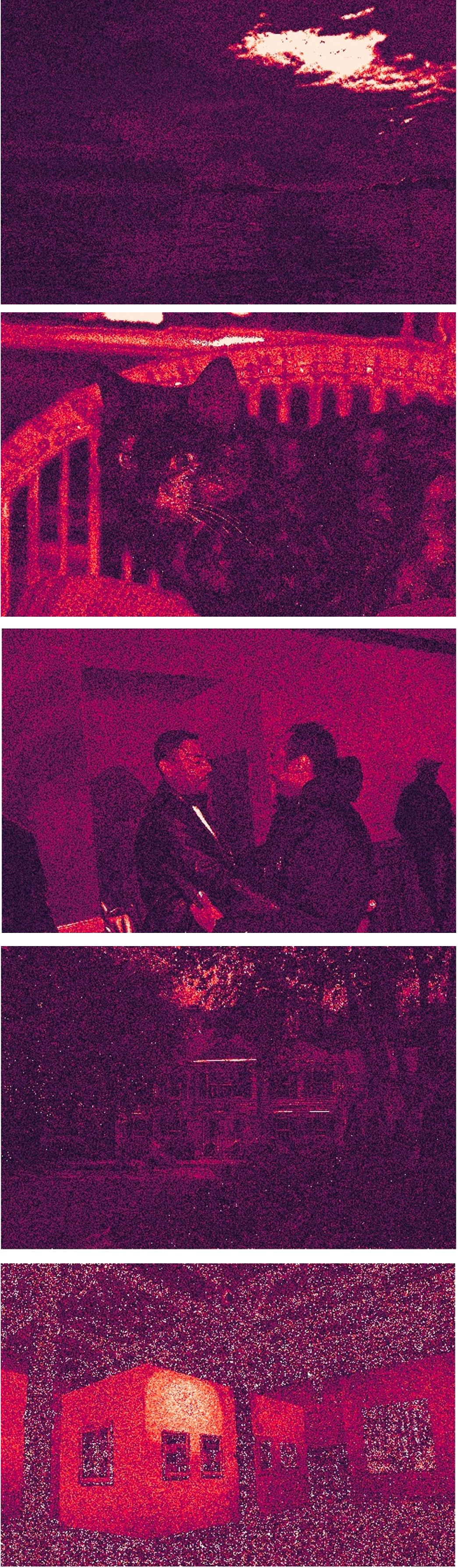}}
    \hspace{-0.3cm}
    \subcaptionbox{SAM~\cite{punnappurath2021spatially} \\(bpp: 9.52e-3)}{
    \includegraphics[height=0.625\linewidth]{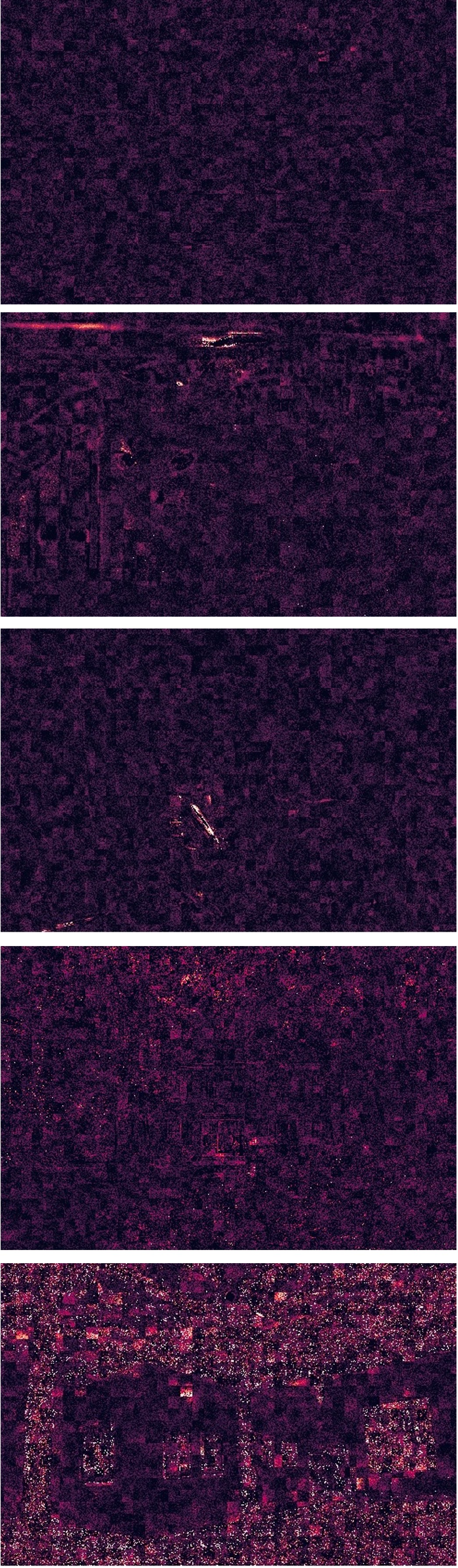}}
    \hspace{-0.3cm}
    \subcaptionbox{Nam~\textit{et al.}\cite{nam2022learning} \\(bpp: 8.44e-1)}{
    \includegraphics[height=0.625\linewidth]{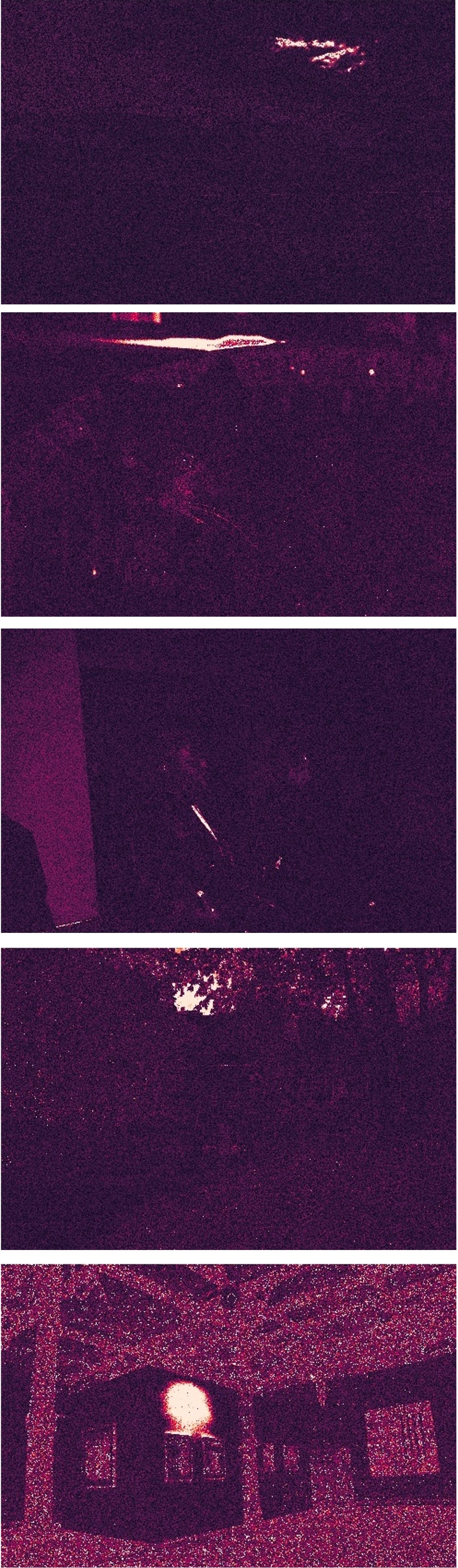}}
    \hspace{-0.3cm}
    \subcaptionbox{Ours\\(bpp: 2.92e-3)}{
    \includegraphics[height=0.625\linewidth]{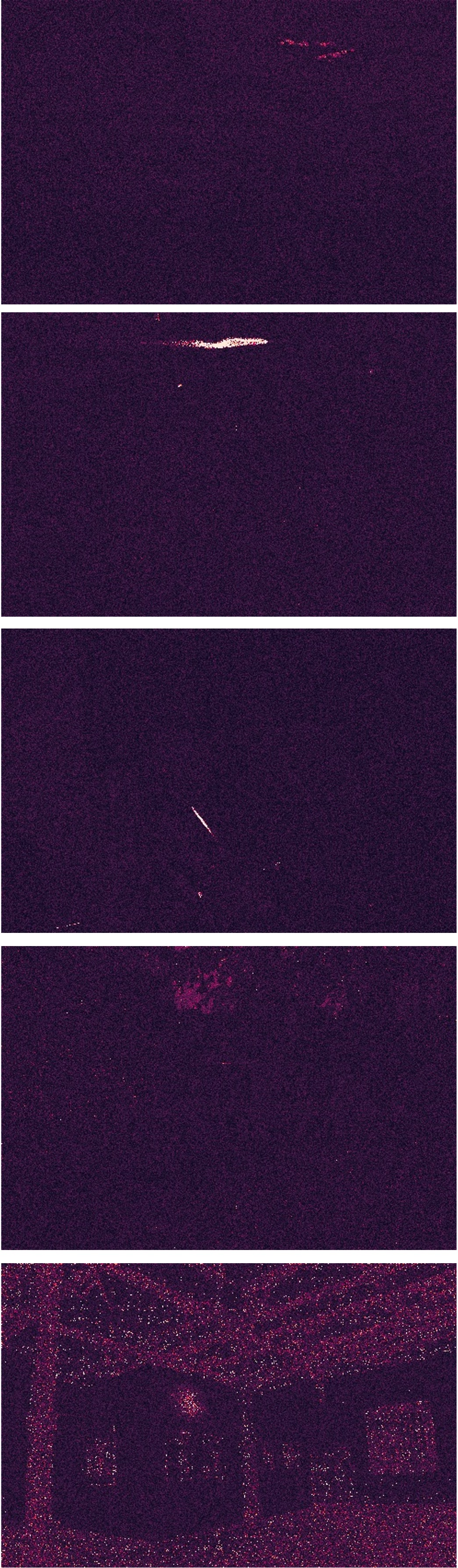}}\hspace{0.cm}
    \includegraphics[height=0.625\linewidth]{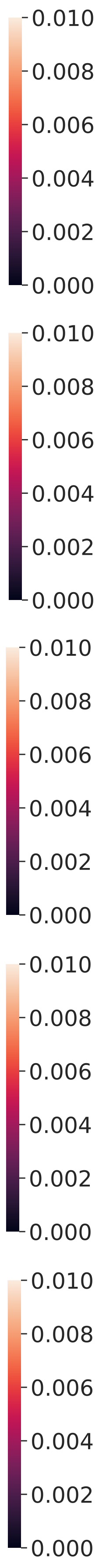}\hspace{-0.1cm}
    \subcaptionbox{Raw image\\(After gamma correction)}{
    \includegraphics[height=0.625\linewidth]{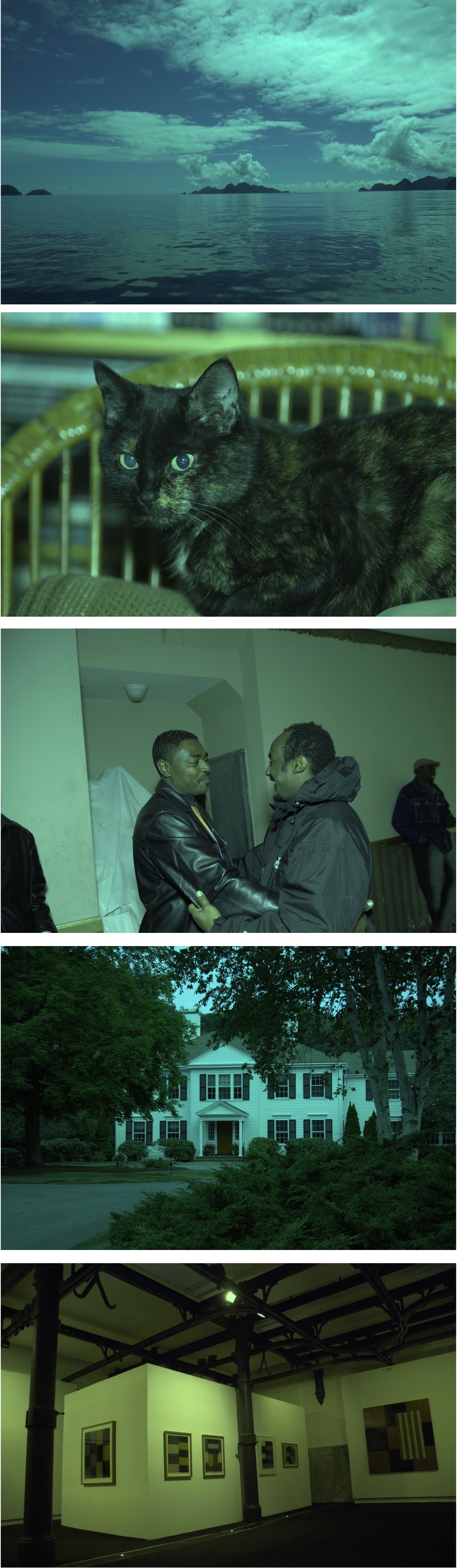}}\hspace{-0.2cm
    }}
    \caption{
    Qualitative comparison of the raw image reconstruction results on the AdobeFiveK~\cite{fivek} dataset, where the sRGB is rendered from a software ISP and the spatial resolution remains the same as the original one. We visualize the maximum value of the error among three channels on the pixel level. For better visualization, we apply gamma correction to the raw image to increase visibility. 
    }
    \label{fig:adobe}
\end{figure*}

\begin{table*}[t]
\centering
\caption{The quantitative results on NUS dataset~\cite{cheng2014illuminant} processed and released by~\cite{nam2022learning} where the sRGB is uncompressed.} 
\scalebox{0.96}{
\begin{threeparttable}
\begin{tabular}{ccccccccc}

\toprule
\multirow{2}{*}{Method} & \multirow{2}{*}{Input space} &\multirow{2}{*}{bpp $\downarrow$} & \multicolumn{2}{c}{Samsung NX2000} & \multicolumn{2}{c}{Olympus E-PL6} & \multicolumn{2}{c}{Sony SLT-A57} \\ 
                        &  &                    & PSNR $\uparrow$   & SSIM $\uparrow$  & \multicolumn{1}{c}{PSNR $\uparrow$}  & SSIM $\uparrow$  & \multicolumn{1}{c}{PSNR $\uparrow$}  & SSIM $\uparrow$ \\
\hline
SAM~\cite{punnappurath2021spatially} &  16-bit &0.7500 & 47.03 & 0.9962 & 49.35 & 0.9978 & 50.44 & 0.9982\\
Nam \textit{et al.}~\cite{nam2022learning}~\tnote{1} & 8-bit & 0.8438 & 46.24 & 0.9964 & 46.84 & 0.9966 & 47.66 & 0.9974\\
Nam \textit{et al.}~\cite{nam2022learning} & 16-bit & 0.8438 & 48.08 & 0.9968 & 50.71 & 0.9975 & 50.49 & 0.9973\\
\cite{nam2022learning} w/ fine-tuning & 8-bit & 0.8438 & 49.40 & 0.9972 & 50.37 & 0.9976 & 51.63 & 0.9983\\
\cite{nam2022learning} w/ fine-tuning & 16-bit & 0.8438 & 49.57 & 0.9975 & 51.54 & 0.9980 & 53.11 & 0.9985\\

R2LCM~\cite{wang2023raw} & 8-bit & 1.2250 & 50.78 & 0.9986 & 53.56 & 0.9933 & 53.87 & 0.9991\\
Ours & 8-bit & \textbf{0.3763} & \textbf{56.74} & \textbf{0.9996} & \textbf{59.04} & \textbf{0.9997} & \textbf{58.21} & \textbf{0.9996} \\


\bottomrule
\end{tabular}
\begin{tablenotes}
    \item[1] The reason that the bpp of SAM~\cite{punnappurath2021spatially} and Nam \textit{et al.} ~\cite{nam2022learning} is different under the same number of sampled pixels is that Nam \textit{et al.} ~\cite{nam2022learning} need extra bits to save the locations of sampled raw pixels as shown in Fig. \ref{fig:sampling}.
\end{tablenotes}
\end{threeparttable}
}
\label{tab:nus}
\end{table*}

\section{Experiments}
\subsection{Experimental settings}

\noindent \textbf{Datasets.} 
We utilize two widely-used datasets, NUS dataset \cite{cheng2014illuminant} and AdobeFiveK dataset~\cite{fivek}, to evaluate the effectiveness of our proposed methods. These datasets are all natural images collected from different scenarios and devices. 
Specifically, AdobeFiveK dataset~\cite{fivek} includes 5000 photographs taken by different photographers and devices so that it covers a wide range of scenes and lighting conditions.
%
%
%

\begin{figure}[t]
    \centering
    \includegraphics[width=0.325\linewidth]{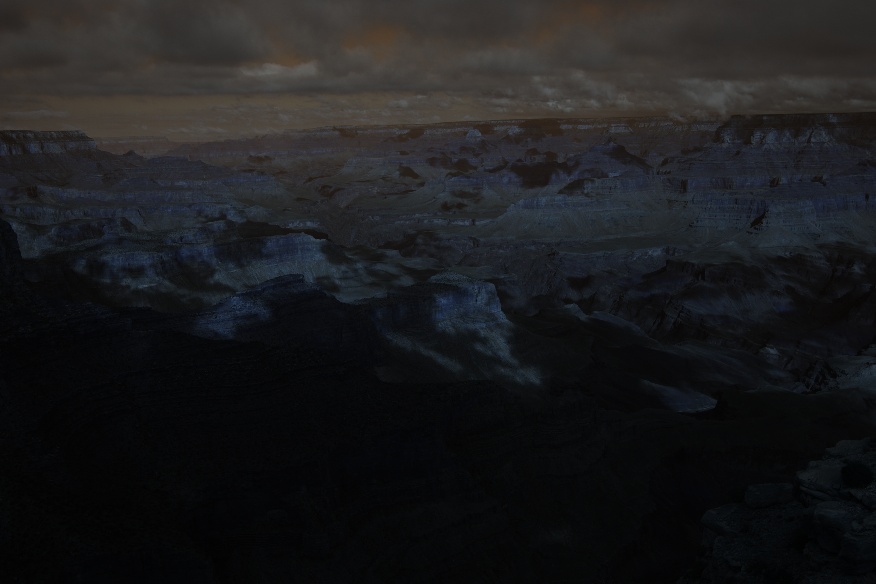}
    \includegraphics[width=0.325\linewidth]{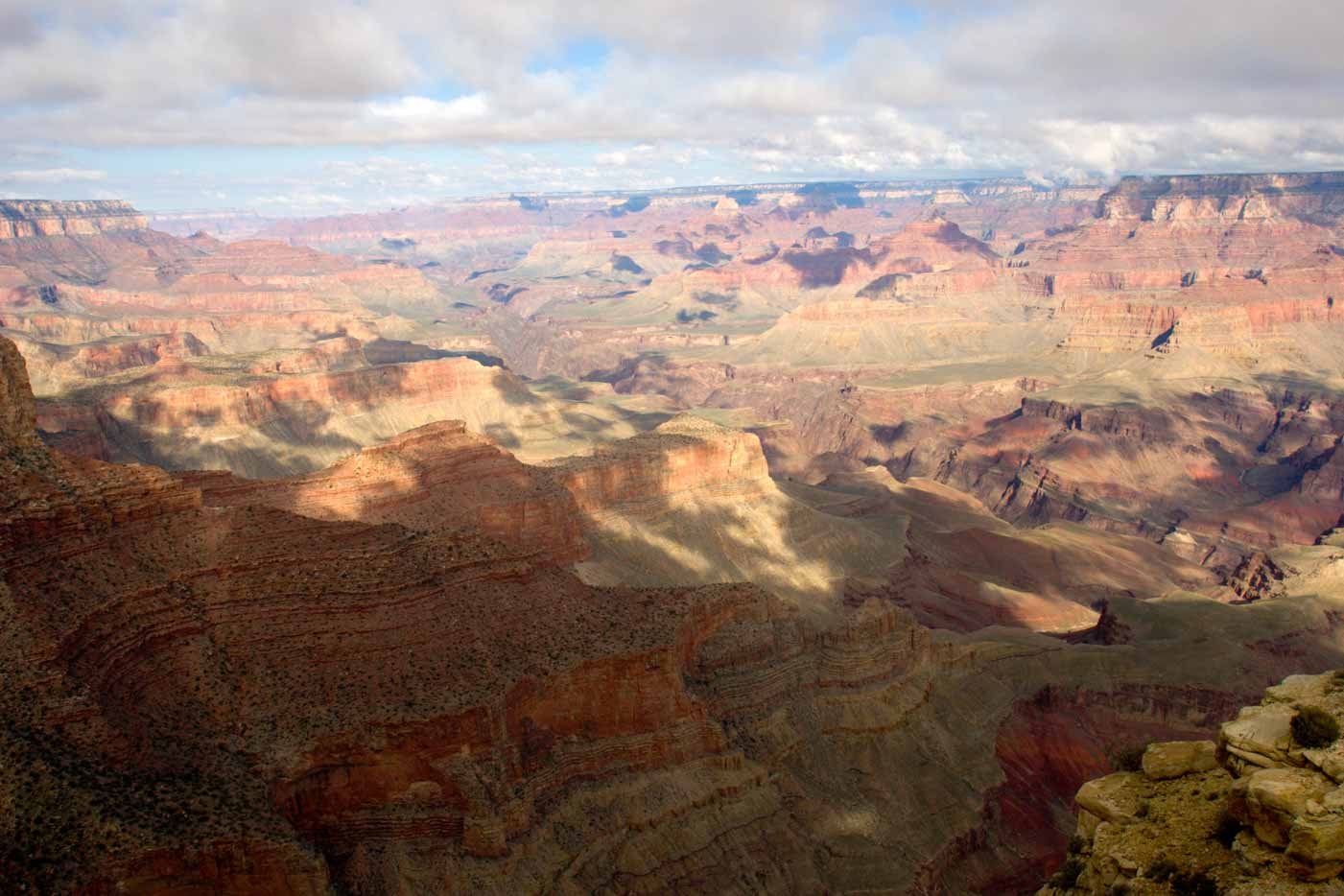}
    \includegraphics[width=0.325\linewidth]{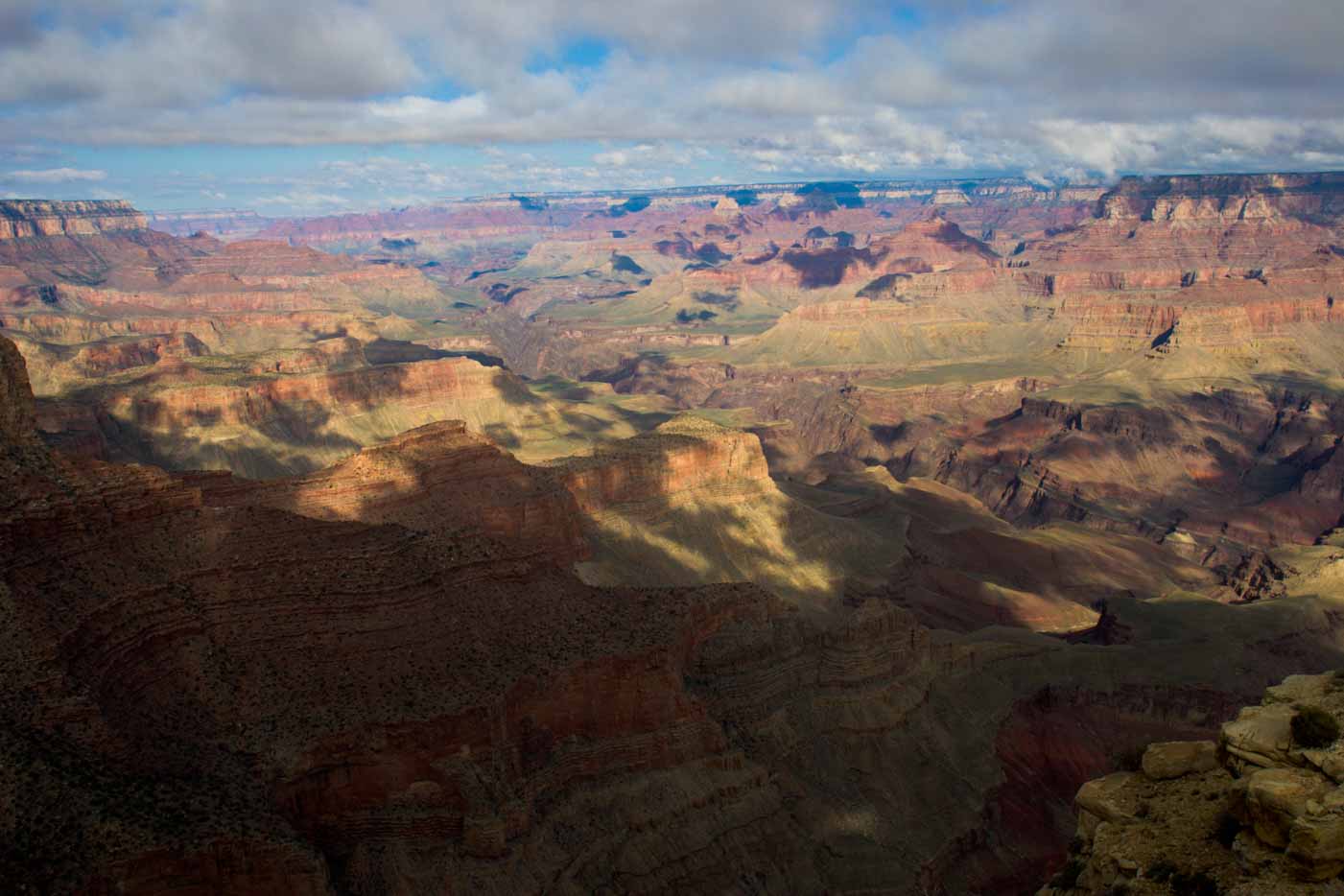}
    \includegraphics[width=0.325\linewidth]{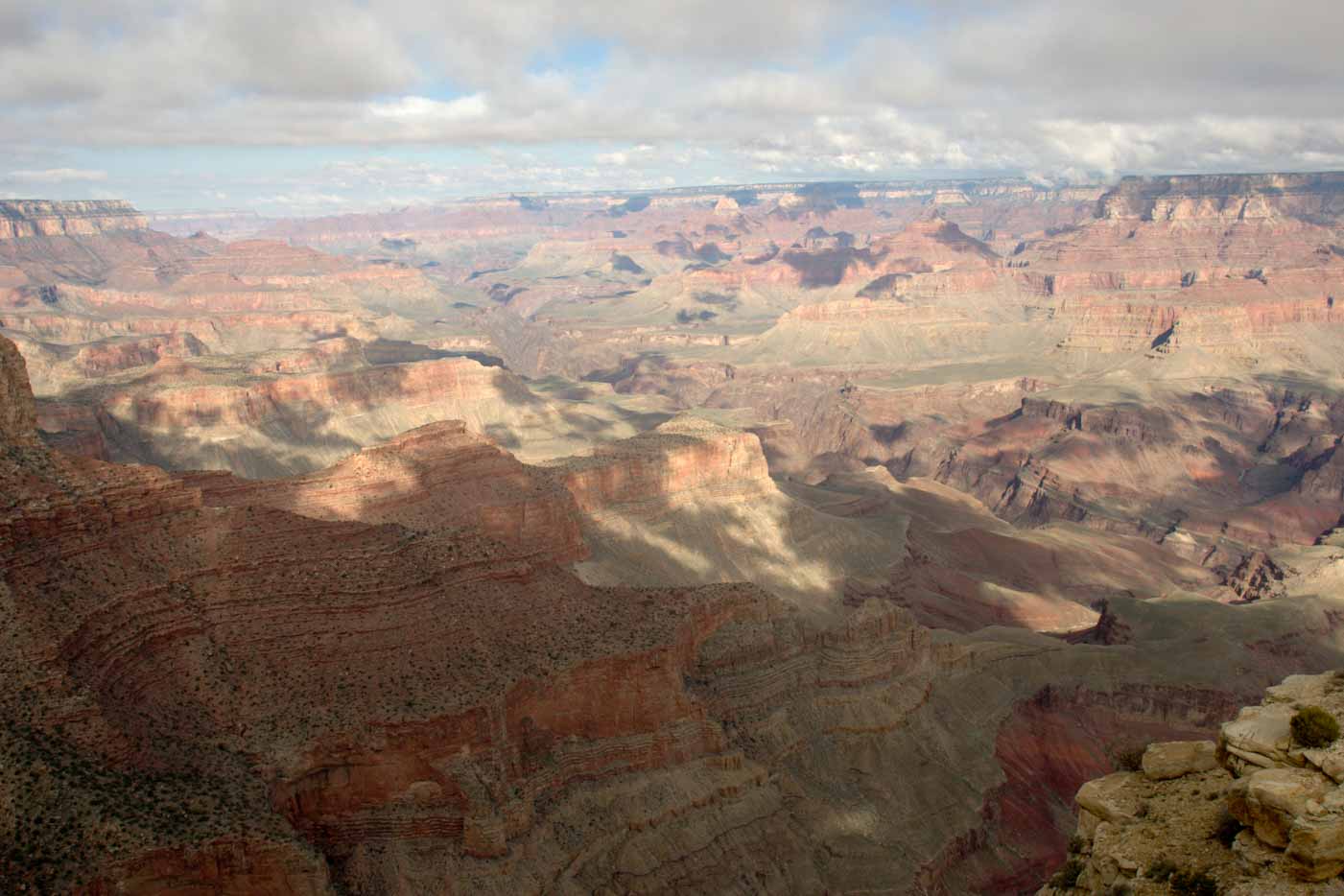}
    \includegraphics[width=0.325\linewidth]{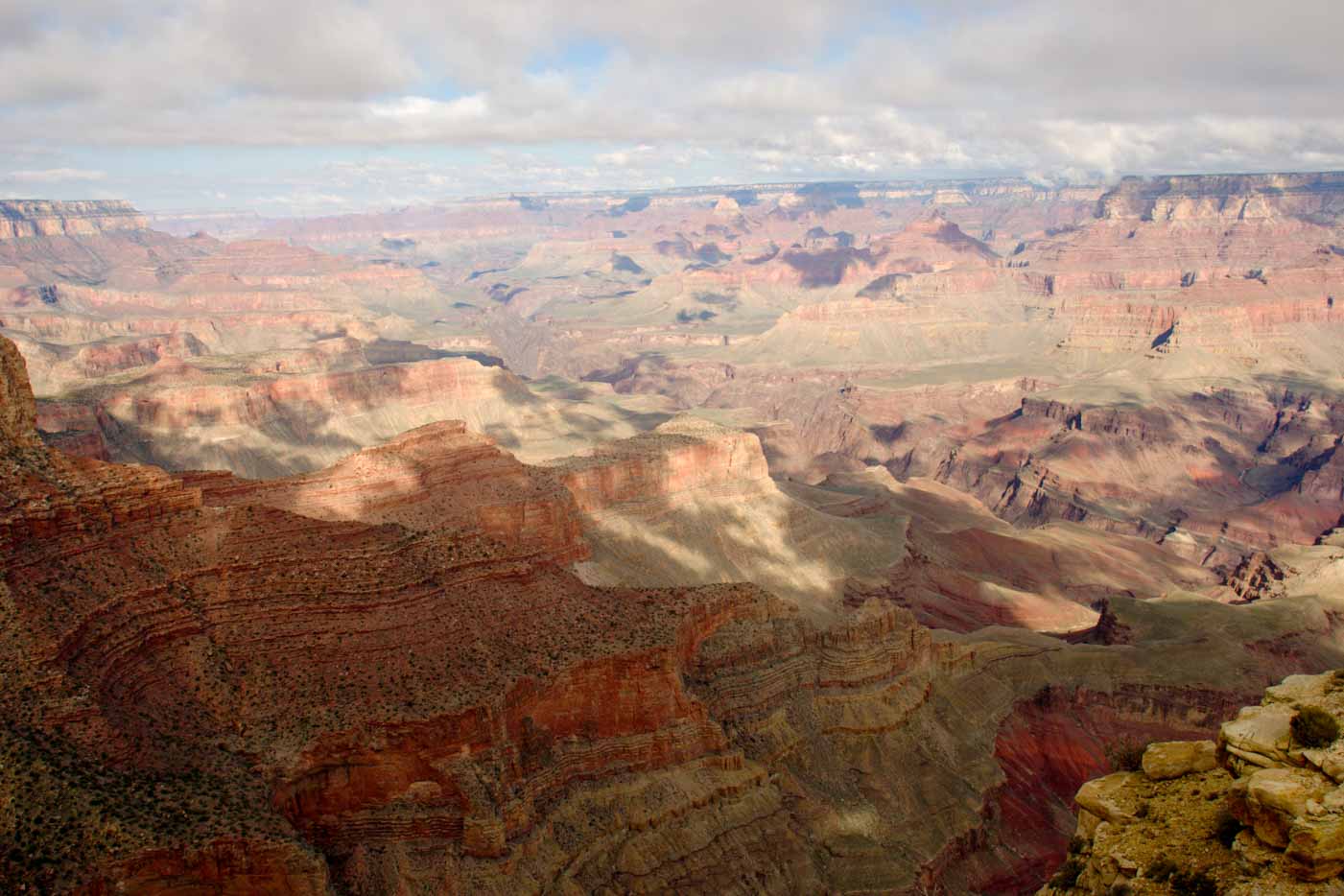}
    \includegraphics[width=0.325\linewidth]{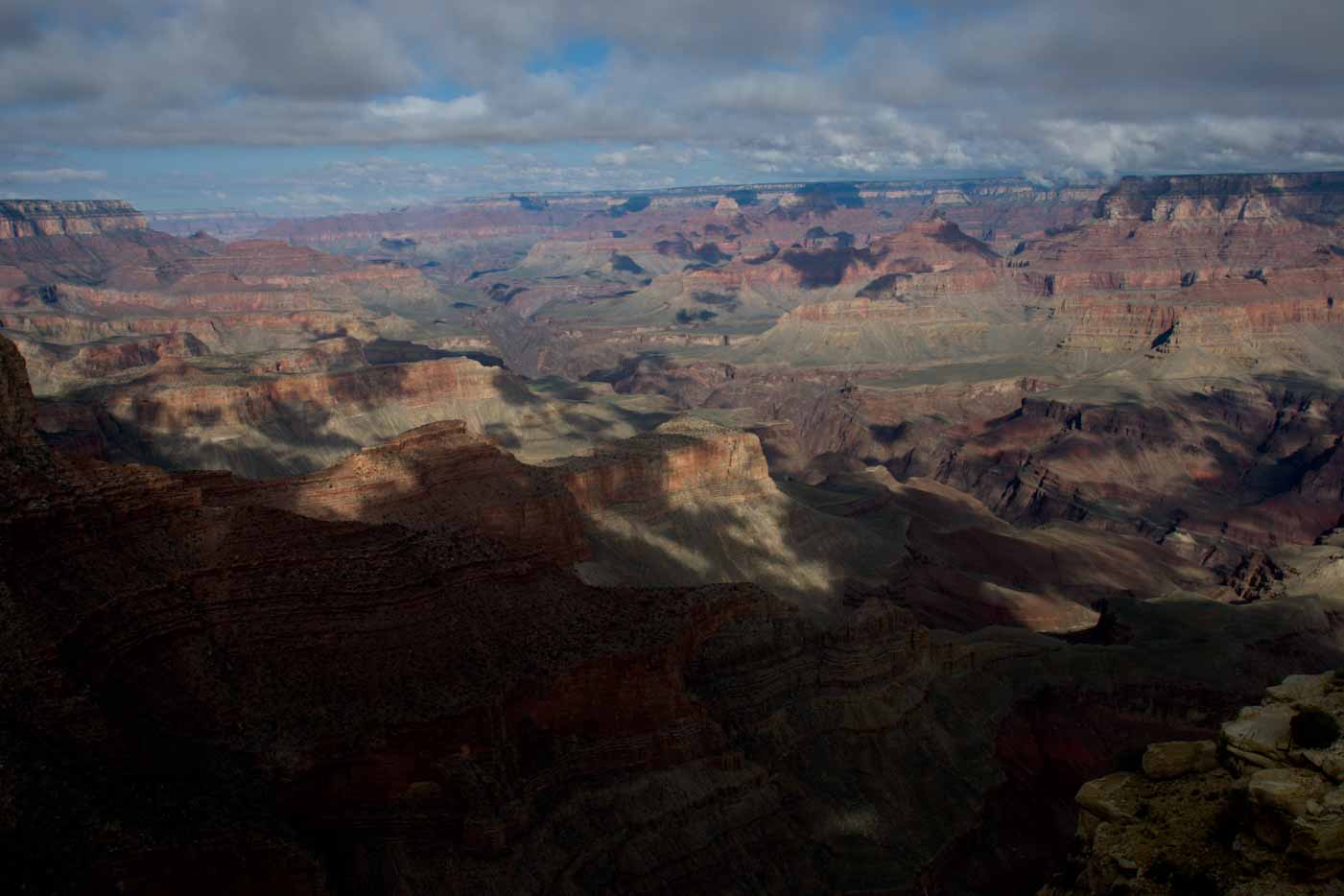} 
    \caption{Images from the AdobeFiveK dataset. The first image is the raw image and the others are from five different photographers. The images retouched by different photographers can be regarded as from diverse complicated ISPs.}
    \label{fig:adobe_retonched}
\end{figure}

\noindent \textbf{Baselines.} We compared the proposed method with several SOTA methods, including InvISP ~\cite{xing2021invertible}, SAM~\cite{punnappurath2021spatially}, and Nam \wh{\textit{et al.}}~\cite{nam2022learning}. Specifically, InvISP~\cite{xing2021invertible} is a SOTA raw image reconstruction model that \wh{utilizes} a single invertible network to learn the mapping from sRGB image to raw image and vice verse. SAM~\cite{punnappurath2021spatially} is a test-time adaptation model that \wh{saves} the uniformly sampled raw pixels as metadata. Nam \wh{\textit{et al.}} ~\cite{nam2022learning} \wh{learn} the sampling process and reconstruction process using two separate neural networks \cite{yang2020superpixel}. Besides, our conference version R2LCM~\cite{wang2023raw} is also used for comparison to further evaluate the effectiveness of the proposed improvements. 

\noindent \textbf{Implementation details.} 
For training, we utilize two sets of configurations for the settings of different resolutions.
For the setting with the original resolution, we use a batch size of $1$ and a patch size of $1024$ to reduce the I/O time. 
For the setting with down-sampled spatial resolution, the patch size and batch size are set two $256$ and $8$ respectively.  
Adam is used as the optimizer with a learning rate of 1e-4. 
We train the models for 300 epochs for AdobeFiveK dataset and 500 epochs for NUS dataset. We reduce the learning rate by a factor of 0.1 if there is no improvement in terms of the evaluation loss after every 60 epochs. For the sRGB-guided context model, we set $N=4$ for all settings. 

\begin{table}[tbp]
    \centering
    \caption{Quantitative evaluation on AdobeFiveK dataset, where the sRGB is rendered from a software ISP and the spatial resolution remains the same as the original one.}
    \begin{tabular}{cccc}
    \toprule
    Method & bpp & PSNR & SSIM \\
    \midrule
    InvISP ~\cite{xing2021invertible} & N/A & 52.69 & 0.99938  \\
    SAM~\cite{punnappurath2021spatially} & 9.566e-4 & 49.61 & 0.99874 \\
    SAM~\cite{punnappurath2021spatially} & 9.5219-3 & 54.76 & 0.99945\\
    Nam \textit{et al.} ~\cite{nam2022learning} & 8.438e-1  & 56.72 & 0.99958 \\
    \cite{wang2023raw} (w/o metadata) & N/A &53.03 & 0.99926 \\
    \\[\dimexpr-\normalbaselineskip+1pt]
    \hline
    \\[\dimexpr-\normalbaselineskip+2pt]
    R2LCM~\cite{wang2023raw} & {4.901e-4} & {58.14} & \textbf{0.99969}  \\
    Ours & \textbf{3.760e-4} & \textbf{58.44} & \textbf{0.99969} \\
    \\[\dimexpr-\normalbaselineskip+1pt]
    \hline
    \\[\dimexpr-\normalbaselineskip+2pt]
    R2LCM~\cite{wang2023raw} & {1.045-02} & {59.02} & {0.99942} \\
    Ours  & \textbf{2.916e-03} & \textbf{59.09} & \textbf{0.99973} \\
    \bottomrule 
    \end{tabular}
    \label{tab:adobe}
\end{table}

\subsection{Experimental results}
For the evaluation metrics, we utilize PSNR and SSIM \cite{wang2004image} which are widely used to evaluate the reconstruction quality with \wh{the} reference image. 
We also utilize bpp (bit per pixel) to \wh{evaluate} the coding efficiency of the model.

\begin{figure*}
    \centering
    \scalebox{0.96}{
    \begin{subfigure}{0.271\linewidth}
    \includegraphics[width=\linewidth]{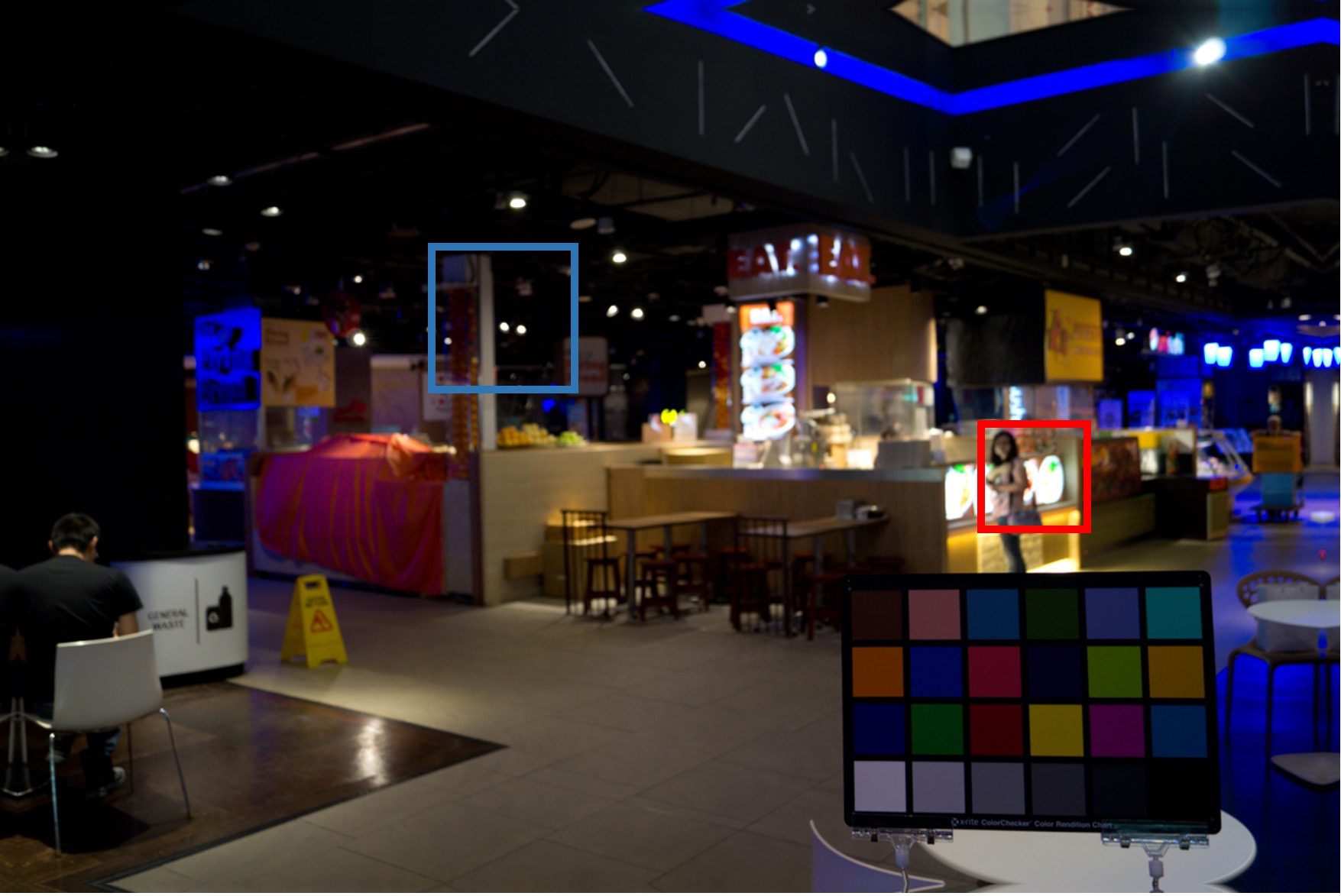}
    \caption{Input}
    \end{subfigure}
    \begin{subfigure}{0.18\linewidth}
    \includegraphics[width=\linewidth]{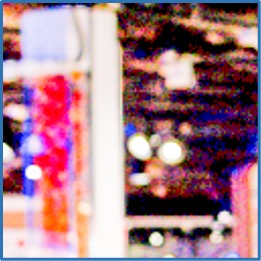}
    \caption{sRGB}
    \end{subfigure}
    \begin{subfigure}{0.18\linewidth}
    \includegraphics[width=\linewidth]{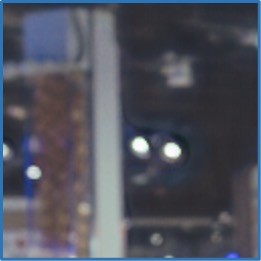}
    \caption{Nam et al. \cite{nam2022learning}}
    \end{subfigure}
    \begin{subfigure}{0.18\linewidth}
    \includegraphics[width=\linewidth]{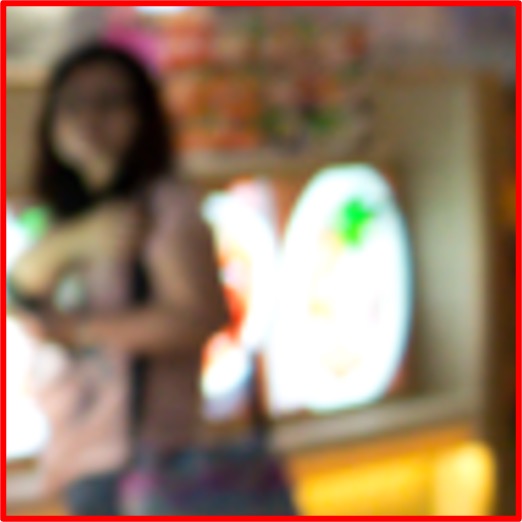}
    \caption{sRGB}
    \end{subfigure}
    \begin{subfigure}{0.18\linewidth}
    \includegraphics[width=\linewidth]{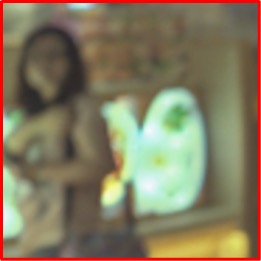}
    \caption{Nam et al.~\cite{nam2022learning}}
    \end{subfigure}
    }
    \scalebox{0.96}{
    \begin{subfigure}{0.271\linewidth}
    \includegraphics[width=\linewidth]{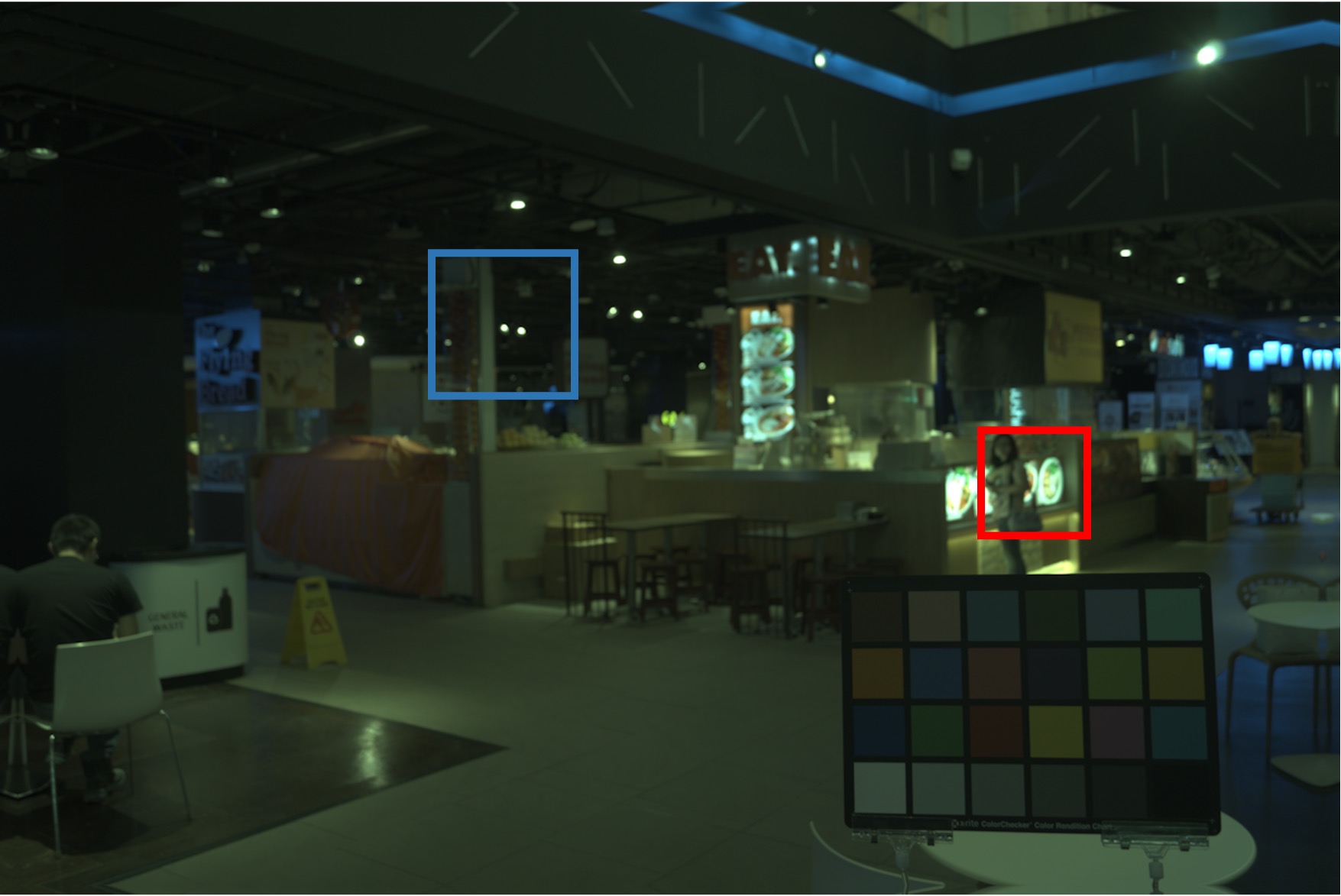}
    \caption{Input}
    \end{subfigure}
    \begin{subfigure}{0.18\linewidth}
    \includegraphics[width=\linewidth]{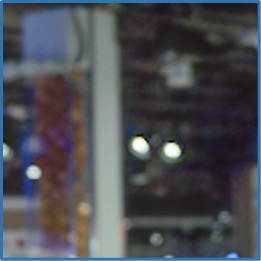}
    \caption{Ours}
    \end{subfigure}
    \begin{subfigure}{0.18\linewidth}
    \includegraphics[width=\linewidth]{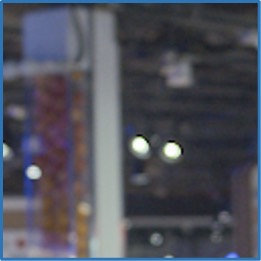}
    \caption{Ground truth}
    \end{subfigure}
    \begin{subfigure}{0.18\linewidth}
    \includegraphics[width=\linewidth]{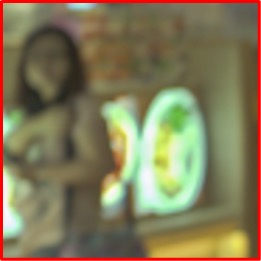}
    \caption{Ours}
    \end{subfigure}
    \begin{subfigure}{0.18\linewidth}
    \includegraphics[width=\linewidth]{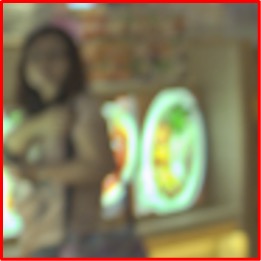}
    \caption{Ground truth}
    \end{subfigure}
    }
    \caption{Qualitative comparison between the reconstruction quality of the shadow (in the blue bounding box) and highlight areas (in red). To ensure fair comparison and improve perceptual visibility, all results from the raw space were subjected to the same post-processing process. Additionally, the brightness of sRGB zoom-in patches are fine-tuned to better display details in it. 
    Highlights and shadows are challenging areas in our task since the severe information loss by range clipping and small reconstruction loss during training.
    The proposed method can better preserve the information using smaller metadata (bpp: 0.3219 for this image) than the previous SOTA method~\cite{nam2022learning} (bpp: 0.8438).}
    \label{fig:my_label}
\end{figure*}

\begin{figure*}[htbp]
    \centering
    \hspace{-0.4cm}
    \scalebox{0.86}{
    \subcaptionbox{Input\\(8 bit sRGB image)}{
    \includegraphics[height=1\linewidth, clip]{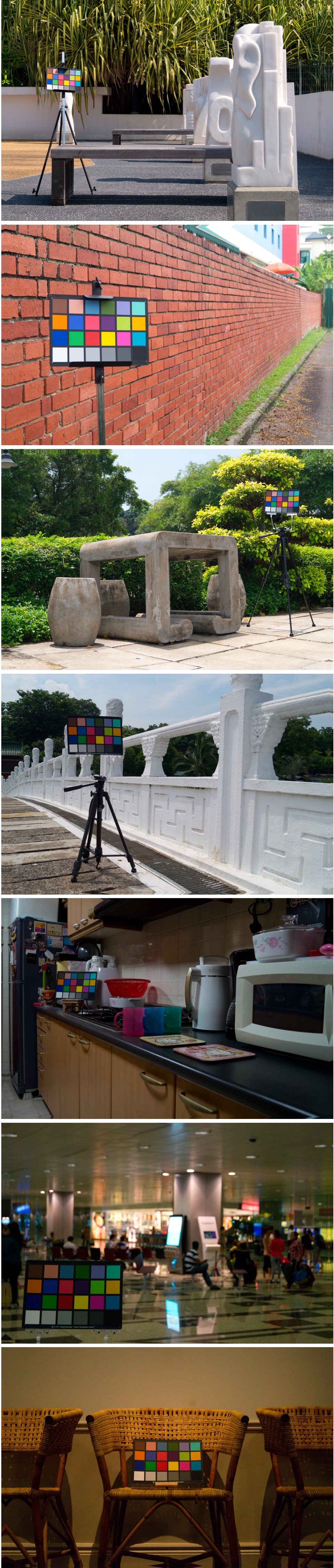}
    }
    \hspace{-0.3cm}
    \subcaptionbox{Nam~\textit{et al.}\cite{nam2022learning} \\(bpp: 0.844)}{
    \includegraphics[height=1\linewidth]{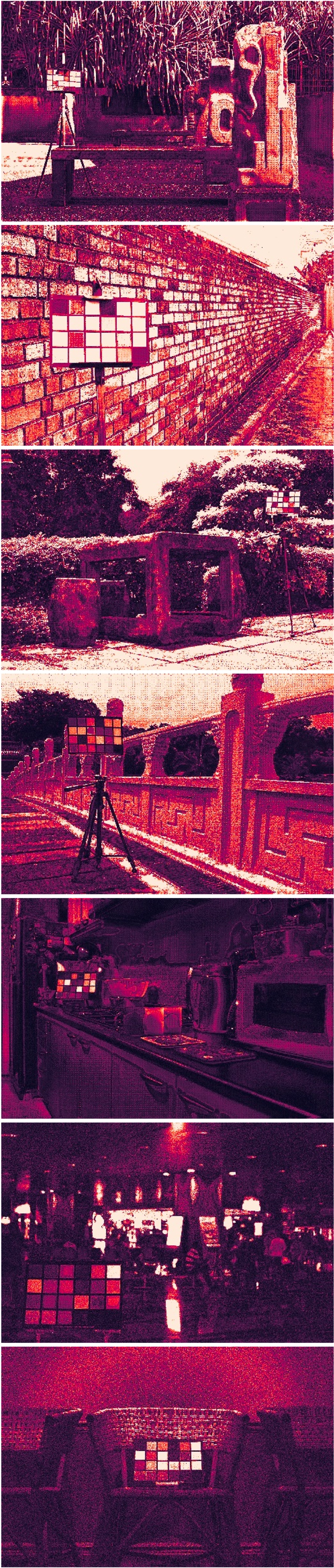}}
    \hspace{-0.25cm}
    \subcaptionbox{R2LCM~\cite{wang2023raw} \\(bpp: 1.261)}{
    \includegraphics[height=1\linewidth]{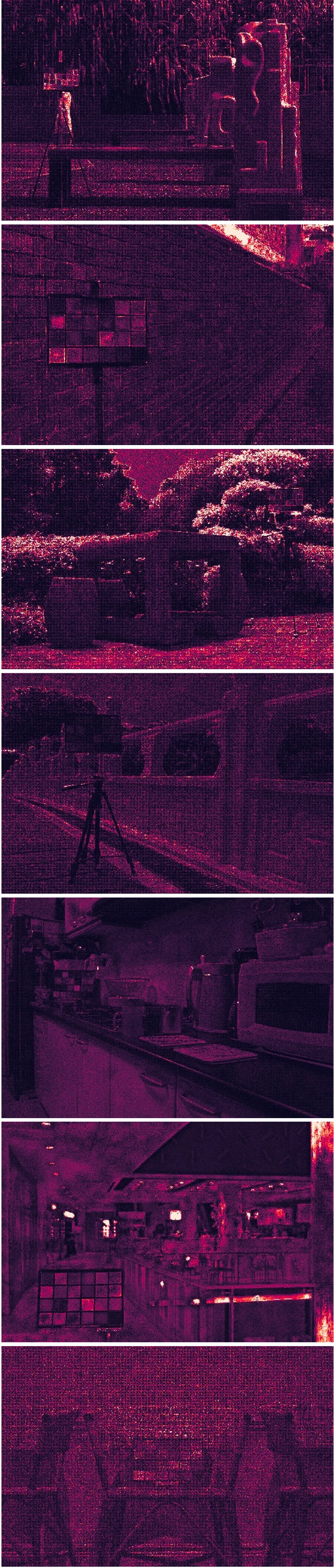}}
    \hspace{-0.25cm}
    \subcaptionbox{Ours\\(bpp: 0.262)}{
    \includegraphics[height=1\linewidth]{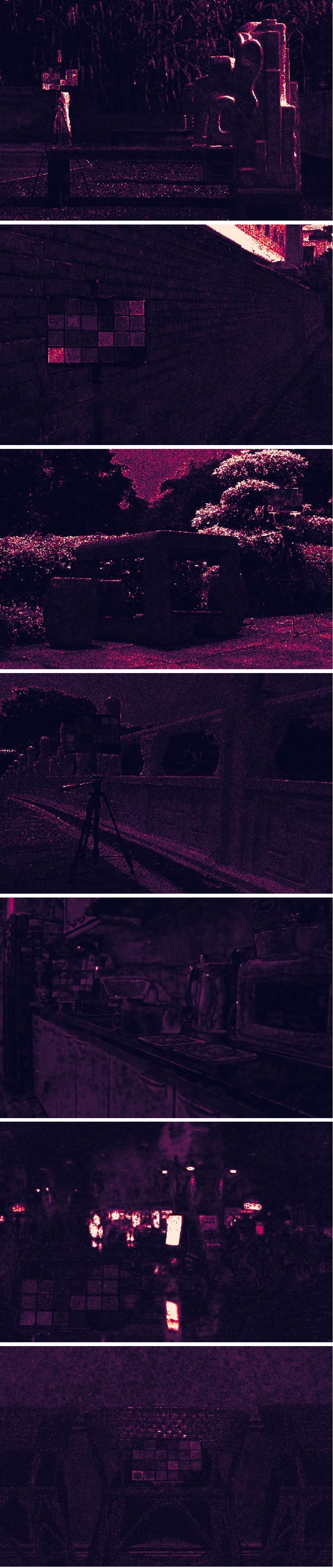}}\hspace{0.cm}
    \includegraphics[height=1\linewidth]{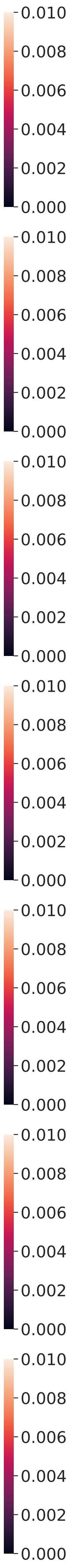}\hspace{-0.1cm}
    \subcaptionbox{Raw image\\(After gamma correction)}{
    \includegraphics[height=1\linewidth]{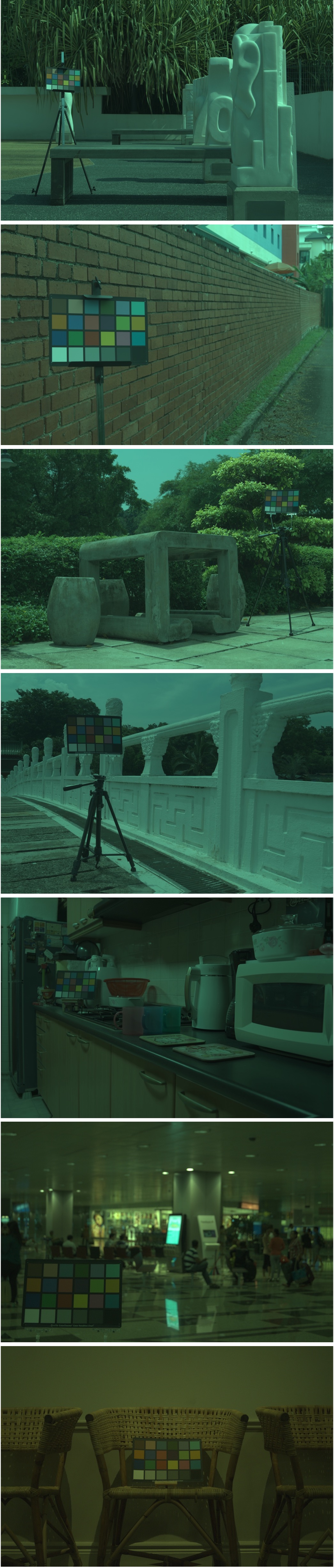}}\hspace{-0.2cm
    }}
    \caption{
    Qualitative comparison of the raw image reconstruction results on the NUS dataset~\cite{cheng2014illuminant} processed and released by~\cite{nam2022learning} where the sRGB images are uncompressed. We visualize the maximum value of the error among three channels on the pixel level. For better visualization, we apply gamma correction to the raw image to increase visibility. 
    }
    \label{fig:nus_main}
\end{figure*}

\subsubsection{Results on uncompressed sRGB data.}
\noindent\textbf{Results on AdobeFiveK with software ISP.}
Following previous work \cite{nam2022learning, punnappurath2021spatially, wang2023raw}, we use the raw image after demosaic and render sRGB images using a software ISP. 
The quantitative evaluation results are shown in Table \ref{tab:adobe}.
As we can see in the table, raw image reconstruction models with metadata can achieve better performance than SOTA raw image reconstruction models without metadata~\cite{xing2021invertible}. Besides, compared with previous metadata-based SOTA methods \cite{punnappurath2021spatially, nam2022learning}, our method achieves better reconstruction quality with lower storage overhead.
In addition, for the deep learning-based method, the model without using the metadata exhibits worse performance than that w/ metadata using the same backbone, demonstrating the necessity of the saved metadata.
Besides, compared with our conference version R2LCM~\cite{wang2023raw}, the model in this work has better reconstruction quality with lower storage overhead among different compression rates.
Visual comparisons can be found in Fig.~\ref{fig:adobe}.

\noindent \textbf{Results on AdobeFiveK with retouched sRGB images.}
\begin{figure}[tbp]
    \centering
    \includegraphics[height=2.9cm, trim=15 0 15 0, clip]{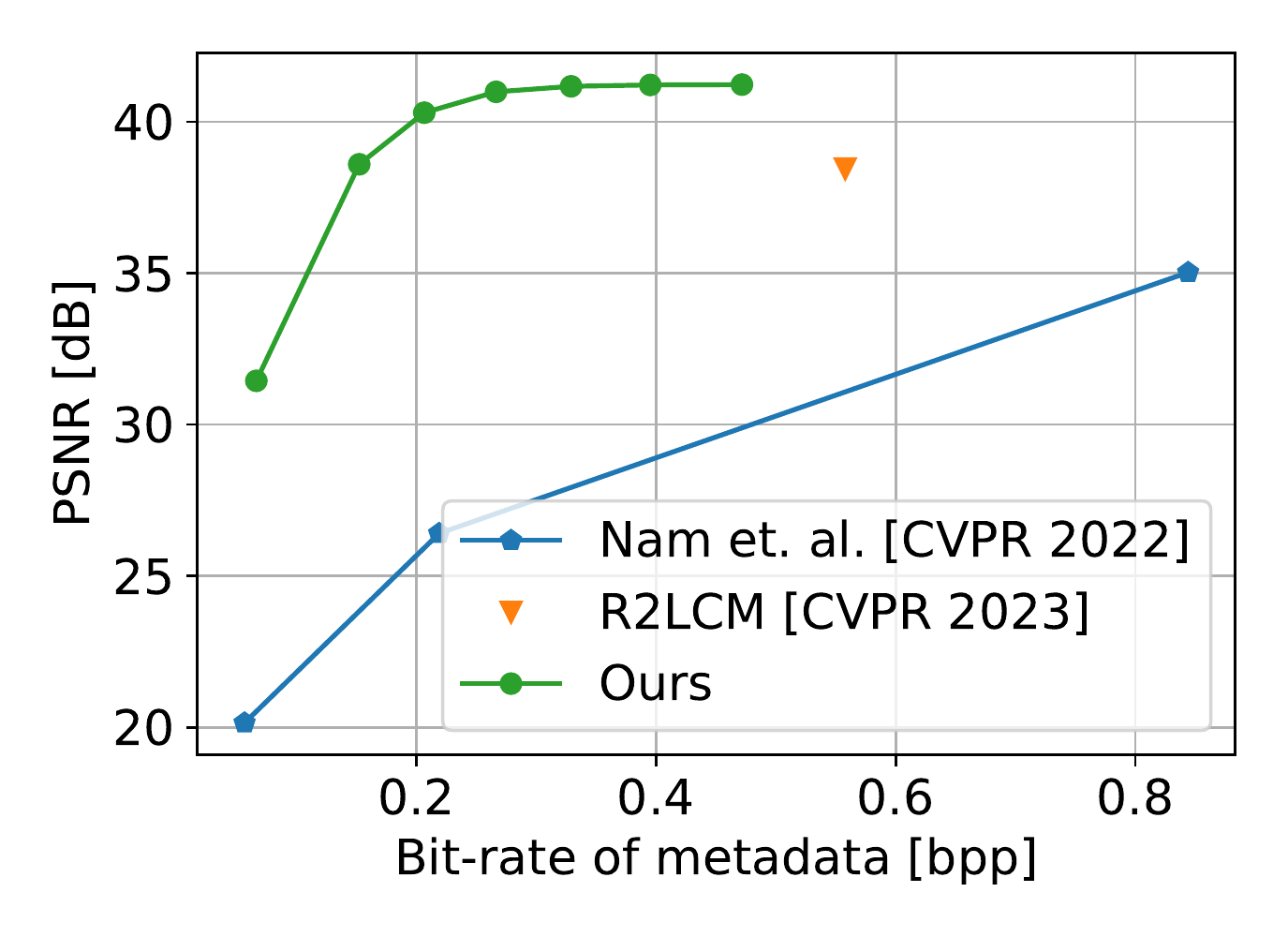}
    \includegraphics[height=2.9cm, trim=15 0 15 0, clip]{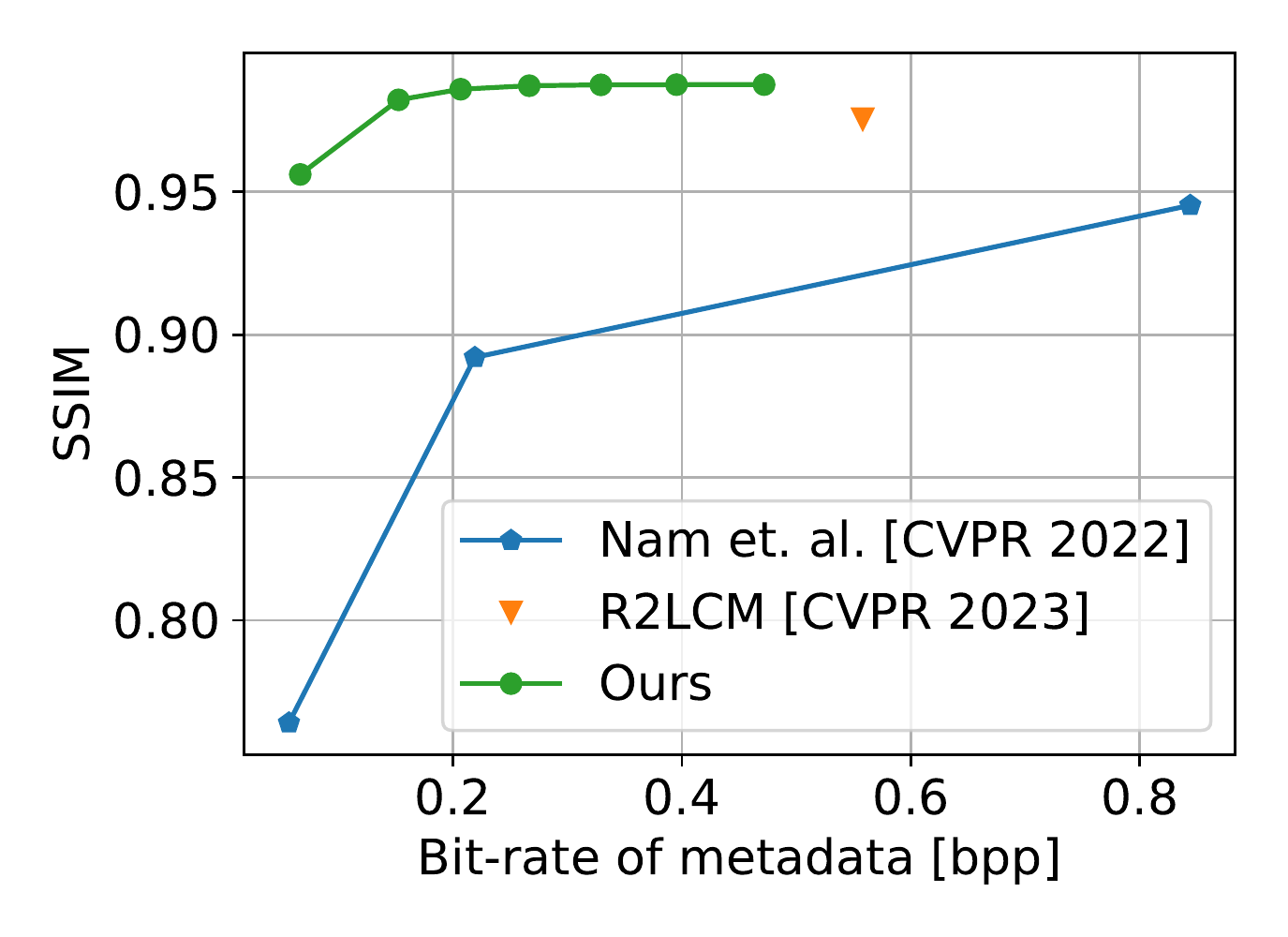}
    \caption{Rate-distoration curves over AdobeFiveK dataset with retouched sRGB images. We compare the proposed method with the SOTA methods Nam~\textit{et. al.} ~\cite{nam2022learning} w/ online fine-tuning and our conference work R2LCM~\cite{wang2023raw}. For the proposed method, $\gamma$ is set to $[0,1,2,3,4,5,7.5]$ to achieve different bit rates.}
    \label{fig:complicate_isp_result}
\end{figure}
Furthermore, to assess the robustness of the proposed method against more complex Image Signal Processors (ISPs), we employ sRGB images retouched by five professional photographers instead of using a simplified software ISP, as illustrated in Figure \ref{fig:adobe_retonched}.
To ensure a fair comparison with the previous state-of-the-art (SOTA) method~\cite{nam2022learning}, we apply the same downsampling factor of 4 to the dataset.
For each raw image, we randomly select from the five retouched sRGB images. Besides, the image pairs with severe misalignment are excluded from the organized dataset.
This results in a training set of 4765 pairs and a testing set of 99 pairs. We evaluate the proposed method alongside the previous SOTA method~\cite{nam2022learning} and our conference work R2LCM~\cite{wang2023raw}. The results are depicted in Figure \ref{fig:complicate_isp_result}.
As depicted in the figure, the increased complexity of the mapping between sRGB and raw images, as well as the downscaled image resolution, lead to significant performance degradation across all models in terms of PSNR. However, the proposed method consistently outperforms \cite{nam2022learning} by a substantial margin.

\begin{figure*}
    \begin{subfigure}{\linewidth}
    \includegraphics[width=0.49\linewidth, trim=0 10 0 10, clip]{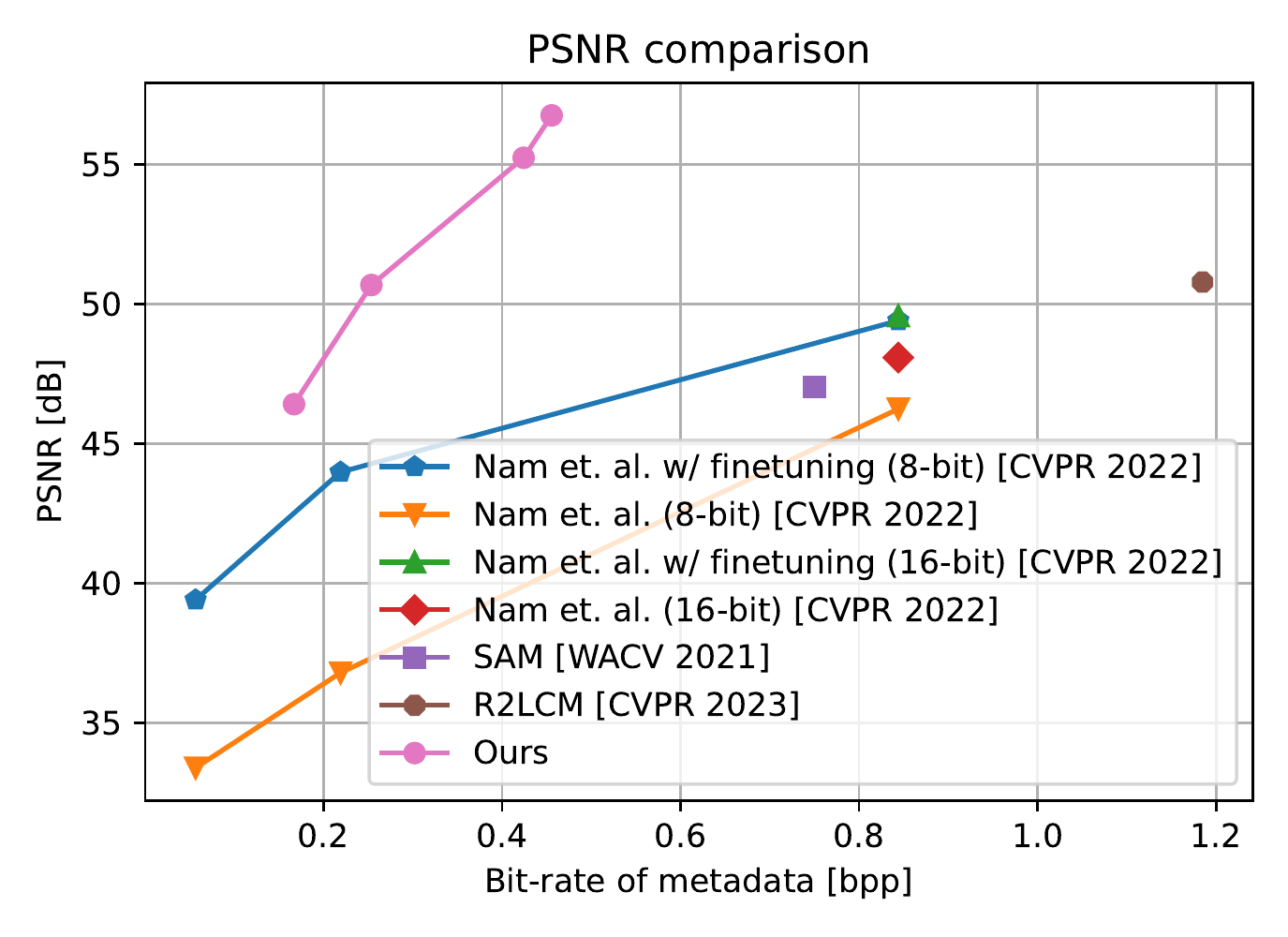}
    \includegraphics[width=0.49\linewidth, trim=0 10 0 10, clip]{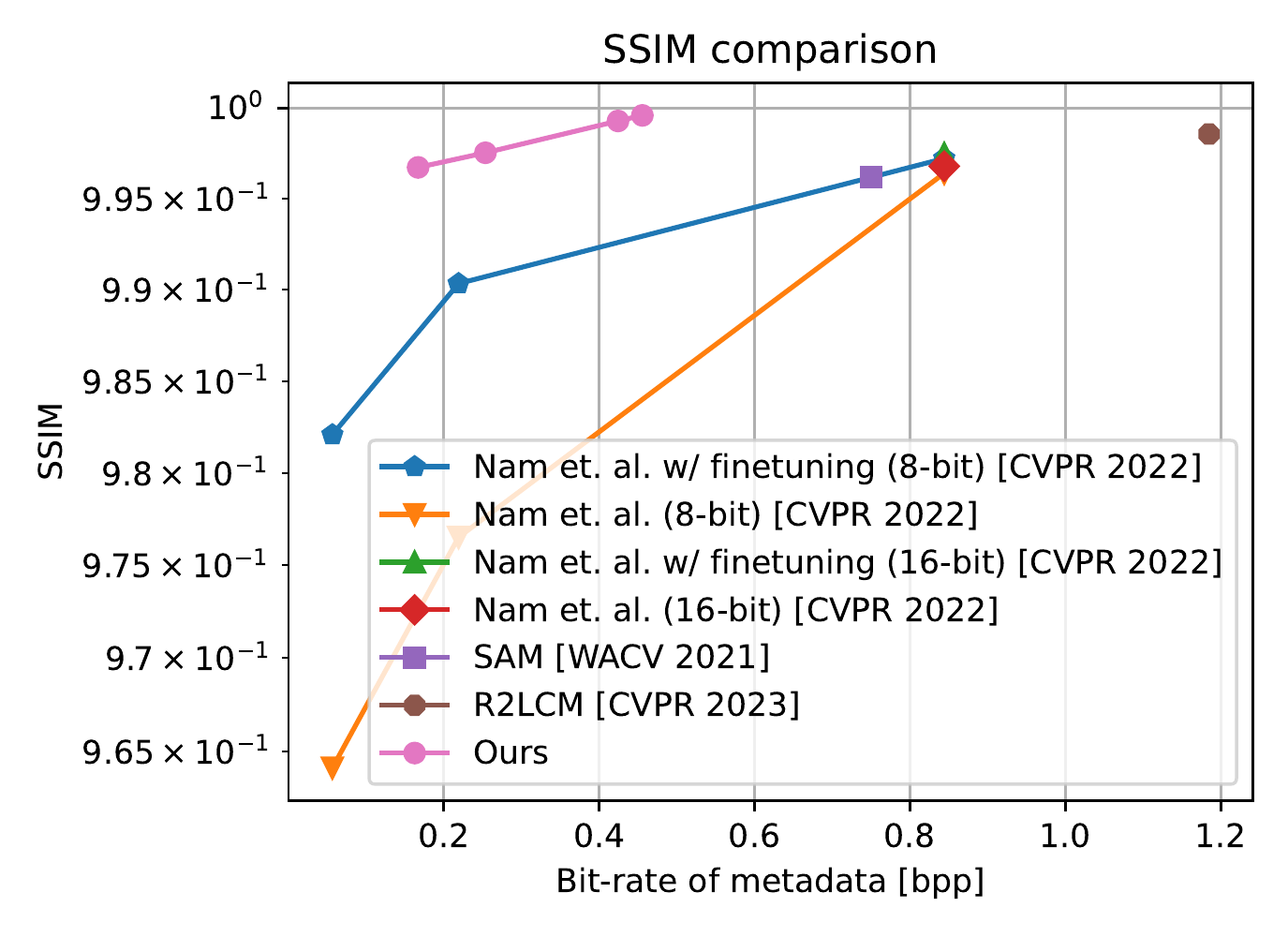}
    \caption{Samsung NX2000}
    \end{subfigure}\\
        \begin{subfigure}{\linewidth}
    \includegraphics[width=0.49\linewidth, trim=0 10 0 10, clip]{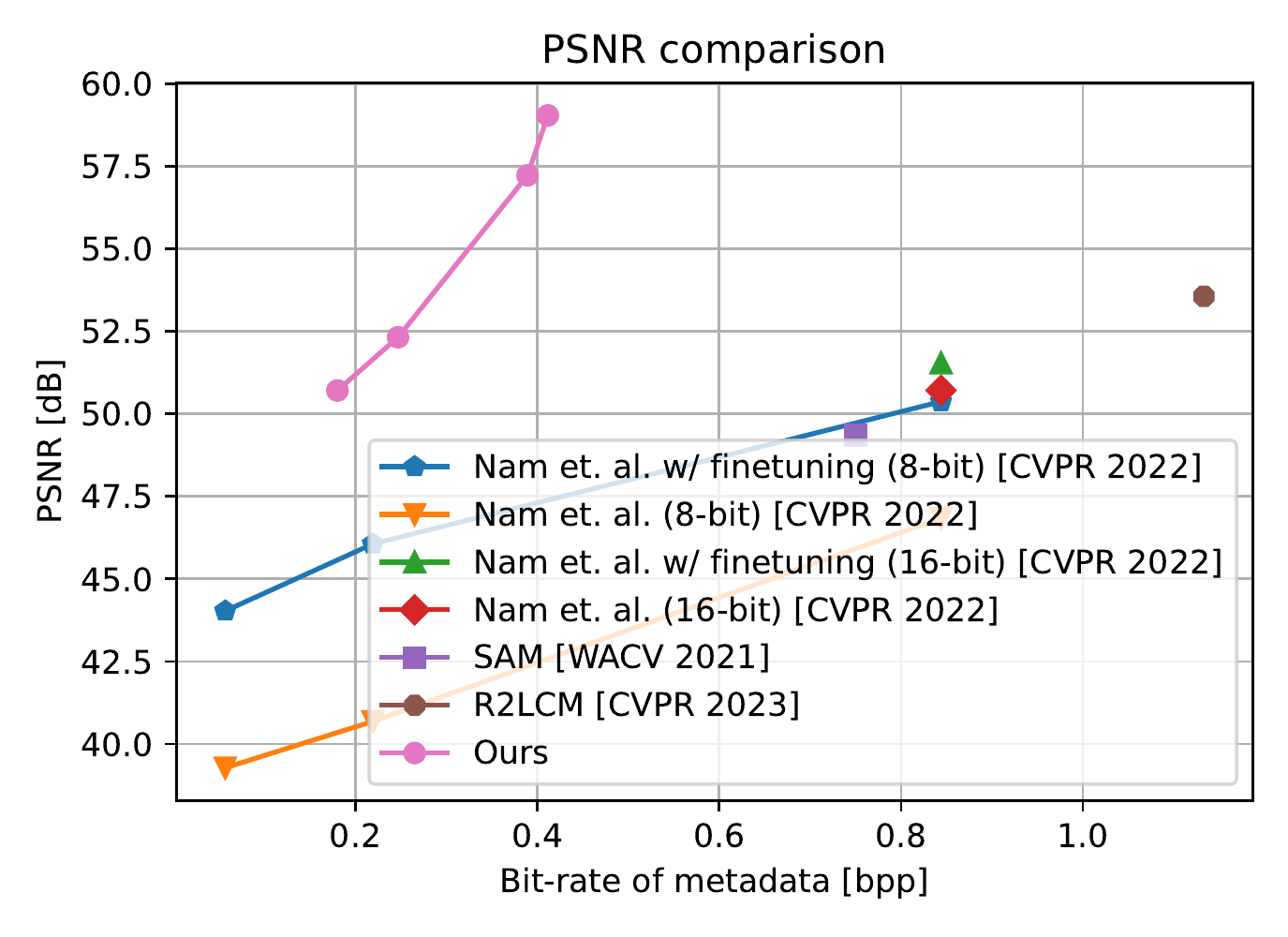}
    \includegraphics[width=0.49\linewidth, trim=0 10 0 10, clip]{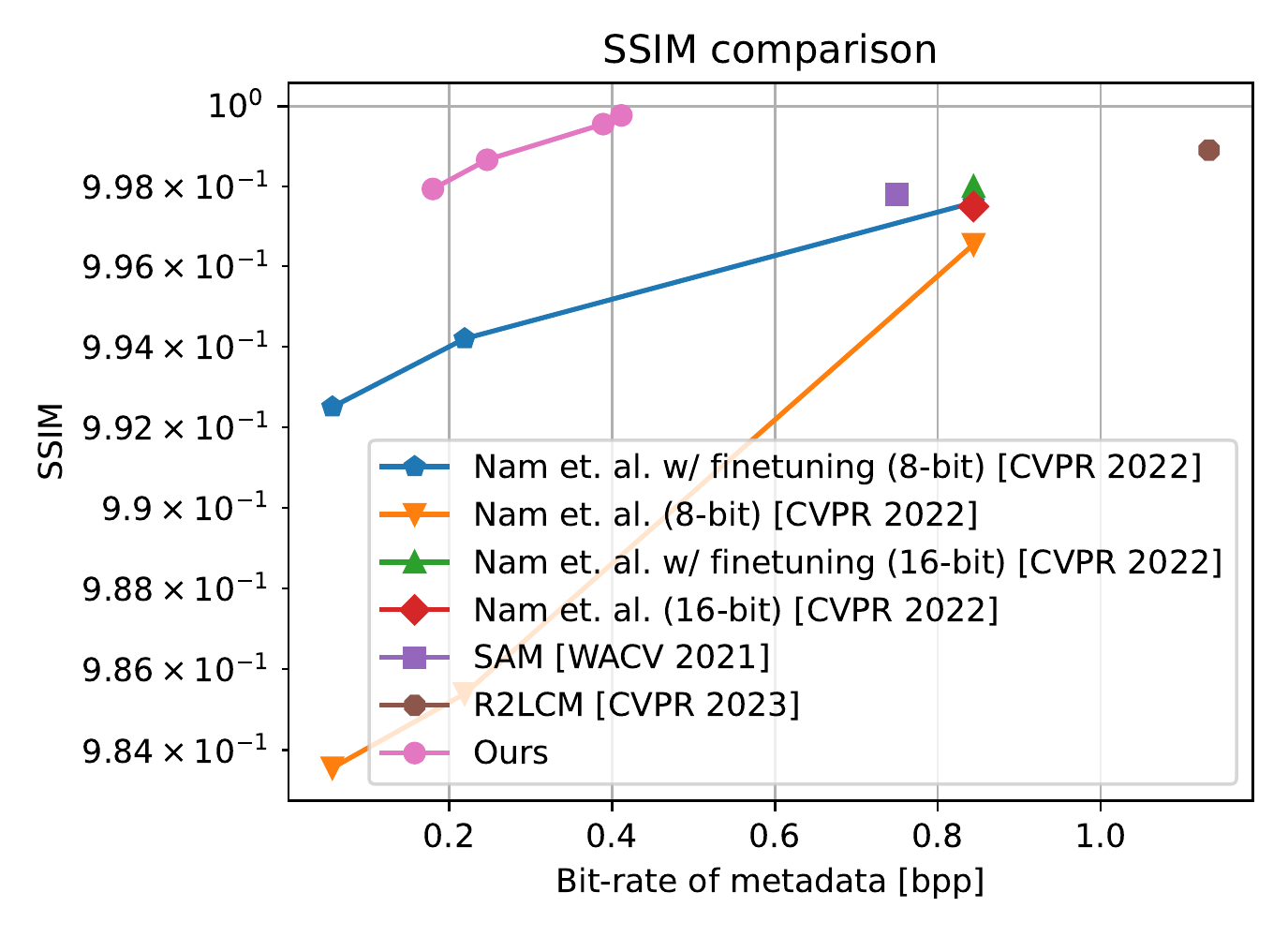}
    \caption{Olympus E-PL6}
    \end{subfigure}\\
        \begin{subfigure}{\linewidth}
    \includegraphics[width=0.49\linewidth, trim=0 10 0 10, clip]{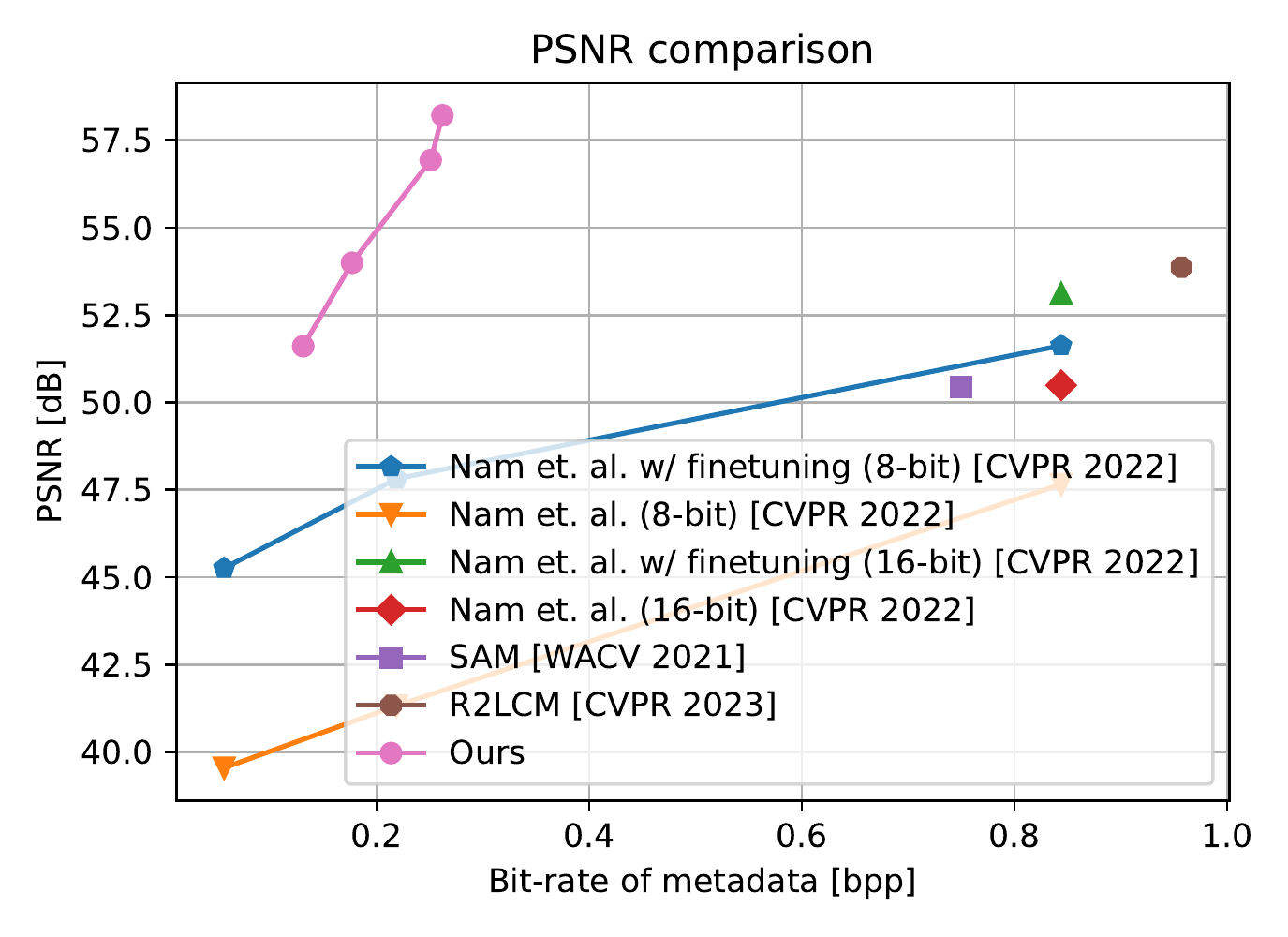}
    \includegraphics[width=0.49\linewidth, trim=0 10 0 10, clip]{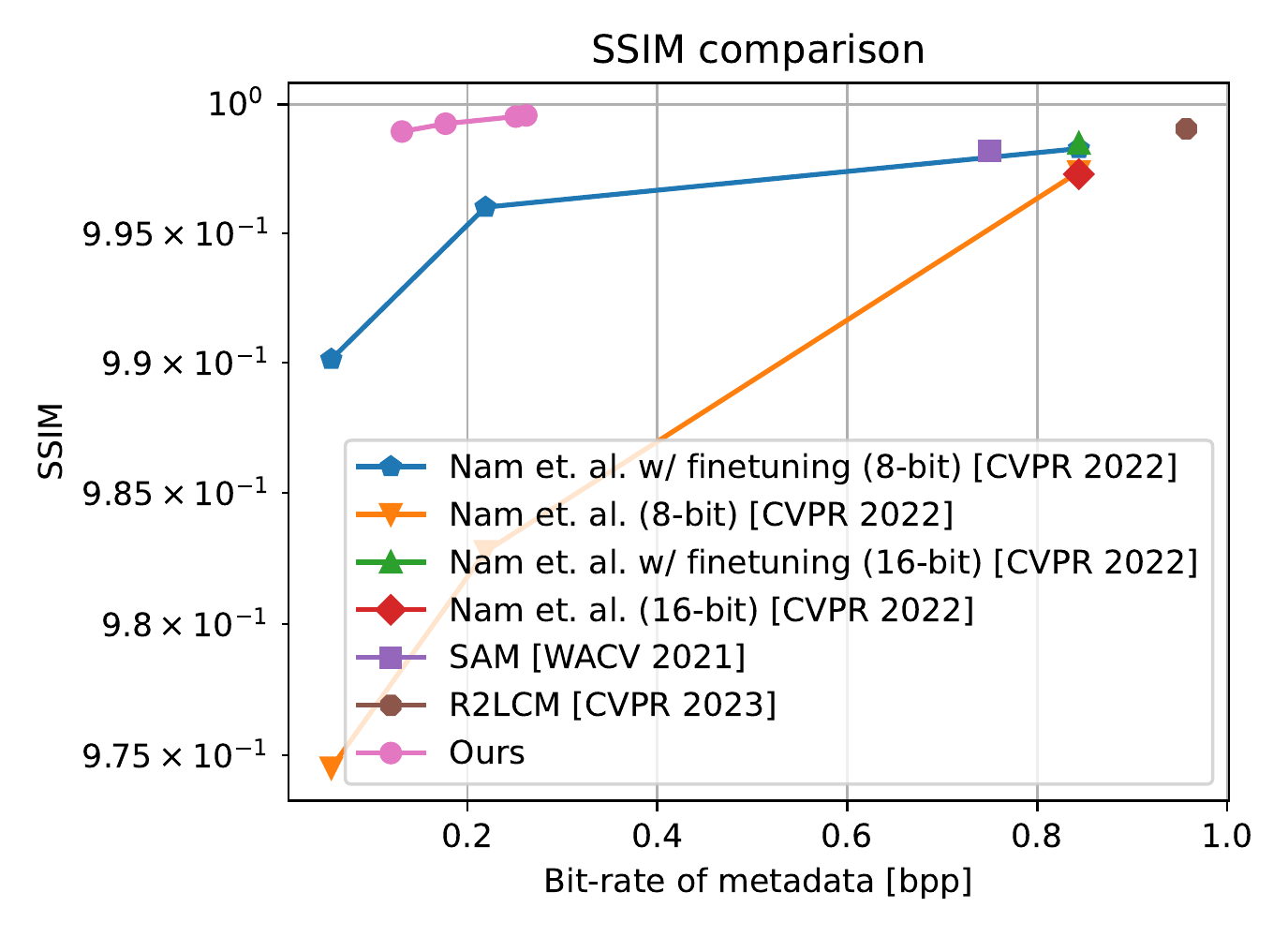}
    \caption{Sony SLT-A57}
    \end{subfigure}
    \caption{RD curves over NUS dataset~\cite{cheng2014illuminant} following the setting of Nam \textit{et. al}~\cite{nam2022learning}. The results in the first and second columns are measured on PSNR and SSIM respectively. For models that provide results under variable rates, only a single model is trained for each curve, \textit{i.e.}, by changing the hyper-parameter of the trained model to obtain different RD performance. The proposed method achieves the best performance under a large range of bit rates among different devices.}
    \label{fig:nus_curve}
\end{figure*}

\noindent\textbf{Results on NUS dataset.}
Unlike our conference version~\cite{wang2023raw}, we adhere to the setting of the benchmark by Nam \textit{et al.}\cite{nam2022learning}, where a downsampled dataset with a factor of 4 instead of the original resolution is utilized. Consequently, we retrain our conference model \cite{wang2023raw} on this dataset and re-evaluate its performance. The results, presented in Table \ref{tab:nus}, demonstrate significant performance improvement compared to the state-of-the-art method by Nam \textit{et al.}\cite{nam2022learning}, including its test-time optimization version, as well as our conference work R2LCM~\cite{wang2023raw}. Furthermore, our approach requires fewer bits of metadata compared with other competing methods. Additionally, we analyze the performance of different models across various coding rates, as shown in Fig. \ref{fig:nus_curve}. The variable bit rates of Nam \textit{et al.}\cite{nam2022learning} is achieved by changing the ratio of sampled raw pixels, and ours is achieved by the proposed strategy. Some visual comparsion can be found in Fig. \ref{fig:nus_main}.

\subsubsection{Results on compressed JPEG images.}

We further explore a more challenging and realistic scenario in which we aim to reconstruct raw images based on compressed JPEG images. To ensure a fair comparison, we maintain the same experimental setup as described in our conference work~\cite{wang2023raw}. Specifically, we employ the original spatial resolution and assess the robustness of our method by training a single model across different devices and JPEG quality factors. The results, presented in Table \ref{tab:nus-compressed}, demonstrate our method's ability to dynamically allocate varying bit rates to JPEG images based on their respective quality factors, \textit{i.e.}, assigning higher bit rates to images with lower JPEG quality. Notably, our method achieves superior reconstruction quality while utilizing the least amount of metadata, outperforming previous SOTA methods as well as our conference work R2LCM~\cite{wang2023raw}.

\begin{table*}[t]
\centering
\caption{The quantitative results of NUS dataset~\cite{cheng2014illuminant} conditioned on the compressed JPEG image with different quality factors. For the metric of bpp, we highlight the comparison of the proposed method with the metadata-based methods that are built on deep neural networks (in gray), which relatively exhibit better performance.}
\setlength\tabcolsep{10pt}
\scalebox{0.9}{
\begin{tabular}{ccccccccc}

\toprule
\multirow{2}{*}{Quality} & \multirow{2}{*}{Method} & \multirow{2}{*}{bpp $\downarrow$} & \multicolumn{2}{c}{Samsung NX2000} & \multicolumn{2}{c}{Olympus E-PL6} & \multicolumn{2}{c}{Sony SLT-A57} \\ 
& &                      & \multicolumn{1}{c}{PSNR $\uparrow$}   & SSIM $\uparrow$  & \multicolumn{1}{c}{PSNR $\uparrow$}  & SSIM $\uparrow$  & \multicolumn{1}{c}{PSNR $\uparrow$}  & SSIM $\uparrow$ \\
\\[\dimexpr-\normalbaselineskip+1pt]
\hline
\\[\dimexpr-\normalbaselineskip+2pt]
\multirow{5}{*}{10} & InvISP & N/A & 26.62 & 0.8836 & 29.12 & 0.8980 & 29.12 & 0.9002  \\
& SAM & 9.556e-4 & 24.42 & 0.8946 & 25.24 & 0.9094 & 25.56 & 0.9110 \\
& SAM & 9.522e-3 & 27.94 & 0.9234 & 28.22 & 0.9376 & 27.83 & 0.9374 \\
\rowcolor[HTML]{F6F6F6}
\cellcolor[HTML]{FFFFFF} 
& Nam \textit{et al.} & 8.438e-1 & 33.06 & 0.9373 & 34.03 & 0.9477 & 34.29 & 0.9506 \\
\rowcolor[HTML]{F6F6F6}
\cellcolor[HTML]{FFFFFF} 
& R2LCM & 7.736e-4 & 33.13 & 0.9386 & 34.04 & 0.9482 & 34.31 & 0.9515 \\
\rowcolor[HTML]{F6F6F6}
\cellcolor[HTML]{FFFFFF} 
& Ours & \best{2.973e-4} & \second{34.11} & \second{0.9476} & \second{35.23} & \second{0.9593} & \second{35.48} & \second{0.9618} \\
\rowcolor[HTML]{F6F6F6}
\cellcolor[HTML]{FFFFFF} 
& Ours &\second{1.298e-1}& \best{36.49} & \best{0.9547}& \best{38.19} & \best{0.9682}& \best{38.72} & \best{0.9710} \\ 
\\[\dimexpr-\normalbaselineskip+1pt]
\hline
\\[\dimexpr-\normalbaselineskip+2pt]
\multirow{4}{*}{30} & InvISP & N/A & 28.71 & 9.9316 & 31.76 & 0.9421 & 30.89 & 0.9459\\
& SAM & 9.556e-4 & 28.88 & 0.9344 & 30.21 & 0.9465 & 29.65 & 0.9458 \\
& SAM & 9.522e-3 & 34.24 & 0.9553 & 35.87 & 0.9648 & 36.12 & 0.9677 \\ 
\rowcolor[HTML]{F6F6F6}
\cellcolor[HTML]{FFFFFF} 
& Nam \textit{et al.} & 8.438e-1 & 37.21 & 0.9630 & 38.70 & 0.9723 & 39.06 & 0.9750\\
\rowcolor[HTML]{F6F6F6}
\cellcolor[HTML]{FFFFFF} 
& R2LCM & {3.613e-4}& {37.40} & {0.9640}& {38.81} & {0.9729}& {39.18} & {0.9757} \\
\rowcolor[HTML]{F6F6F6}
\cellcolor[HTML]{FFFFFF} 
& Ours & \best{2.607e-4} & \second{37.85} & \second{0.9657} & \second{39.42} & \second{0.9752} & \second{39.82} & \second{0.9779} \\
\rowcolor[HTML]{F6F6F6}
\cellcolor[HTML]{FFFFFF} 
& Ours & \second{8.872e-2}& \best{38.67} & \best{0.9676}& \best{40.50} & \best{0.9773}& \best{40.98} & \best{0.9802} \\ %
\\[\dimexpr-\normalbaselineskip+1pt]
\hline
\\[\dimexpr-\normalbaselineskip+2pt]
\multirow{5}{*}{50} & InvISP & N/A & 30.02 & 0.9416 & 32.91 & 0.9529 & 32.97 & 0.9579\\
& SAM & 9.556e-4 & 30.78 & 0.9448 & 32.41 & 0.9559 & 32.05 & 0.9567 \\
& SAM & 9.522e-3 & 36.32 & 0.9629 & 37.77 & 0.9705 & 38.24 & 0.9739 \\
\rowcolor[HTML]{F6F6F6}
\cellcolor[HTML]{FFFFFF} 
& Nam \textit{et al.} & 8.438e-1 & 38.34 & 0.9686 & 40.07 & 0.9767 & 40.04 & 0.9797  \\
\rowcolor[HTML]{F6F6F6}
\cellcolor[HTML]{FFFFFF} & R2LCM & {3.368e-4}& {38.67} & {0.9699}& {40.33} & {0.9776}& {40.73} & {0.9806}\\
\rowcolor[HTML]{F6F6F6}
\cellcolor[HTML]{FFFFFF} 
 & Ours & \best{2.571e-4} & \second{38.93} & \second{0.9707} & \second{40.65} & \second{0.9786} & \second{41.10} & \second{0.9816} \\
\rowcolor[HTML]{F6F6F6}
\cellcolor[HTML]{FFFFFF} 
 & Ours & \second{8.530e-2}& \best{39.49} & \best{0.9718}& \best{41.38} & \best{0.9799}& \best{41.85} & \best{0.9829} \\ 
\\[\dimexpr-\normalbaselineskip+1pt]
\hline
\\[\dimexpr-\normalbaselineskip+2pt]
\multirow{5}{*}{70} & InvISP & N/A & 30.86 & 0.9458 & 32.91 & 0.9553 & 32.97 & 0.9592 \\
& SAM & 9.556e-4 & 32.08 & 0.9529 & 34.14 & 0.9620 & 33.90 & 0.9637 \\
& SAM & 9.522e-3 & 37.42 & 0.9684 & 38.96 & 0.9745 & 39.38 & 0.9780 \\
\rowcolor[HTML]{F6F6F6}
\cellcolor[HTML]{FFFFFF} 
& Nam \textit{et al.} & 8.438e-1 & 39.13 & 0.9724 & 41.01 & 0.9769 & 41.42 & 0.9825 \\
\rowcolor[HTML]{F6F6F6}
\cellcolor[HTML]{FFFFFF} & R2LCM & {3.210e-4}& {39.59} & {0.9742}& {41.36} & {0.9807}& {41.75} & {0.9836}  \\
\rowcolor[HTML]{F6F6F6}
\cellcolor[HTML]{FFFFFF} 
& Ours & \best{2.406e-4} & \second{39.82} & \second{0.9748} & \second{41.62} & \second{0.9814} & \second{42.07} & \second{0.9843}\\
\rowcolor[HTML]{F6F6F6}
\cellcolor[HTML]{FFFFFF} 
&  Ours & \second{8.462e-2}& \best{40.21} & \best{0.9755}& \best{42.13} & \best{0.9822}& \best{42.58} & \best{0.9852}  \\
\\[\dimexpr-\normalbaselineskip+1pt]
\hline
\\[\dimexpr-\normalbaselineskip+2pt]
\multirow{5}{*}{90} & InvISP & N/A & 31.55 & 0.9476 & 33.74 & 0.9598 & 33.68 & 0.9643  \\
& SAM & 9.556e-4 & 34.37 & 0.9663 & 36.60 & 0.9712 & 36.78 & 0.9747 \\
& SAM & 9.522e-3 & 39.17 & 0.9787 & 40.79 & 0.9812 & 41.20 & 0.9843\\
\rowcolor[HTML]{F6F6F6}
\cellcolor[HTML]{FFFFFF} 
& Nam \textit{et al.} & 8.438e-1 & 40.32 & 0.9782 & 42.33 & 0.9838 & 42.82 & 0.9864  \\
\rowcolor[HTML]{F6F6F6}
\cellcolor[HTML]{FFFFFF} 
& R2LCM  & {2.944e-4}& {41.19} & {0.9821}& {42.98} & {0.9856}& {43.43} & {0.9882}\\
\rowcolor[HTML]{F6F6F6}
\cellcolor[HTML]{FFFFFF} 
& Ours & \best{2.319e-4} & \second{41.54} & \second{0.9832} & \second{43.29} & \second{0.9862} & \second{43.79} & \second{0.9888} \\
\rowcolor[HTML]{F6F6F6}
\cellcolor[HTML]{FFFFFF} 
& Ours  & \second{8.691e-2}& \best{41.85} & \best{0.9836}& \best{43.59} & \best{0.9866}& \best{44.10} & \best{0.9893} \\
\bottomrule
\end{tabular}
}
\label{tab:nus-compressed}
\end{table*}

    

\subsection{Ablation study}
\begin{figure}[tbp]
    \centering
    \scalebox{0.84}{
    \hspace{-0.4cm}
    \subcaptionbox{input}{
    \includegraphics[height=0.65\linewidth, clip]{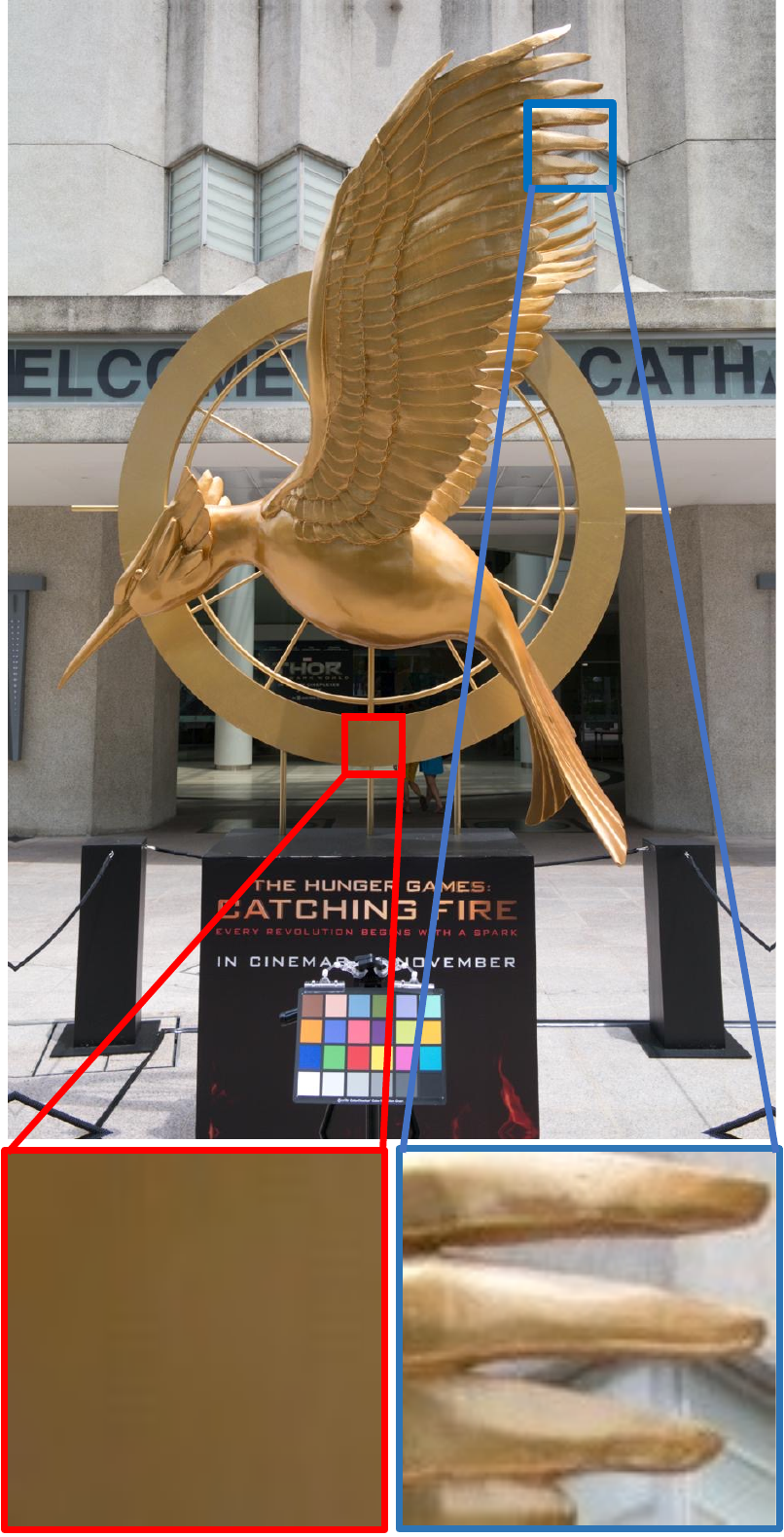}
    }
    \hspace{-0.2cm}
    \subcaptionbox{Nam \textit{et al.}. \cite{nam2022learning}\\
    bpp: 0.8443\\
    PSNR: 45.08}{
    \includegraphics[height=0.65\linewidth]{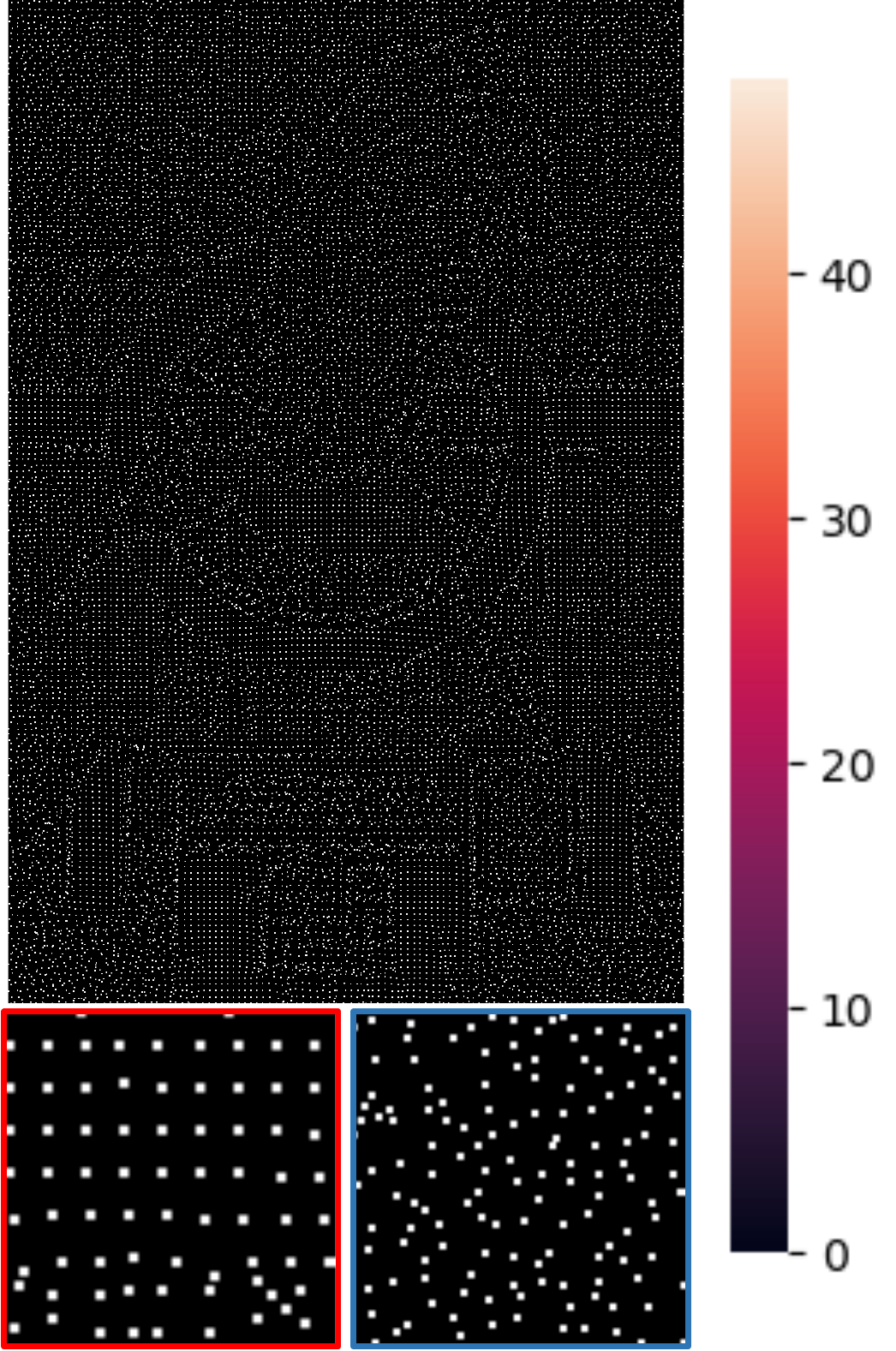}}
    \hspace{-0.3cm}
    \subcaptionbox{Ours\\
    bpp: 0.4162\\
    PSNR: 52.26}{
    \includegraphics[height=0.65\linewidth]{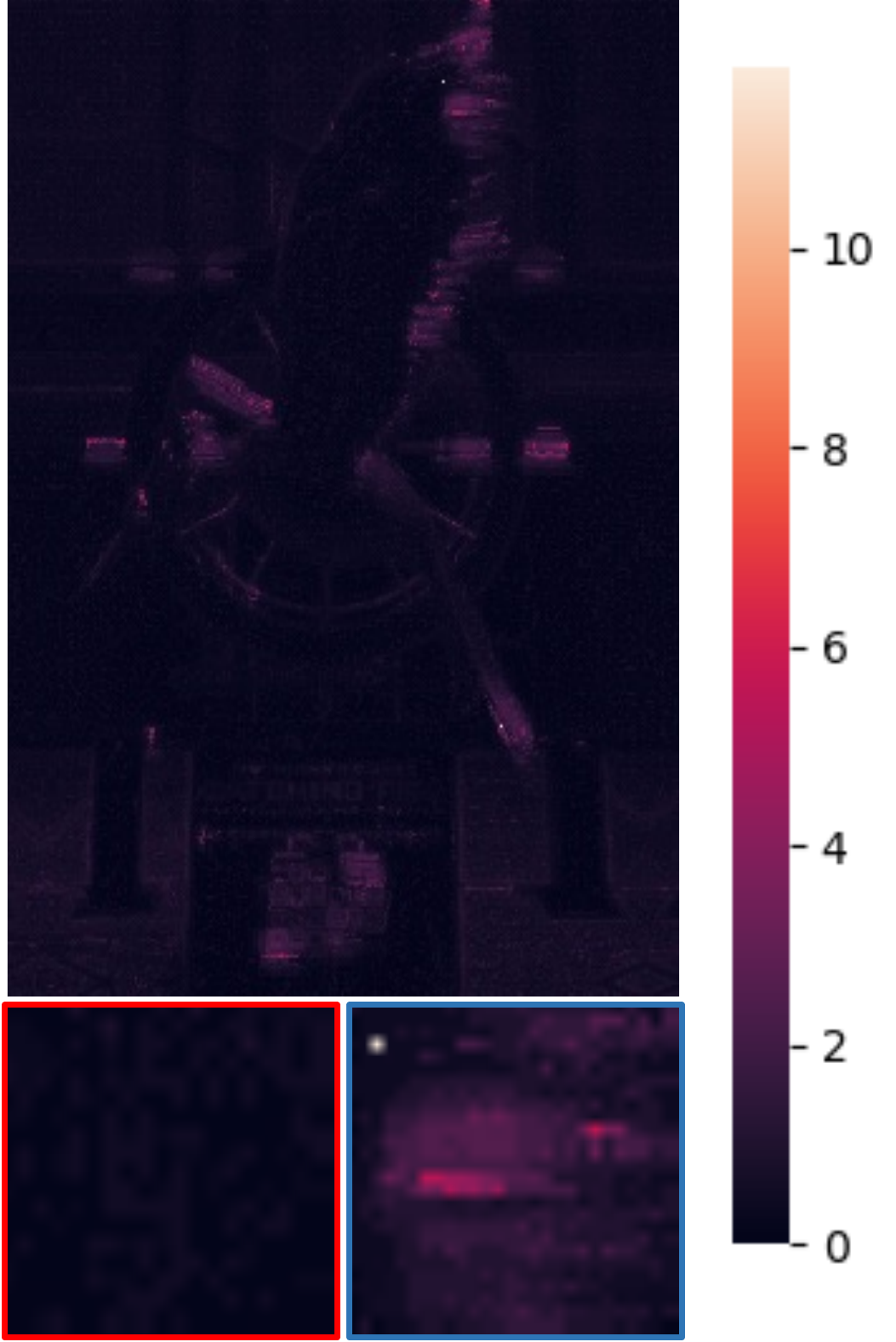}}\hspace{-0.2cm}}
    \caption{
    The comparison of the bit allocation. 
    (a) The input sRGB image. 
    (b) The bits allocation map of Nam \textit{et al.}. \cite{nam2022learning}.
    (c) The bits allocation map of the proposed method.
    For better visualization, we enlarge the size of sampled pixels in (b). 
    It is worth noting that each sampled raw pixel needs 48 bits to save in (b). 
    \textit{Best zoom in for more details.}
    }
    \label{fig:sampling}
\end{figure}

\noindent\textbf{The comparison of bit allocation.}
The proposed method offers a notable advantage by enabling end-to-end learning of bit allocation. 
As depicted in Fig. \ref{fig:isp}, the non-uniform loss of information emphasizes the importance of an effective bit allocation algorithm in metadata-based raw image reconstruction. 
Therefore, we visually compare the bit allocation of current SOTA methods with our proposed approach in Fig. \ref{fig:sampling}.
In contrast to the uniform sampling approach in SAM~\cite{punnappurath2021spatially}, Nam \textit{et al.}~\cite{nam2022learning} propose a superpixel-based sampling network. 
However, the training of the reconstruction and sampling networks is conducted in two separate phases.
Besides, although their sampling method exhibits local non-uniformity, it maintains global uniformity, thereby limiting the coding efficiency and reconstruction quality.
As shown in the figure, our method demonstrates the capability to adaptively allocate different bits to different areas. 
Specifically, for flat areas (\textit{e.g.}, the red bounding box), our approach utilizes a minimal number of bits for encoding. Conversely, in areas with more complex contextual information (\textit{e.g.}, the boundary area within the blue bounding box), our method allocates a relatively larger number of bits. Furthermore, even in areas where a large bit rate is assigned, the bit depth of the latent features required by our method is still much lower than sampling in the raw pixel space with a fixed bit depth.

\begin{figure}
\centering
    \scalebox{0.84}{
    \hspace{-0.4cm}
    \subcaptionbox{input and \protect\\$-\log_2(q(\z))$}{
    \includegraphics[height=0.55\linewidth, clip]{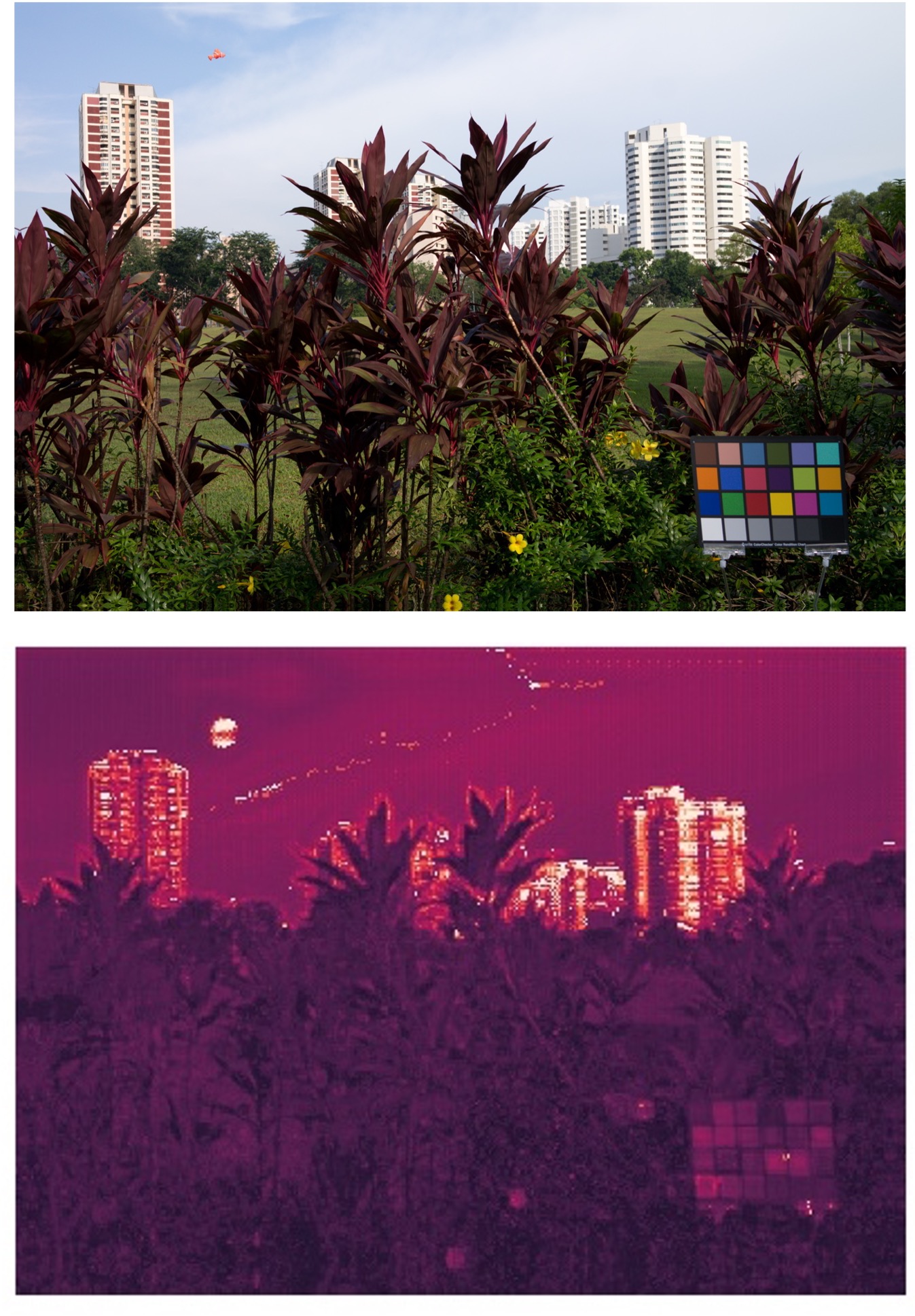}
    }
    \hspace{-0.42cm}
    \subcaptionbox{$\mathbf{M}^0$ and \protect\\$-\log_2(q(\z|\mathbf{M}^0\odot \z))$}{
    \includegraphics[height=0.55\linewidth]{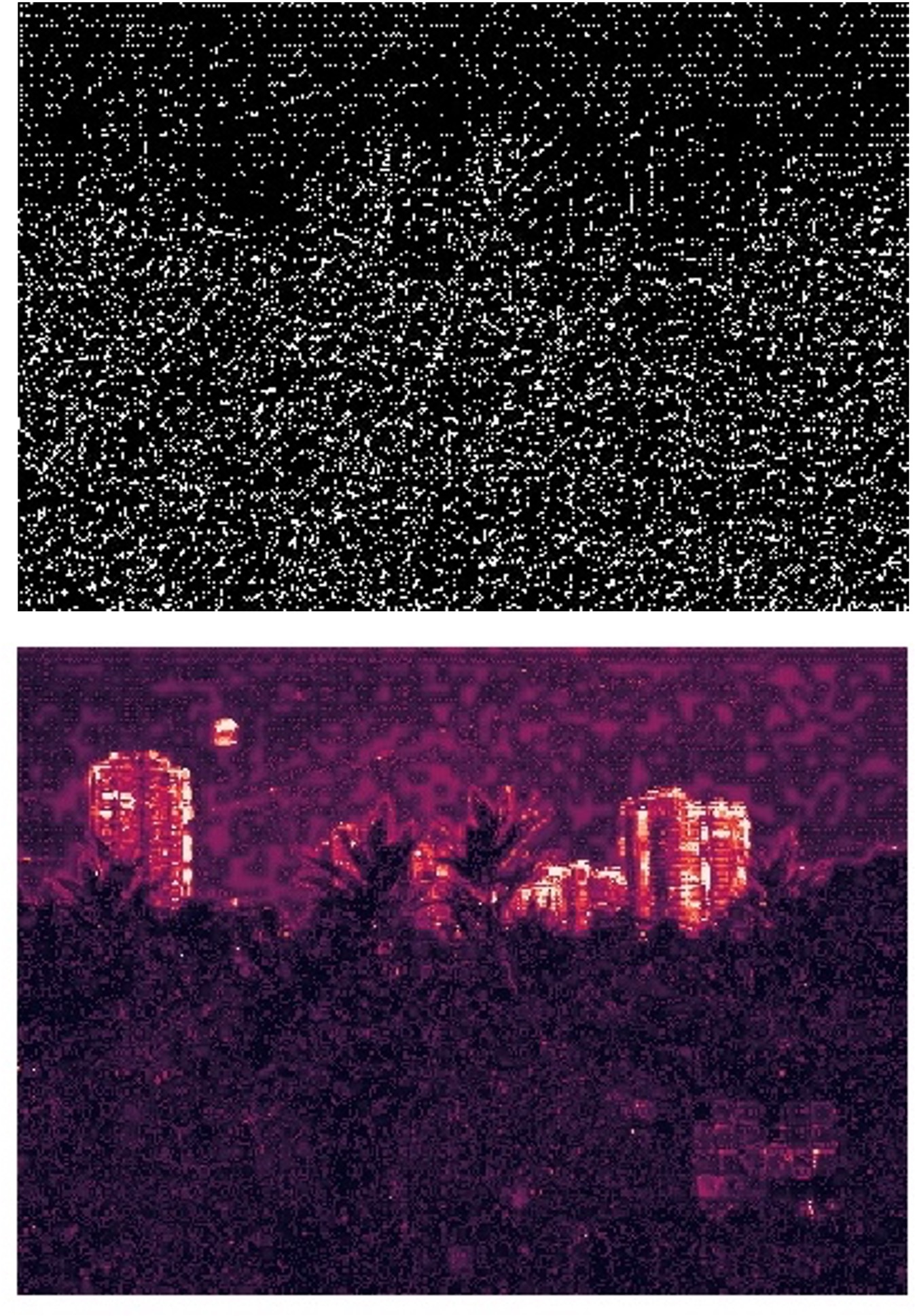}}
    \hspace{-0.35cm}
    \subcaptionbox{$\mathbf{M}^1$ and \protect\\$-\log_2(q(\z|\sum_{i=0}^1\mathbf{M}^i \cdot \z))$}{
    \includegraphics[height=0.55\linewidth]{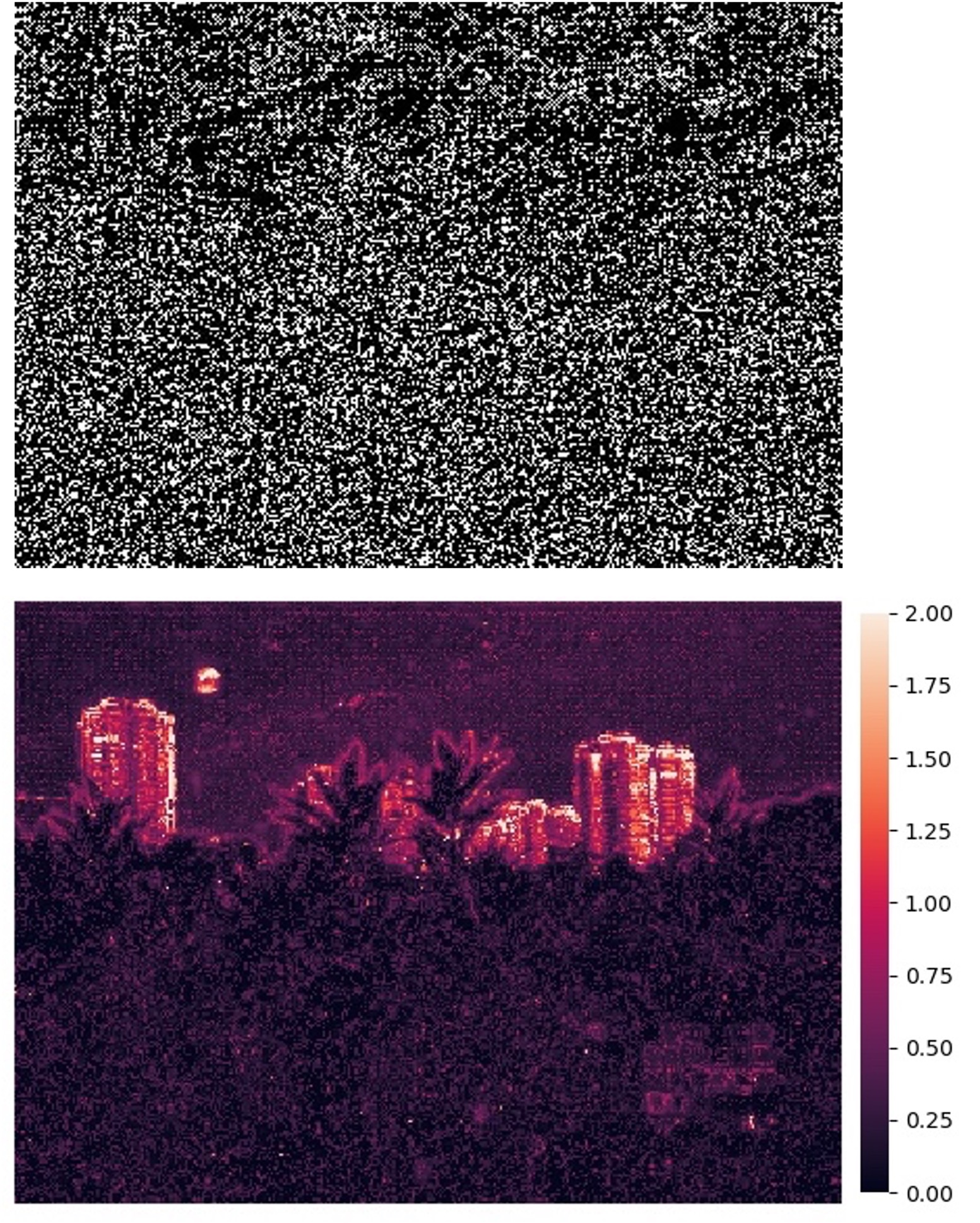}}\hspace{-0.2cm}}
\caption{
The visualization of the effect of the sRGB-guided context model.
The images in the first row are the sRGB image, the sampling mask of the first step $\mathbf{M}^0$, and the second step $\mathbf{M}^1$, respectively.
The images in the second row are the needed bit map w/o the prior knowledge and w/ the already encoded latent features in $\mathbf{M}^0$ and $\mathbf{M}^0+\mathbf{M}^1$ from left to right. As shown in the figure, the proposed context model can significantly improve the coding efficiency.
}
\label{fig:context_vis}
\end{figure}

\noindent\textbf{The sRGB-guided context model.} 
To gain a deeper understanding of the proposed context model's functionality, we visualize each step of the encoding/decoding process of $\mathbf{z}$ in Fig. \ref{fig:context_vis}.
This visualization reveals the significant relationship between the order of the compression/decompression process and the image context, showcasing the effective utilization of sRGB image information by our proposed context model.
The sparse sampling masks assist the model in better predicting the distribution of adjacent to-be-processed latent features. 
Additionally, our method gradually increases the number of sampled pixels in the latent space as more pixels become available, enabling better prediction of the distribution of unseen pixels and resulting in enhanced coding efficiency.
As shown in Fig. \ref{fig:context_vis} (b-c), with the assistance of the already decoded information $\M^0 \odot \mathbf{\hat{z}}$ and $(\M^0 + \M^1) \odot \mathbf{\hat{z}}$, we gradually achieve a more and more accurate estimation of the likelihood (\textit{i.e.}, a smaller bit rate) of $\mathbf{z}$. 
We also conduct quantitative evaluations to assess the effectiveness of our proposed sRGB-guided context model, as shown in Table \ref{tab:ablation_context}. 
We compare our method with two baseline models: one without the context model and another that encodes and decodes with the reversed predicted order masks. 
The table clearly illustrates that our model, which incorporates the proposed context model, achieves similar reconstruction quality while utilizing only approximately 50\% of the metadata space. 
Furthermore, compared to the model with the context model using the reversed order, our model also demonstrates significant metadata savings.
These results emphasize the importance of the encoding/decoding order and highlight the effectiveness of our approach.

\begin{table}[htbp]
\centering
\caption{The ablation study of the proposed sRGB-guided context model. We evaluate the performance of models on the released  dataset of \cite{nam2022learning}. ``context model in \cite{wang2023raw}" means we replace the proposed improved context model in this work with the one in~\cite{wang2023raw} and keep others the same, and ``w/ reversed order" indicates the utilization of the reversed order of the predicted order masks. Since the quantized value is influenced by the order of encoding/decoding, adopting different processing orders can lead to slight differences in the reconstructed results.}

\begin{tabular}{cccccc}
\toprule
 & & {BPP} & {PSNR} & {SSIM} \\
\midrule
\multirow{4}{*}{\rotatebox{90}{\textbf{Sony}}} 
& context model in \cite{wang2023raw} & 0.7823 & 60.35 & 0.99975 \\
& w/o context model & 0.5621 & 58.25 & 0.99958 \\
& w/ reversed order & 0.3740 & 58.25 & 0.99958 \\
& w/ context model & \textbf{0.2619} & 58.21 & 0.99958 \\
\midrule
\multirow{4}{*}{\rotatebox{90}{\textbf{Olym}}} & context model in \cite{wang2023raw} & 0.9172 & 59.51 & 0.99979 \\
& w/o context model & 1.0411 & 59.72 & 0.99979 \\
& w/ reversed order & 0.6392 & 59.06 & 0.99969 \\
& w/ context model & \textbf{0.4115} & 59.04 & 0.99969 \\
\midrule
\multirow{4}{*}{\rotatebox{90}{\textbf{Sams}}} 
& context model in \cite{wang2023raw} & 0.7837 & 57.80 & 0.99971 \\
& w/o context model & 0.8368 & 57.06 & 0.99965 \\
& w/ reversed order & 0.5711 & 56.75 & 0.99961 \\
& w/ context model & \textbf{0.4555} & 56.75 & 0.99961 \\
\bottomrule
\end{tabular}

\label{tab:ablation_context}
\end{table}

\noindent\textbf{Different backbone designs.} 
We evaluate the performance of different backbone designs and the results are shown in Table \ref{tab:ablation_bacbone}. The table reveals that the proposed asymmetric hybrid design outperforms other strategies in terms of RD performance. Specifically, the proposed asymmetric hybrid design enables us to achieve superior reconstruction quality while minimizing storage overhead. It is noteworthy that although the resolution-maintained backbone has the potential for higher reconstruction quality, reaching this potential becomes challenging due to the constraint of the rate loss and a more serious approximation error of the quantization step.

\begin{table}[tbp]
\centering
\caption{The ablation study of different backbone designs evaluated on NUS-Sony dataset~\cite{nam2022learning,cheng2014illuminant}. ``-" represents the spatial resolution keeps unchanged through the network.}

\begin{tabular}{cccccc}
\toprule
Design & $g_a^0$ & $g_s^0$ & {BPP} & {PSNR}  \\
\midrule
Downsampling oriented & $\downarrow 16$ & $\uparrow 16$ & 0.13 & 48.37 \\
Resolution maintained & - & - & 1.36 & 55.04 \\
Finetuned downsampling & $\downarrow 2$ & $\uparrow 2$ & 1.06 & 56.27 \\
\\[\dimexpr-\normalbaselineskip+1pt]
\hline
\\[\dimexpr-\normalbaselineskip+2pt]
\multirow{2}{*}{Asymmetric hybrid} & $\downarrow 4$ & - & 0.26 & 58.21 \\
& $\downarrow 4$ & - & 0.13 & 51.61 \\
\bottomrule
\end{tabular}
\label{tab:ablation_bacbone}
\end{table}

\begin{figure}[tbp]
    \centering
    \begin{subfigure}{0.48\linewidth}
        \includegraphics[width=\linewidth]{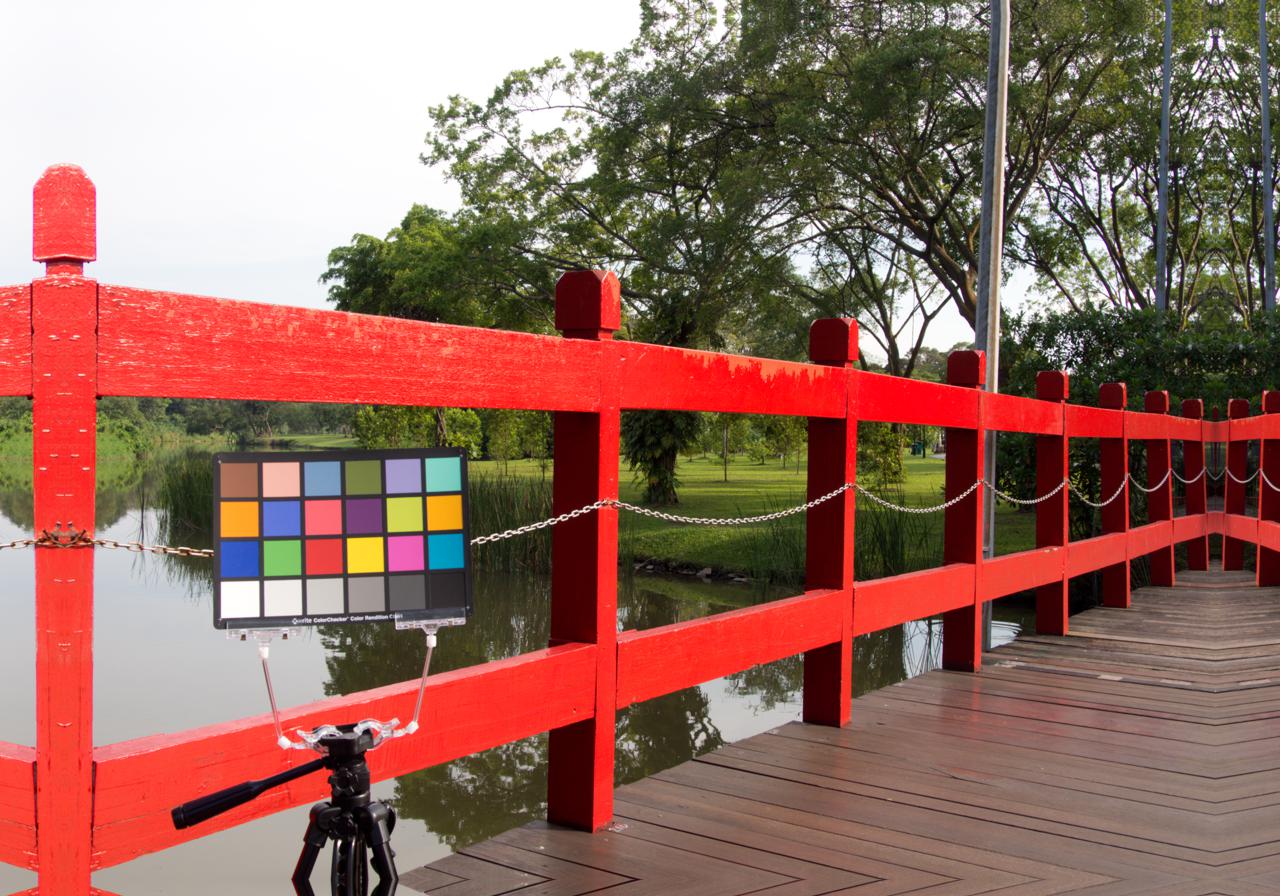}
    \caption{sRGB image}
    \end{subfigure}
    \begin{subfigure}{0.48\linewidth}
        \includegraphics[width=\linewidth]{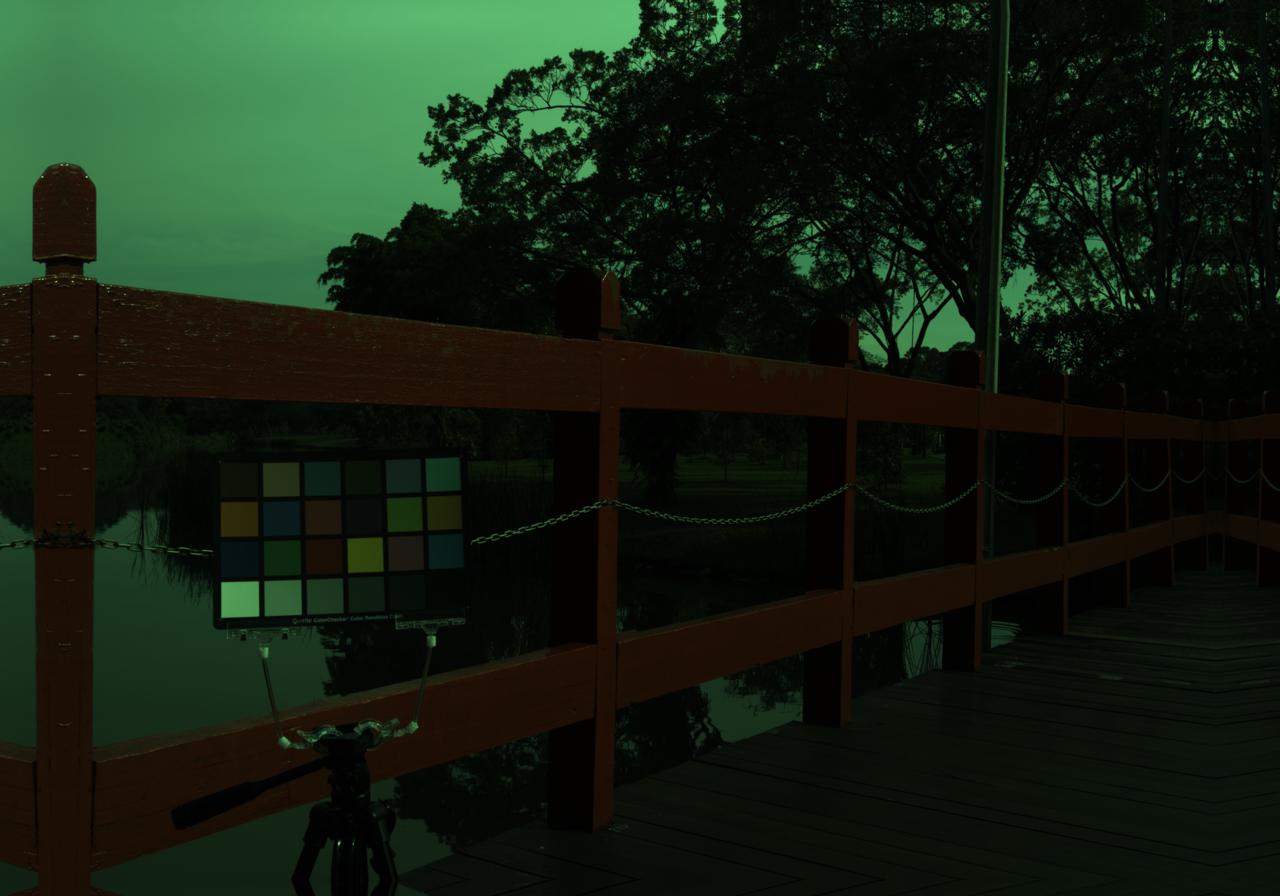}
    \caption{Raw image}
    \end{subfigure}

    \begin{subfigure}{0.49\linewidth}
        \includegraphics[width=\linewidth]{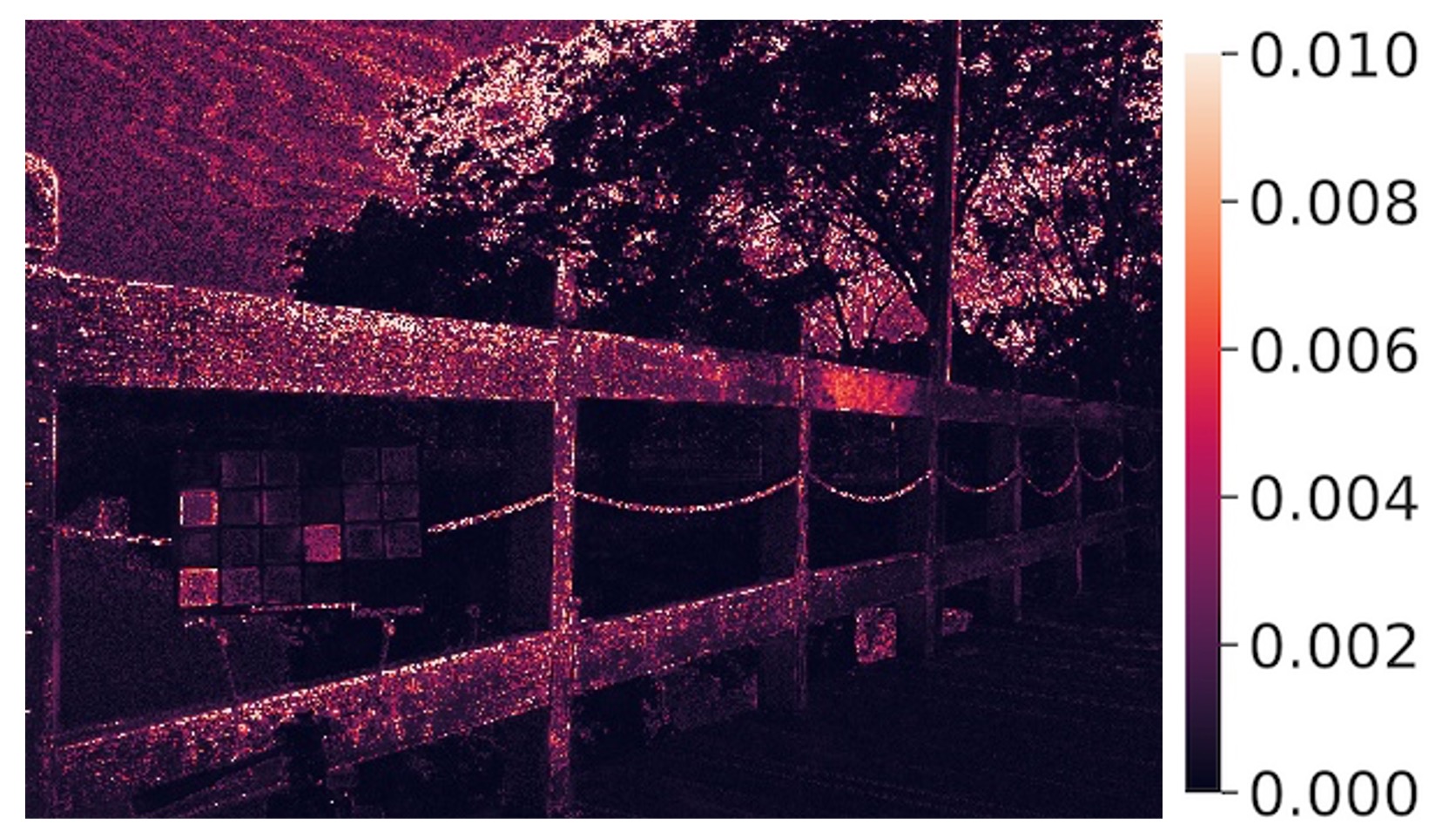}
    \caption{Error map}
    \end{subfigure}
    \begin{subfigure}{0.49\linewidth}
        \includegraphics[width=\linewidth]{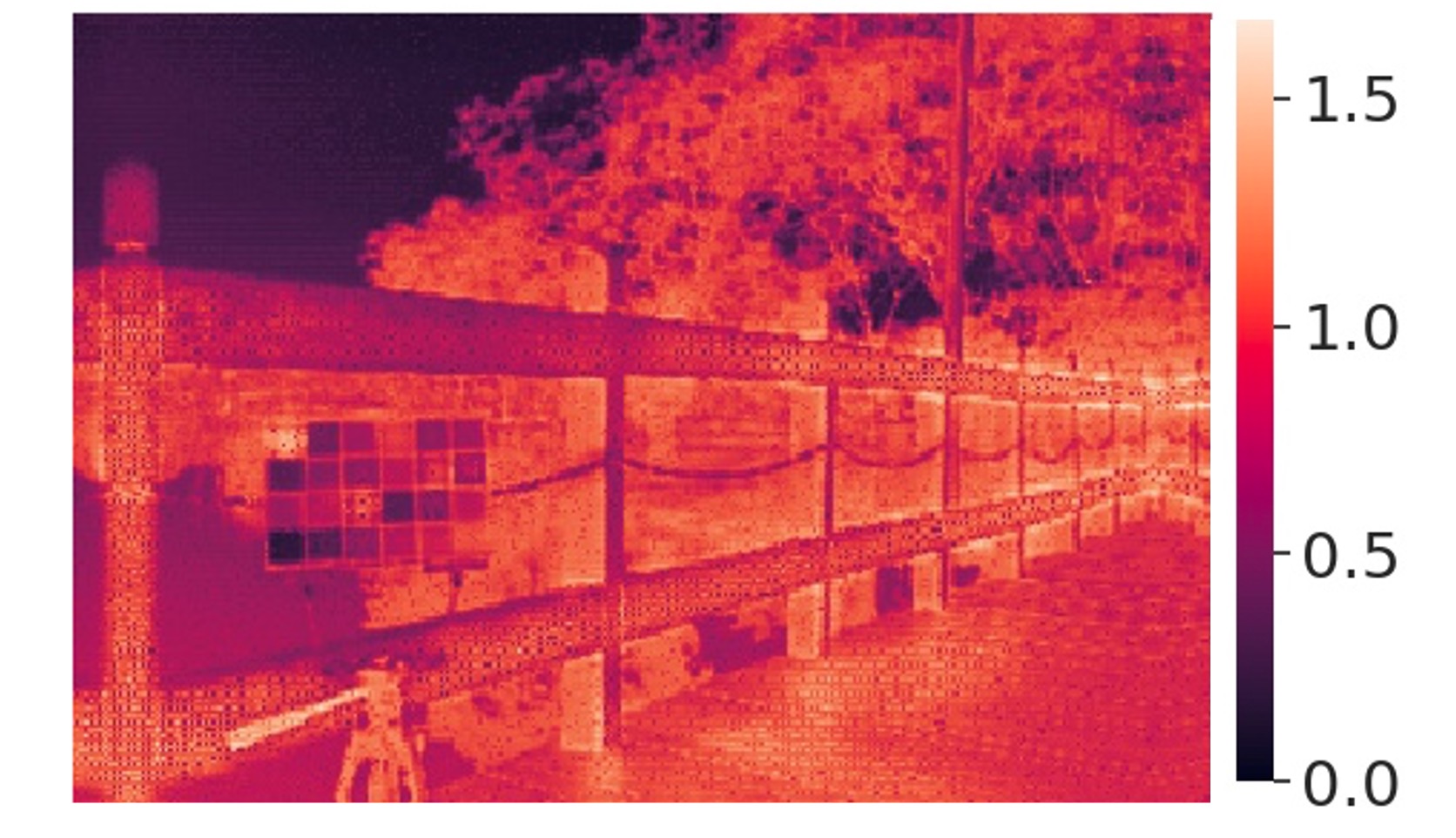}
    \caption{Quantization precision}
    \end{subfigure}
    \caption{An illustration of the proposed adaptive quantization bin width. We visualize the quantization precision in (d) by averaging the quantization bin width in the channel dimension. As we can see in the figures, the proposed model adaptively allocates more bits to the highlights, which exhibit higher error, and to areas with complex contextual information.}
    \label{fig:quant_error}
\end{figure}

\begin{figure}[tbp]
    \centering

    \includegraphics[width=0.8\linewidth]{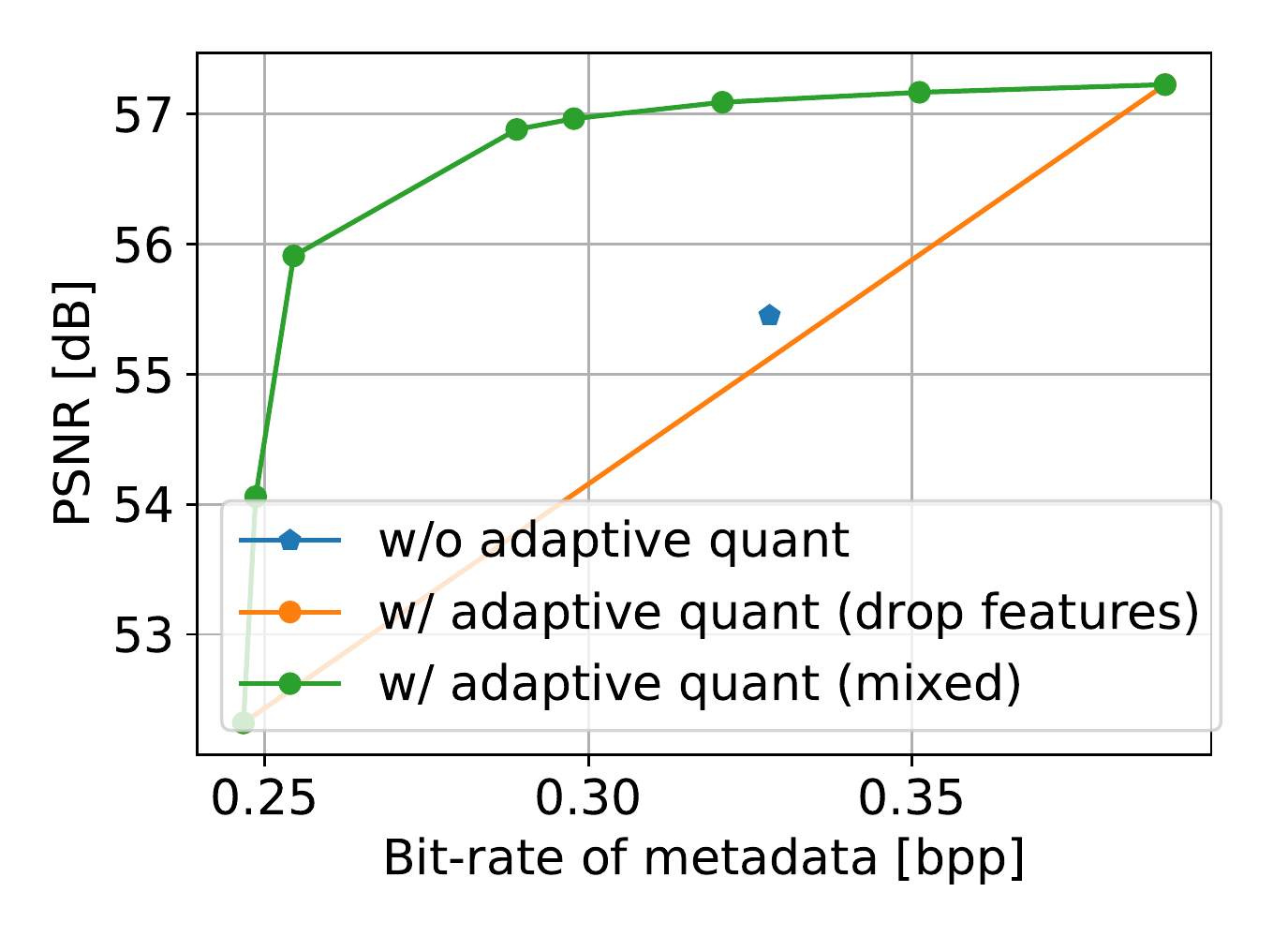}
    \caption{A comparison of the model trained w/ and w/o the proposed adaptive quantization strategy. The model trained using adaptive quantization precision demonstrates superior rate-distortion performance. Besides, we employed a ``mixed" approach to achieve performances under different bit rates by utilizing the proposed variable bit rate strategy, enabling a continuous adjustment of the bit rate. ``drop features" means we discard features in different numbers of sampling masks.}
    \label{fig:adaptive_quant}
\end{figure}

\noindent\textbf{The adaptive quantization strategy.}
We conducted a comprehensive evaluation of the effectiveness of the proposed adaptive quantization strategy, which dynamically assigns varying levels of quantization precision to different regions. 
We trained models with and without the adaptive quantization strategy and present the results in Fig.~\ref{fig:adaptive_quant}. 
The figure clearly illustrates that the model trained with the adaptive quantization strategy outperforms the model without it, as indicated by its curve position above the latter's point. 
This demonstrates that the enhanced representation capability of the model greatly contributes to improved rate-distortion (RD) performance. 
Additionally, we visually depict the predicted bin width in Fig. \ref{fig:quant_error}. The figure shows that the model tends to allocate higher quantization precision to highlights, which exhibit higher error, and to areas with complex contextual information.

\begin{table*}[tbp]
    \centering
    \caption{The comparison of the computational cost of different methods used in Table~\ref{tab:nus}, given an $896\times 1280$ input. \textit{N/A} in SAM~\cite{punnappurath2021spatially} means it is a parameter free method. The arithmetic coding in our method can be further accelerated by running it on GPU.} 
    \scalebox{0.97}{
    \begin{tabular}{ccccc}
    \toprule
         & Parameters & FLOPs & Compress \& Decompress time\\
    \midrule
        InvISP~\cite{xing2021invertible} & 1.41MB & 1.50T & 0.43s\\
        SAM~\cite{punnappurath2021spatially} (bpp 9.56e-4) & N/A & -- & 1.41s\\
        SAM~\cite{punnappurath2021spatially} (bpp 9.52e-3) & N/A & -- & 8.87s\\
        Nam~\textit{et al.}\cite{nam2022learning} & 15.55MB & 438G  & 0.05s (w/o online fine-tuning) \\
        Nam~\textit{et al.}\cite{nam2022learning} & 15.55MB & 4.227T  & 0.95s (w/ online fine-tuning) \\
        R2LCM~\cite{wang2023raw} & 0.55MB & 527.40G & 0.91s (0.74s of arithmetic coding on CPU)\\
        \rowcolor[HTML]{EFEFEF} Ours & 2.35MB & 223.21G & 0.32s (0.11s for encoding and 0.21s for decoding)\\
    \bottomrule
    \end{tabular}}
    \label{tab:cost}
\end{table*}

\subsection{Computational cost}
Our method can be trained and evaluated on a single RTX A5000 GPU. We conducted a comparative analysis of the computational costs between our proposed method and other SOTA methods, as presented in Table 1.
The table demonstrates that our proposed method achieves comparable or even faster speeds than existing SOTA methods such as InvISP~\cite{xing2021invertible}, the test-time model SAM~\cite{punnappurath2021spatially}, and our previous conference work~\cite{wang2023raw}. 
Furthermore, we achieve a notable increase in speed compared to the online fine-tuning version of Nam et al. (2022) \cite{nam2022learning}. In their approach, online fine-tuning plays a crucial role in achieving better rate-distortion (RD) performance as shown in Fig. \ref{fig:nus_curve} and Table \ref{tab:nus}.
Furthermore, the speed bottleneck of our proposed method can be further mitigated by utilizing existing public libraries for GPU implementation of arithmetic coding.
Additionally, it is worth noting that the computational cost of encoding is considerably smaller than that of decoding in our work, resulting in lower latency during the capture phase.

\section{Conclusion}

In this work, we introduce a novel framework for reconstructing raw images using learned compact metadata. Unlike previous SOTA methods that sample in the raw pixel domain, our proposed end-to-end learned coding technologies enable the encoding of metadata in the latent space using an adaptive bits allocation strategy. This approach yields improved reconstruction quality and higher coding efficiency.
Furthermore, we present several advancements compared to our previous conference version. These include enhanced designs of the backbone, context model, quantization, and training strategy, resulting in further improvements in rate-distortion performance. Additionally, we propose a variable bitrate strategy, allowing us to cover a wide range of bitrates seamlessly within a single model.
We conduct extensive comparisons of our proposed method on various benchmark datasets. The results consistently demonstrate that our method surpasses both existing SOTA methods and our previous conference work in terms of both the reconstruction quality and coding effiency.

\backmatter

\section*{Declarations}

\noindent \textbf{Funding.} This work was done at Rapid-Rich Object Search (ROSE) Lab, Nanyang Technological University. This research is supported in part by the NTU-PKU Joint Research Institute (a collaboration between the Nanyang Technological University and Peking University that is sponsored by a donation from the Ng Teng Fong Charitable Foundation), the Basic and Frontier Research Project of PCL, the Major Key Project of PCL, and the MOE AcRF Tier 1 (RG61/22) and Start-Up Grant.

\noindent \textbf{Conflict of interest.}  The authors declare that they have no known competing financial interests or personal relationships that could have appeared to influence the work reported in this paper.

\noindent \textbf{Availability of data and materials.} This work does not propose a new dataset. All the datasets we used are publicly available. 

\noindent \textbf{Code availability.} The code of this work will be released after acceptance.

\bibliographystyle{sn-vancouver} 
\bibliography{egbib}

\end{document}